\documentclass{article} 
\usepackage{iclr2025_conference,times}

\usepackage{hyperref}
\usepackage{url}

\usepackage{graphicx}
\usepackage{multirow}
\usepackage{algorithm}
\usepackage{algpseudocode}
\usepackage{amsmath, amssymb}
\usepackage{color}
\usepackage{colortbl}
\usepackage{listings}  

\usepackage[normalem]{ulem}
\useunder{\uline}{\ul}{}

\title{DetailCLIP: \\ Detail-Oriented CLIP for Fine-Grained Tasks}

\author{
  Amin Karimi Monsefi$^{\S}$\thanks{Equal contribution.} \quad
  Kishore Prakash Sailaja$^{\S}$\footnotemark[1] \quad
  Ali Alilooee$^{\S}$ \quad \\
  \textbf{\hspace*{0.2em}Ser-Nam Lim$^{\ddag}$ \quad
  Rajiv Ramnath$^{\S}$} \\
 \texttt{\shortstack[l]{\{karimimonsefi.1, prakashsailaja.1, alilooeedolatabad.1,\\ ramnath.6\}@osu.edu, sernam@ucf.edu}} \\
  $^{\S}$The Ohio State University,\quad
  $^{\ddag}$University of Central Florida
}

\iclrfinalcopy 
\begin{document}

\maketitle

\begin{abstract}
    In this paper, we introduce DetailCLIP, a self-improving vision-language foundation model designed to enhance fine-grained feature understanding through self-supervised learning. Foundation models like CLIP have demonstrated strong performance in global image-text alignment but often fail to capture detail-oriented features necessary for tasks such as segmentation. To address this, DetailCLIP integrates self-curated learning objectives that iteratively improve both high-level semantics and detailed visual representations. Specifically, our method employs patch-level self-distillation and pixel-level reconstruction losses to generate refined internal representations, while an attention-based token filtering mechanism curates semantically relevant information during training. By generating and refining self-curated learning signals, DetailCLIP improves segmentation performance and demonstrates superior generalization across diverse tasks. These task-agnostic objectives position DetailCLIP as a self-improving foundation model, enhancing multi-modal systems like CLIP with fine-grained feature understanding. \textcolor{red}{\href{https://github.com/KishoreP1/DetailCLIP}{https://github.com/KishoreP1/DetailCLIP}}
\end{abstract}

\section{Introduction}

The rapid advancements in computer vision have led to the development of foundation models that can understand visual data with impressive accuracy. Among these, CLIP (Contrastive Language-Image Pretraining)~\citep{radford2021learning} stands out as a pioneering approach that leverages large-scale contrastive learning between images and text to create a shared embedding space. This has proven immensely successful in tasks like classification, where the model can understand and relate visual content to textual descriptions without direct supervision~\citep{yu2024convolutions, wei2023iclip, lan2024proxyclip}.

However, while CLIP excels in generalization and high-level semantic understanding, its reliance on contrastive losses presents significant challenges when adapting the model for more granular tasks like image segmentation~\citep{wei2023iclip, zhou2023zegclip, lin2023clip, yang2023attentive}. Fine-grained  
tasks like segmentation requires a comprehensive understanding of the entire scene and the precise delineation of object boundaries at the pixel level. However, the global representations learned through contrastive learning often fail to capture this level of detail, leading to suboptimal performance when using CLIP for granular tasks~\citep{jiao2023learning, cha2023learning, zhou2023non, karimi2023crashformer, perera2024segformer3d, wang2022cris}.

Furthermore, traditional self-supervised learning (SSL) approaches~\citep{zhou2021ibot, caron2021emerging, kakogeorgiou2022hide, zhou2024reducing, navard2024probabilistic, farahani2023smart}, which have gained popularity for their ability to learn from unlabeled data, often prove inadequate when applied to fine-grained tasks. These methods, including contrastive learning~\citep{oord2018representation, chen2020simple, tian2020makes, he2020momentum} or clustering-based techniques~\citep{monsefi2024masked}, primarily focus on learning representations that are useful for high-level tasks such as classification or detection. However, they often fail to capture the intricate details and spatial relationships necessary for precise boundary delineation and detailed feature extraction. This limitation is particularly evident in masked image modeling (MIM) approaches such as MAE~\citep{he2022masked}, SimMIM~\citep{xie2022simmim}, and iBOT~\citep{zhou2021ibot}, which, while effective at learning global features, struggle to preserve fine-grained details. Consequently, these models often require extensive training epochs to achieve moderate performance improvements, yet still fall short in tasks that demand a high degree of granularity. The lack of textual context in these SSL models further exacerbates their limitations, as they miss the additional semantic guidance that text provides, which is crucial for enhancing the model's ability to focus on detail-oriented tasks. Incorporating text in the pre-training process can significantly improve the model's performance by aligning visual features with rich semantic information, leading to better generalization and more accurate granular tasks~\citep{bao2021beit, yi2022masked, zakeri2019weakly}.


This underscores the need for self-curated learning mechanisms that provide stronger, task-agnostic signals, enabling models to refine both high-level and fine-grained representations. To address this, we introduce DetailCLIP, a self-improving framework that overcomes the limitations of traditional CLIP-based models and self-supervised approaches for detail-oriented tasks. By enhancing both semantic understanding and fine-grained feature extraction, DetailCLIP achieves superior performance in tasks requiring meticulous attention to detail.


DetailCLIP introduces an attention-based token removal mechanism that selectively retains the most relevant image regions, guided by task-specific and textual information. As shown in Figure~\ref{fig:token:removal}, this mechanism focuses the model on critical areas, enhancing its ability to capture and represent subtle details accurately in the output.

DetailCLIP improves pixel-level accuracy through a pixel-level reconstruction feature that preserves detailed image features, enabling it to handle fine-grained tasks requiring precision. Unlike models like MAE~\citep{he2022masked} and SimMIM~\citep{xie2022simmim}, which use random token removal, DetailCLIP employs an attention-based token removal mechanism to retain only the most relevant features. This targeted approach enhances performance on tasks demanding high pixel-level accuracy, setting DetailCLIP apart from previous methods.


DetailCLIP incorporates a self-distillation strategy~\citep{caron2021emerging}, where the student model reconstructs masked images under the guidance of a teacher model using unmasked images. This iterative process refines intricate features, enhancing both segmentation accuracy and performance on classification tasks by preserving semantic integrity and detailed feature recognition. Unlike previous models such as SLIP~\citep{mu2022slip}, MaskCLIP~\citep{dong2023maskclip}, and A-CLIP~\citep{yang2023attentive}, which focus on coarse-grained comparisons, DetailCLIP combines both coarse-grained and fine-grained self-distillation, enabling superior performance in tasks requiring high precision.


Finally, similar to CLIP-based models~\citep{radford2021learning, luo2023segclip}, DetailCLIP emphasizes the crucial connection between vision and language, which is essential for tasks that require a deep understanding of visual content and its associated textual concepts. 

To summarize, our contributions are fourfold:
\begin{itemize}

    \item A \textbf{self-improving token filtering mechanism} driven by attention, which selectively retains tokens with strong semantic relevance to both textual descriptions and fine-grained tasks.
    
    \item A \textbf{patch-level self-distillation strategy}, where the model learns from masked images by comparing internal representations.
    
    \item A \textbf{pixel-level reconstruction} of masked images to provide a stronger self-supervised learning signal, enabling accurate recovery of intricate details for precision tasks.
    
    \item Integration of these objectives makes DetailCLIP an effective \textbf{self-improving vision-language model}, excelling in zero-shot and fine-tuning for detail-oriented tasks.
 
 \end{itemize}

\section{Related Work}

\subsection{Self Supervised Learning}

Self-Supervised Learning (SSL) has emerged as a dominant force in visual pre-training, with various strategies driving its success across different pretext tasks~\citep{ben2024reverse, su2024flsl, almalki2024self, oord2018representation, chen2020simple, tian2020makes, he2020momentum, bao2021beit, he2022masked, xie2022simmim, monsefi2024masked}. These approaches enable SSL to extract meaningful visual features from large-scale unlabeled image datasets~\citep{oquab2023dinov2, xie2022masked, liu2023frequency, zheng2024mfae, kakogeorgiou2022hide, zhou2021ibot, chen2020big, chen2021exploring, caron2021emerging}. Among the key methodologies are contrastive learning, Masked Image Modeling (MIM), Masked Frequency Modeling (MFM), and self-supervised Knowledge Distillation (KD), each designed to optimize representation learning in different ways.

Contrastive learning, a popular SSL technique, aligns different views of the same image within a shared embedding space while distinguishing them from views of other images~\citep{oord2018representation, chen2020simple}. Although this approach is effective for learning global features, it often faces challenges in handling tasks that require detailed precision, such as image segmentation. The reason is that contrastive learning primarily focuses on aligning overall image representations, which can lead to the neglect of fine-grained, pixel-level details.

On the other hand, Masked Image Modeling (MIM) has shown greater promise for tasks requiring detailed feature extraction. MIM involves masking portions of an image and training models to reconstruct or predict the missing parts, encouraging the model to understand the intricate relationships between visible and hidden regions~\citep{bao2021beit, xie2022simmim, he2022masked, yi2022masked}.

\subsection{CLIP-Based Approaches}

The integration of self-supervised learning with vision-language pre-training has driven the development of several CLIP-based models, each striving to bridge the gap between visual and textual data~\citep{cha2023learning, zhou2023non, li2023scaling, jia2021scaling}. These models build on the original CLIP framework, introducing innovations to enhance the quality of representations and improve alignment between images and text.

For instance, SLIP~\citep{mu2022slip} enhances CLIP by incorporating self-supervised learning with image-to-image contrastive learning, resulting in richer and more robust visual representations. MaskCLIP~\citep{dong2023maskclip} advances this concept by integrating masked image modeling, which refines visual features by focusing on specific image regions, ensuring they are more closely aligned with the accompanying text. A-CLIP~\citep{yang2023attentive} further refines this approach by implementing an attentive token removal strategy, selectively preserving tokens that are semantically relevant to the text, thereby enhancing the precision of the visual-text alignment.

While these methods enhance the original CLIP model by improving representation learning and efficiency, they primarily focus on global alignment between images and text. This emphasis on global features makes them less effective for detail-oriented tasks, where capturing and preserving fine-grained visual details is crucial~\citep{jiao2023learning, wei2023iclip, he2023clip, zhou2023zegclip}. In many of these models, token removal is often random or solely based on textual information~\citep{dong2023maskclip, yang2023attentive}. In contrast, DetailCLIP addresses these limitations by employing an attention-based mechanism that incorporates both textual information and fine-grained tasks, ensuring better performance in detail-sensitive applications.

\section{Method}

\begin{figure*}[ht]
    \centering
    \includegraphics[width=0.7\textwidth]{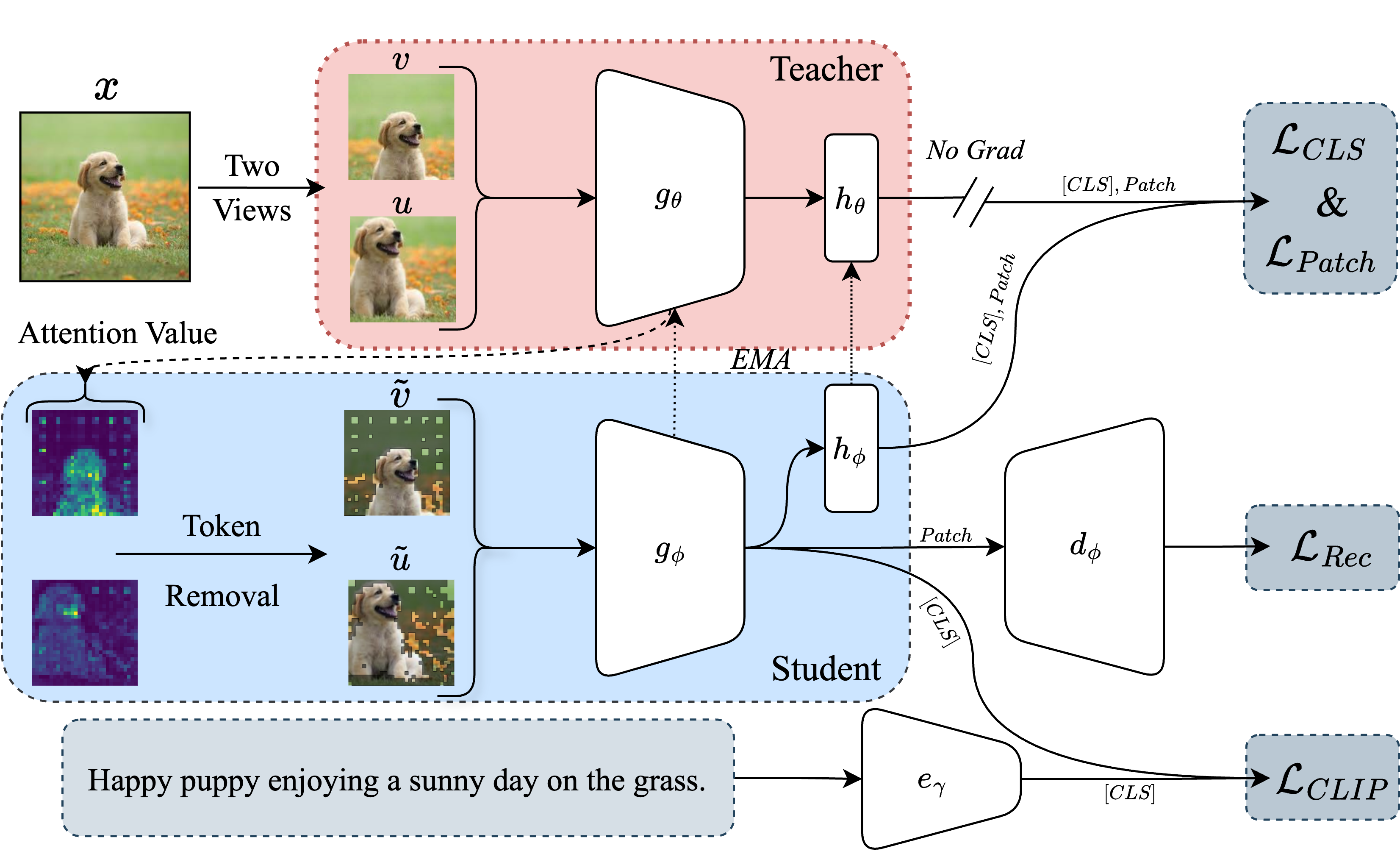}
    \caption{Using a teacher-student architecture, the teacher processes two views of an input image to generate attention values, guiding token removal in the student model. The student reconstructs the image with a vision decoder while optimizing classification loss ($\mathcal{L}_{CLS}$), patch contrastive loss ($\mathcal{L}_{Patch}$), and reconstruction loss ($\mathcal{L}_{Rec}$). The CLIP loss ($\mathcal{L}_{CLIP}$) ensures alignment between vision and text encoders, enhancing segmentation. In the figure, $g$ is the vision encoder, $h$ the projection head, $d$ the decoder, and $e$ the text encoder.}
    \label{fig:DetailCLIP:framework}
\end{figure*}

\subsection{Preliminary and Background}
\label{sec:prelim:back}
CLIP, developed by \citet{radford2021learning}, is a vision-language model that aligns visual and textual representations through large-scale contrastive learning. It trains an image and text encoder jointly to maximize similarity between matching image-text pairs and minimize it for mismatched ones. The training objective uses a symmetric cross-entropy loss on the cosine similarity between image and text embeddings, encouraging correct pairs to have higher similarity than incorrect pairs. Let $I_i$ and $T_j$ represent the embeddings of the $i$-th image and $j$-th text in a batch. The cosine similarity between them is defined as:
\begin{equation}
s_{i,j} = \frac{I_i \cdot T_j}{\|I_i\| \|T_j\|},
\end{equation}

\noindent The logits are scaled by a learnable parameter $\tau$:

\begin{equation}
\text{logits}_{i,j} = \frac{s_{i,j}}{\tau},
\end{equation}

\noindent Let $N$ be the number of pairs of image-text in a batch. The loss function for the image-to-text matching is defined as:

\begin{equation}
\mathcal{L}_{I2T} = -\frac{1}{N} \sum_{i=1}^{N} \log \left(\frac{\exp(\text{logits}_{i,i})}{\sum_{j=1}^{N} \exp(\text{logits}_{i,j})}\right),
\end{equation}

\noindent Similarly, the loss function for the text-to-image matching is:

\begin{equation}
\mathcal{L}_{T2I} = -\frac{1}{N} \sum_{i=1}^{N} \log \left(\frac{\exp(\text{logits}_{i,i})}{\sum_{j=1}^{N} \exp(\text{logits}_{j,i})}\right),
\end{equation}

\noindent Finally, the overall loss for CLIP is the average of the two losses:

\begin{equation}
\mathcal{L}_{CLIP}=\dfrac{1}{2}(\mathcal{L}_{I2T}+\mathcal{L}_{T2I}),
\label{eq:loss_clip}
\end{equation}

\subsection{DetailCLIP Framework}
As shown in Figure~\ref{fig:DetailCLIP:framework}, our framework can be outlined through the following steps, which will be discussed in greater detail. Additionally, Section~\ref{sec:pseudocode} of the appendix provides a pytorch-like pseudocode of the forward pass.

\subsubsection{Patch-Level Comparison}

We employ a teacher-student framework in which the student model is trained to predict both fine-grained and coarse-grained features of images generated by the teacher model. The teacher model provides "target" features for the student to learn. For each input, we consider two views: the original views are fed into the teacher model, while their masked versions are used as input for the student model. For further details on the masking process, refer to Section~\ref{sec:token_removal}. The student model is tasked with predicting both the masked tokens and the global features generated by the teacher.

Similar to methods used in previous works like DINO~\citep{caron2021emerging}, the teacher model's parameters are updated using an exponential moving average (EMA) of the student model's parameters and its own. Specifically, if $\theta$ represents a parameter of the teacher model and $\phi$ represents a parameter of the student model, the update rule for $\theta$ is $\theta \leftarrow (1 - \lambda)\phi + \lambda\theta$, that $\lambda$ starting at 0.996 and gradually increasing to 1.0, following a cosine scheduling approach as outlined in previous works~\citep{chen2021empirical, dong2023maskclip, yang2023attentive}. This strategy ensures that the EMA model remains stable and effective in capturing the most relevant features during training~\citep{grill2020bootstrap}.

We employ two primary loss functions to train the student model to ensure both global and fine-grained feature learning are effectively captured.

\noindent \textbf{Global Loss (CLS Token)}:
The global loss, applied to the [CLS] token, measures the difference between the probability distribution of the [CLS] token in the student model and the [CLS] token in the teacher model using KL divergence:

\begin{equation}
\mathcal{L}_{\text{CLS}} = \text{D}_{KL}\left(p_{\text{s}}^{\text{[CLS]}} \| p_{\text{t}}^{\text{[CLS]}}\right)
\end{equation}
where
$p_{\text{s}}^{\text{[CLS]}}$ is the predicted probability distribution of the [CLS] token from the student model,
$p_{\text{t}}^{\text{[CLS]}}$ is the target probability distribution of the [CLS] token from the teacher model,
$\text{D}_{KL}$ represents the Kullback-Leibler divergence~\citep{kullback1951information}.

\noindent \textbf{Fine-grained Loss (Patch)}:
The patch loss also utilizes KL divergence, comparing the distribution of patch tokens in the student model to those in the teacher model:
\begin{equation}
\mathcal{L}_{\text{patch}} = \frac{1}{|\mathcal{M}|} \sum_{j \in \mathcal{M}} \text{D}_{KL}\left(p_{\text{s}}^j \| p_{\text{t}}^j\right)
\end{equation}
where $\mathcal{M}$ is the set of indices corresponding to the masked patches,
$p_{\text{s}}^j$ is the predicted probability distribution of the $j$-th masked patch from the student model,
$p_{\text{t}}^j$ is the target probability distribution of the $j$-th masked patch from the teacher model,
$\text{D}_{KL}$ represents the Kullback-Leibler divergence.

\subsubsection{Pixel-Level Reconstruction}

As illustrated in Figure~\ref{fig:DetailCLIP:framework}, following token removal, we employ an autoencoding method ($g_{\phi}$-$d_{\phi}$) (MAE~\citep{he2022masked}) to reconstruct the original signal from a partially observed input. By processing only visible patches, the encoder reduces data handling and improves efficiency. The decoder reconstructs the original image using the encoded visible patches and mask tokens representing the masked areas. Since the decoder is smaller and used only during pre-training, most computational effort is focused on training a more powerful encoder.

This process is especially well-suited for fine-grained tasks because it pushes the encoder to extract detailed information from the limited visible patches, sharpening its ability to recognize intricate features. By concentrating on these visible patches, the encoder becomes skilled at capturing subtle details that are critical in precision-demanding tasks like segmentation or detailed object recognition. The decoder's responsibility in reconstructing the masked portions of the image ensures that these fine details are preserved and accurately restored. The loss function ($\mathcal{L}_{\text{Rec}}$) employed during this training is the Mean Squared Error (MSE) between the reconstructed and original images, calculated solely on the masked patches:

\begin{equation}
\mathcal{L}_{\text{Rec}} = \frac{1}{|\mathcal{M}|} \sum_{i \in \mathcal{M}} \| \hat{X}_i - X_i \|^2
\end{equation}
where 
    $\mathcal{M}$ is the indices corresponding to the masked patches,
    $\hat{X}_i$ is the reconstructed patch for the $i$-th masked patch,
    $X_i$ is the original patch for the $i$-th masked patch,
    $\|\cdot\|$ represents the Euclidean norm (or $\ell_2$ norm).

\subsubsection{Token Removal}
\label{sec:token_removal}

Token removal in ViT refines the attention mechanism by eliminating less relevant tokens, reducing computational load, prioritizing important patches, and improving downstream performance. Existing methods include random token removal~\citep{kakogeorgiou2022hide} and attentive removal~\citep{yang2023attentive}. Our approach, tailored to the model’s tasks—textual processing, patch comparison, and image reconstruction—outperforms these methods by effectively handling both fine-grained and coarse-grained details (see Section~\ref{sec:attn_removal} for more detail).

As shown in Figure~\ref{fig:DetailCLIP:framework}, our token removal process begins by passing an image view through the teacher encoder ($g_\theta$). The attention values generated by the encoder are used to mask the $50\%$ of patches with the lowest values. For each patch $p_{\text{t}}$, the attention values are computed as:

\begin{equation}
   AV_{p_{\text{t}}} = \frac{1}{N} \sum_{i=1}^{N} \text{Softmax}\left(\frac{Q_i \cdot K_i(p_{\text{t}})}{\sqrt{d}}\right)
\end{equation}
where
    $N$ is the total number of attention heads across all layers,
    $Q_i$ is the query of the [CLS] from the $i$-th attention head,
    $K_i(p_{t})$ is the key of the patch at position $p_{\text{t}}$ from the $i$-th attention head,
    $d$ is the dimensionality of the query and key.

As shown in Figure~\ref{fig:token:removal} of the appendix, masking image patches by removing the lowest half of the attention values (AV) calculated by the teacher encoder results in the images displayed in the figure.

\subsubsection{Integrated Loss Function}

We introduce the formulation of a comprehensive loss function that combines the various loss components essential for effective multi-task learning. The integrated loss function is defined as:

\begin{equation}
\mathcal{L}_{\text{tot}} = 
\alpha_1 \times \mathcal{L}_{\text{CLS}} + 
\alpha_2 \times \mathcal{L}_{\text{Patch}} + 
\alpha_3 \times \mathcal{L}_{\text{Rec}} + 
\mathcal{L}_{\text{CLIP}},
\label{eq:loss_total}
\end{equation}

\noindent In this equation, the hyperparameters $\alpha_1$, $\alpha_2$, and $\alpha_3$ determine the relative importance of each loss term. These weights are set to $1$ in our experiments unless specified otherwise. The comprehensive loss function is designed to enable the model to concurrently learn across the four tasks introduced in previous sections, ensuring a balanced and integrated training process. In Section~\ref{sec:ablation_weight} of the appendix, we explore the impact of different hyper-parameter choices, demonstrating how varying these weights can influence the overall model performance.

\section{Experiments}

\subsection{Setup}

Our experiments were conducted on a setup with four nodes, each equipped with four NVIDIA A100 GPUs (80GB). For details on computational efficiency, see Section~\ref{sec:compute} in the appendix.

\subsubsection{Training Data and Augmentation Strategy}

Our model is trained on a carefully curated 15 million image subset of the YFCC100M dataset~\citep{thomee2016yfcc100m}, which contains only English-language titles and descriptions. For each image, a valid caption—either a title or description—is randomly selected during training, following the methodology used in SLIP~\citep{mu2022slip}.

To enhance model robustness, we employ a data augmentation strategy similar to that of SLIP. Images are randomly resized and cropped to a size between $50\%$ and $100\%$ of their original dimensions. This augmentation is applied to the images in the online training branches, allowing the model to learn from varied perspectives of the same image. For the teacher branch, we use a larger randomly cropped sub-image compared to the online views, which allows us to accurately compute attention value.

\subsubsection{Architecture and Training Setting}

Figure \ref{fig:DetailCLIP:framework} illustrates the architecture of the DetailCLIP framework, which is composed of six key components: two visual encoders, denoted as $g_{\theta}$, and $g_{\phi}$, a text encoder named $e_{\gamma}$, a vision decoder named $d_{\phi}$, and two heads referred to as $h_{\theta}$ and $h_{\phi}$. 

To ensure a fair comparison with existing models, our visual encoders ($g_{\theta}$ and $g_{\phi}$) are based on the Vision Transformer (ViT-B/16)~\citep{dosovitskiy2020image} architecture. This model features $12$ layers, each with a width of $768$ and $12$ attention heads.

For the text encoder ($e_{\gamma}$), we adopted a 12-layer Transformer architecture with a width of $512$ and $8$ attention heads, following the design principles established by CLIP~\citep{radford2021learning}.

The decoder ($d_{\phi}$) itself is composed of a series of transformer blocks like MAE~\cite{he2022masked}. The decoder handles the entire set of input patches, including both the attention-masked and retained patches. Positional embeddings are added to each token in the set to preserve the spatial information.

The projection heads ($h_{\theta}$ and $h_{\phi}$) in our framework are implemented as 3-layer MLPs, featuring an L2-normalized bottleneck, similar to the approach used in DINO~\citep{caron2021emerging}. The output dimension of the shared projection head is set to $8192$ like iBOT paper~\citep{zhou2021ibot}, for robust feature representation and alignment during training.


For our experiments, we employ the $AdamW$ optimizer~\citep{loshchilov2017decoupled} with a learning rate of $5e-4$ and a weight decay of $0.5$. The training is conducted with a substantial batch size of $4096$.


\subsection{Experimental Analysis} 
\label{seq:eval}

Additional experiments and analyses are provided in the appendix due to page limitations: Section~\ref{sec:attn_removal} (attention-based token removal), Section~\ref{sec:compute} (compute efficiency), Section~\ref{sec:pseudocode} (pseudocode), Section~\ref{sec:ablation_weight} (ablation studies), and Section~\ref{sec:visual_analysis} (visual comparison of model outputs).

\subsubsection{Detail-Oriented Visual Tasks}
\label{seq:detail:oriented:eval}

\begin{table*}[ht]
\small
\centering
\begin{tabular}{cccccccc}
\hline
\multicolumn{1}{c|}{}                                            &                                    &                                  & \multicolumn{1}{c|}{}                                          & \multicolumn{2}{c|}{\textbf{ADE20K}}                                               & \multicolumn{2}{c}{\textbf{COCO}}                               \\ \cline{5-8} 
\multicolumn{1}{c|}{\multirow{-2}{*}{\textbf{Methods}}}          & \multirow{-2}{*}{\textbf{Dataset}} & \multirow{-2}{*}{\textbf{Epoch}} & \multicolumn{1}{c|}{\multirow{-2}{*}{\textbf{Effective View}}} & \textbf{UperNet} & \multicolumn{1}{c|}{\textbf{Linear}}       & \textbf{$AP^b$} & \textbf{$AP^m$} \\ \hline
\rowcolor[HTML]{EFEFEF} 
\multicolumn{8}{l}{\cellcolor[HTML]{EFEFEF}\textit{\textbf{Self-Supervised}}}                                                                                                                                                                                                                                                                                    \\ \hline
\multicolumn{1}{c|}{DeiT}                                        & IN-1K                              & 300                              & \multicolumn{1}{c|}{720M}                                      & 47.4                      & \multicolumn{1}{c|}{-}                        & 44.1                           & 39.8                           \\
\multicolumn{1}{c|}{MAE}                                         & IN-1K                              & 800                              & \multicolumn{1}{c|}{960M}                                      & 46.5                      & \multicolumn{1}{c|}{34.3}                        & 46.2                           & 39.1                           \\
\multicolumn{1}{c|}{DINO}                                        & IN-1K                              & 800                              & \multicolumn{1}{c|}{9600M}                                     & 46.8                      & \multicolumn{1}{c|}{34.5}                        & 47.4                           & 40.1                           \\
\multicolumn{1}{c|}{iBOT}                                        & IN-1K                              & 300                              & \multicolumn{1}{c|}{720M}                                      & 47.3                     & \multicolumn{1}{c|}{34.7}                        & 48.4                           & 42.1                           \\
\multicolumn{1}{c|}{AttMask}                                     & IN-1K                              & 300                              & \multicolumn{1}{c|}{4320M}                                     & 47.5                      & \multicolumn{1}{c|}{35.2}                        & 48.9                           & 42.2                           \\ \hline
\rowcolor[HTML]{EFEFEF} 
\multicolumn{8}{l}{\cellcolor[HTML]{EFEFEF}\textit{\textbf{CLIP-Based Model}}}                                                                                                                                                                                                                                                                                   \\ \hline
\multicolumn{1}{c|}{CLIP}                                        & 400M                               & -                                & \multicolumn{1}{c|}{-}                                         & 46.4                      & \multicolumn{1}{c|}{34.2}                      & 43.6                           & 39.5                           \\
\multicolumn{1}{c|}{SLIP}                                        & YFCC-15M                           & 25                               & \multicolumn{1}{c|}{750M}                                      & 46.6                        & \multicolumn{1}{c|}{36.1}                         & 44.0                           & 40.3                           \\
\multicolumn{1}{c|}{MaskCLIP}                                    & YFCC-15M                           & 25                               & \multicolumn{1}{c|}{750M}                                      & 47.5                        & \multicolumn{1}{c|}{36.3}                     & 45.4                           & 40.9                           \\
\multicolumn{1}{c|}{A-CLIP}                                       & YFCC-15M                           & 25                               & \multicolumn{1}{c|}{750M}                                      & 47.0                   & \multicolumn{1}{c|}{34.7}                         & 45.8                           & 41.7                           \\ \hline
\rowcolor[HTML]{ECF4FF} 
\multicolumn{1}{c|}{\cellcolor[HTML]{ECF4FF}\textbf{DetailCLIP}} & YFCC-15M                           & 25                               & \multicolumn{1}{c|}{\cellcolor[HTML]{ECF4FF}750M}              & \textbf{48.1}                         & \multicolumn{1}{c|}{\cellcolor[HTML]{ECF4FF }\textbf{37.3}} & \textbf{48.9}                           & \textbf{42.5}                           \\
\rowcolor[HTML]{ECF4FF} 
\multicolumn{1}{c|}{\cellcolor[HTML]{ECF4FF}\textbf{DetailCLIP}} & YFCC-15M                           & 50                               & \multicolumn{1}{c|}{\cellcolor[HTML]{ECF4FF}1500M}             & \textbf{48.8}                       & \multicolumn{1}{c|}{\cellcolor[HTML]{ECF4FF} \textbf{39.3}} & \textbf{50.1}                           & \textbf{43.3}                           \\ \hline \hline
\end{tabular}
\caption{Performance comparison on segmentation (ADE20K) and object detection (MS COCO) tasks using the ViT-B architecture. Segmentation results are based on UperNet and linear decoders, while object detection uses bounding box Average Precision. “Effective View” indicates the model’s training exposure, calculated as views per image × epochs × dataset size. }
\label{tab:res:seg-object}
\end{table*}

We evaluated DetailCLIP on detail-oriented visual tasks to test its ability to capture fine-grained details in complex visual environments.

\noindent \textbf{Semantic Segmentation on ADE20K:} We evaluated DetailCLIP on the ADE20K dataset~\citep{zhou2017scene} using both UperNet~\citep{xiao2018unified} and linear decoders. UperNet assessed the model’s ability to capture detailed object boundaries, while the linear decoder tested performance under minimal architectural complexity. Input images had a resolution of $512 \times 512$ pixels, and training was conducted for $160k$ iterations. Performance was measured using the mean Intersection over Union (mIoU) metric.

\noindent \textbf{Object Detection and Instance Segmentation on COCO:} We evaluated DetailCLIP using the Cascade Mask R-CNN model~\citep{cai2019cascade, he2017mask}, following iBOT, to generate bounding boxes and instance masks on the COCO dataset~\citep{lin2014microsoft}. Performance was measured using Average Precision for bounding boxes ($AP^b$) and masks ($AP^m$).

\noindent \textbf{Evaluating Task Outcomes:} Table \ref{tab:res:seg-object} shows DetailCLIP’s strong performance across detail-oriented visual tasks. Using the ViT-B architecture, DetailCLIP achieved a $mIoU$ of $48.8$ on ADE20K with UperNet, surpassing MaskCLIP by $1.3$ points. In the linear decoder setting, DetailCLIP’s $mIoU$ of $39.3$ outperformed SLIP by $3.2$ points. In object detection on MS COCO, DetailCLIP achieved $48.9$ $AP^b$ for bounding boxes, surpassing ACLIP by $3.1$ points, and $42.5$ $AP^m$ for masks, $0.8$ points higher than the best alternative. These improvements highlight DetailCLIP’s effectiveness in capturing both high-level semantics and fine-grained details.

\begin{figure*}[]
\centering
\begin{tabular}{c|c|cccc}
\hline
\textbf{Image} & \textbf{Ground Truth} & \textbf{A-Clip} & \textbf{SLIP} & \textbf{DetailCLIP} \\ \hline

\includegraphics[width=0.135\textwidth]{image_files/semantic_segmentation/raw/ADE_val_00001005.jpg} & \includegraphics[width=0.135\textwidth]{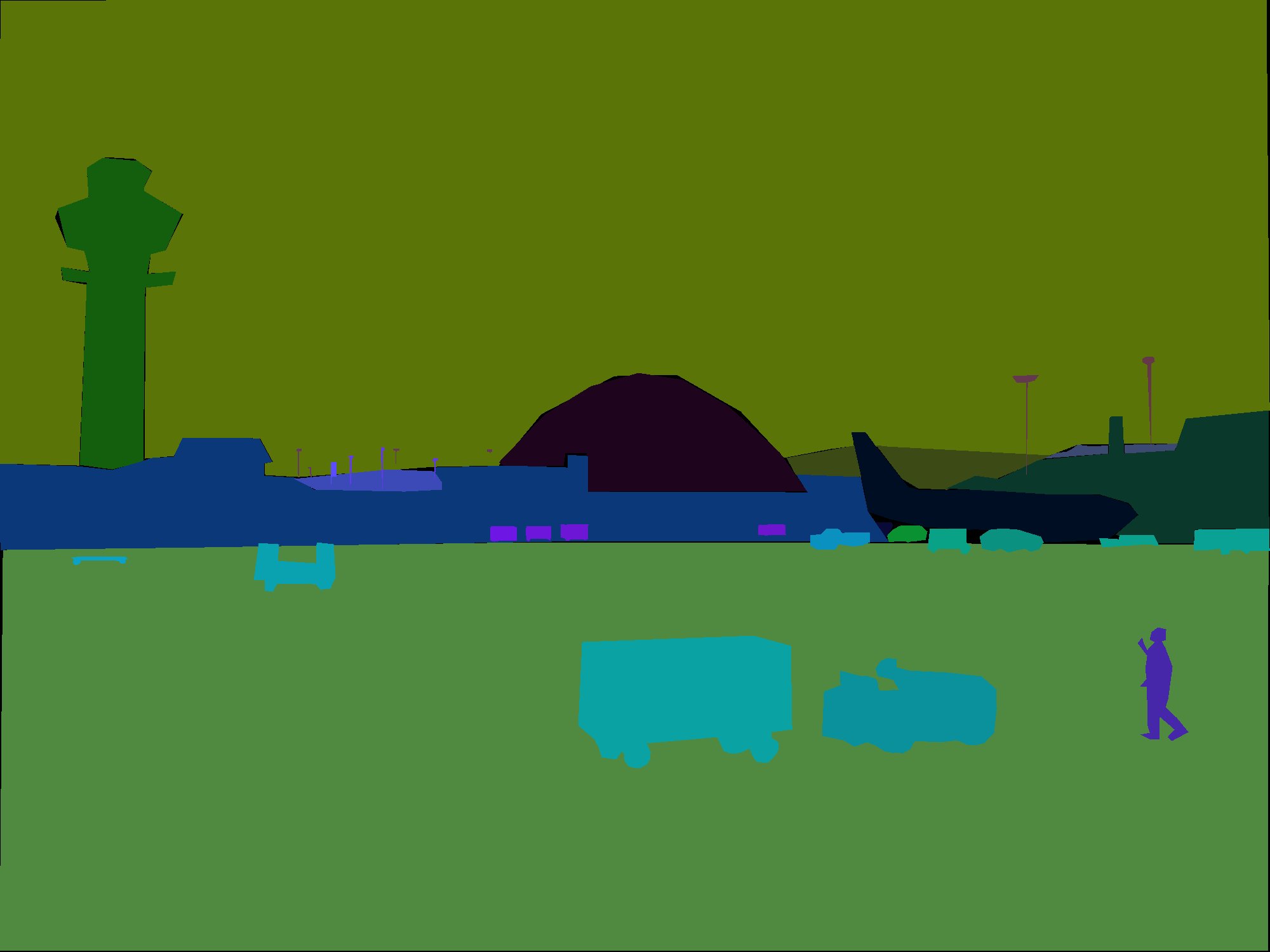} & \includegraphics[width=0.135\textwidth]{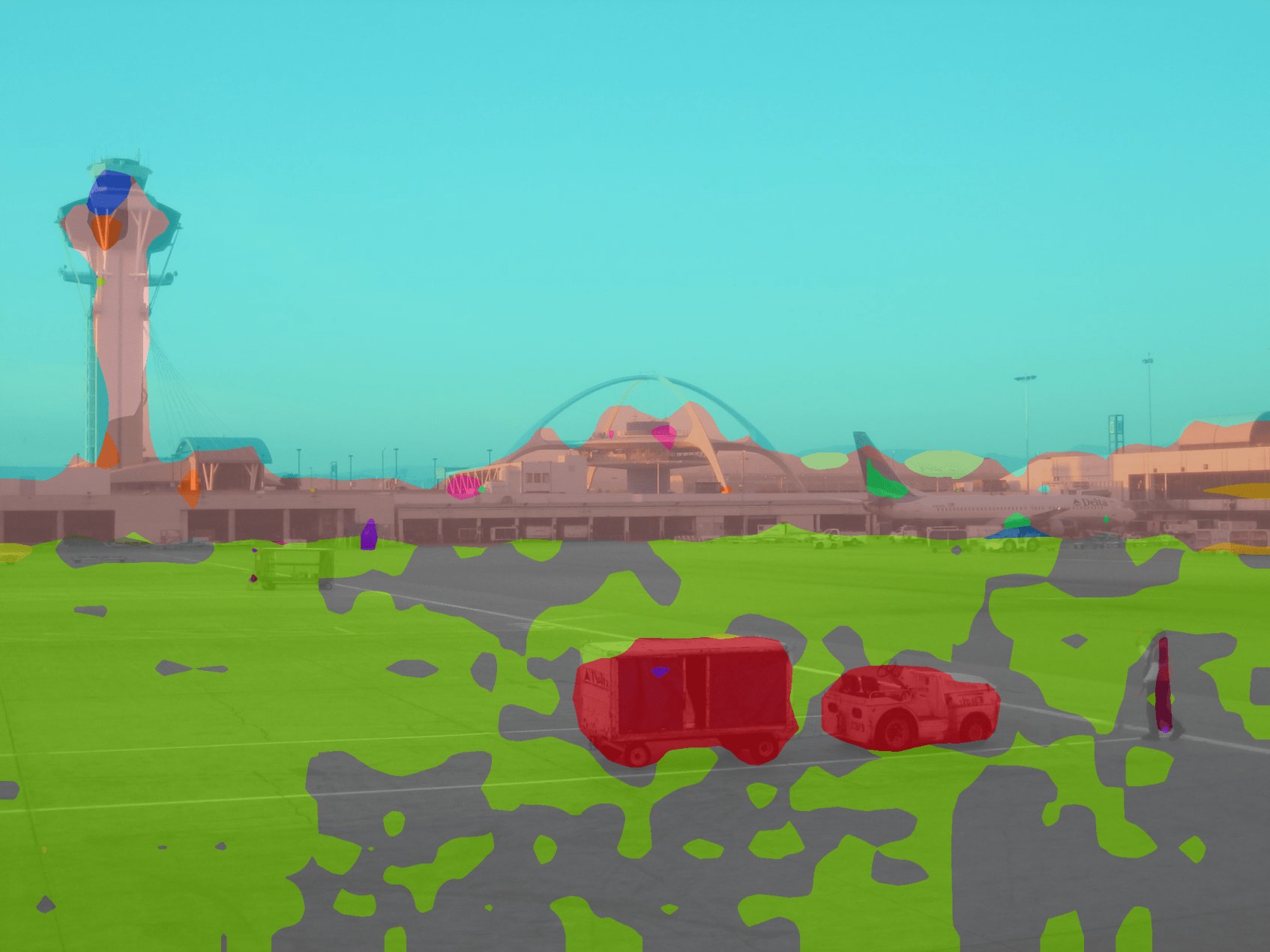} & \includegraphics[width=0.135\textwidth]{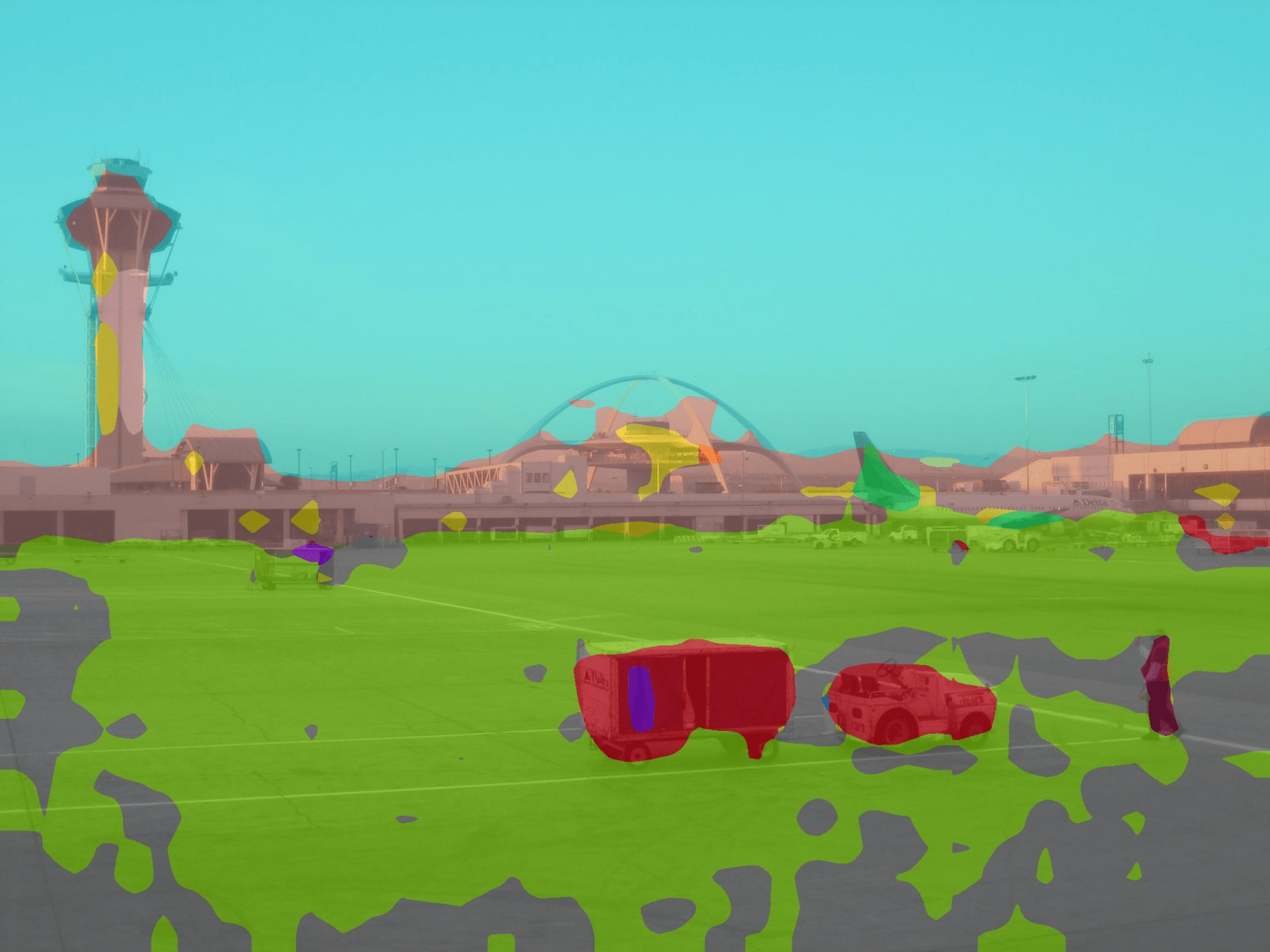}  & \includegraphics[width=0.135\textwidth]{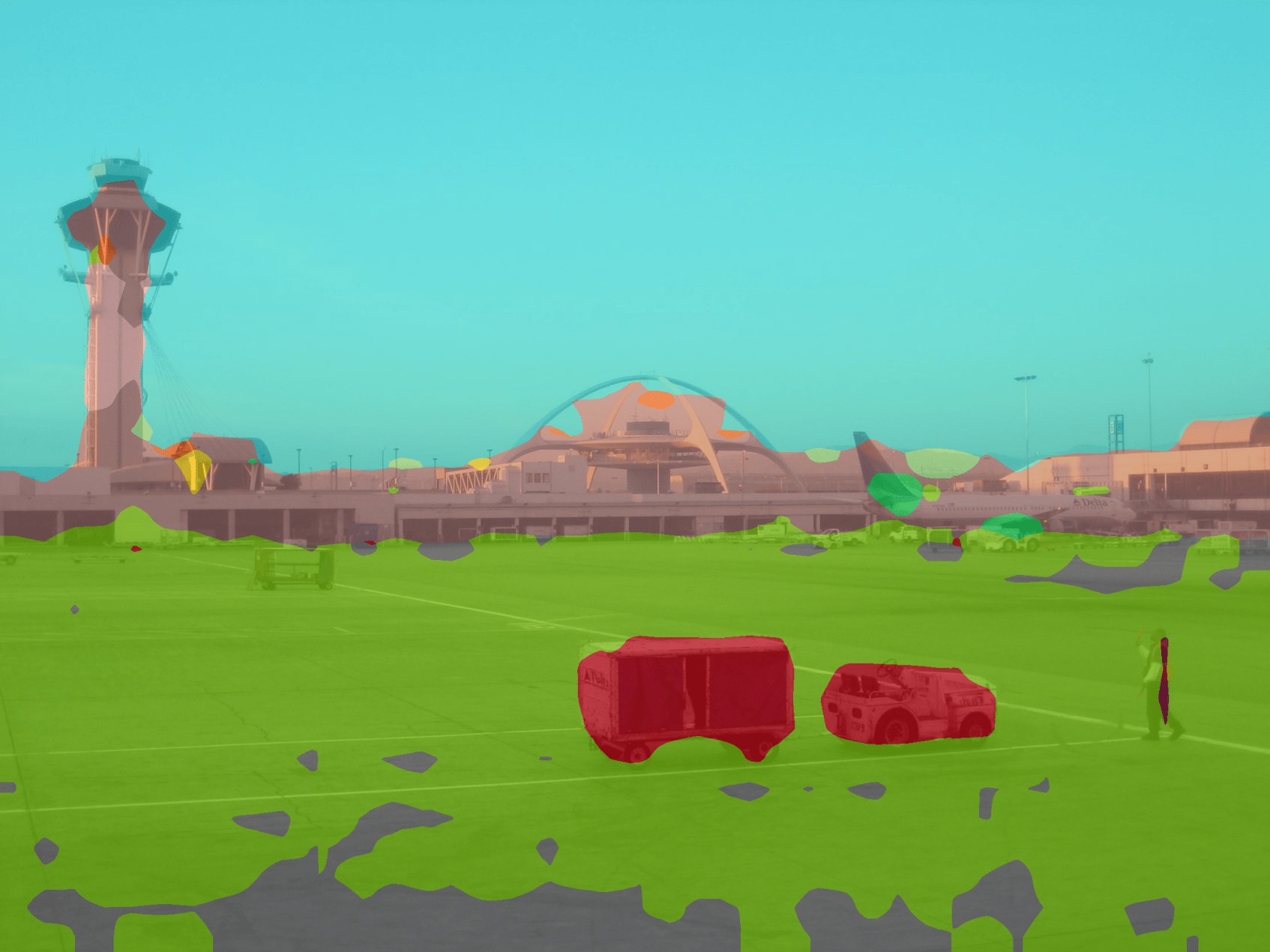} \\

\includegraphics[width=0.135\textwidth]{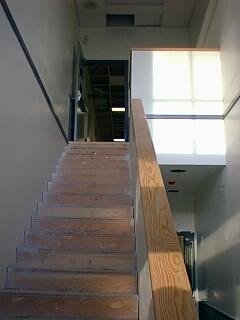} & \includegraphics[width=0.135\textwidth]{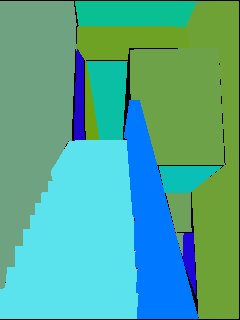} & \includegraphics[width=0.135\textwidth]{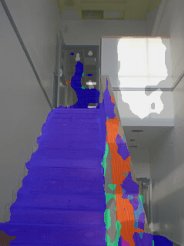} & \includegraphics[width=0.135\textwidth]{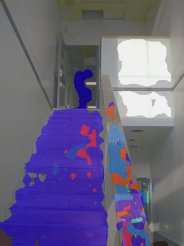}  & \includegraphics[width=0.135\textwidth]{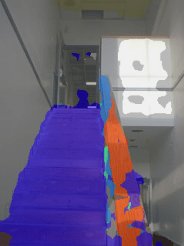}\\

\includegraphics[width=0.135\textwidth]{image_files/semantic_segmentation/raw/ADE_val_00001880.jpg} & \includegraphics[width=0.135\textwidth]{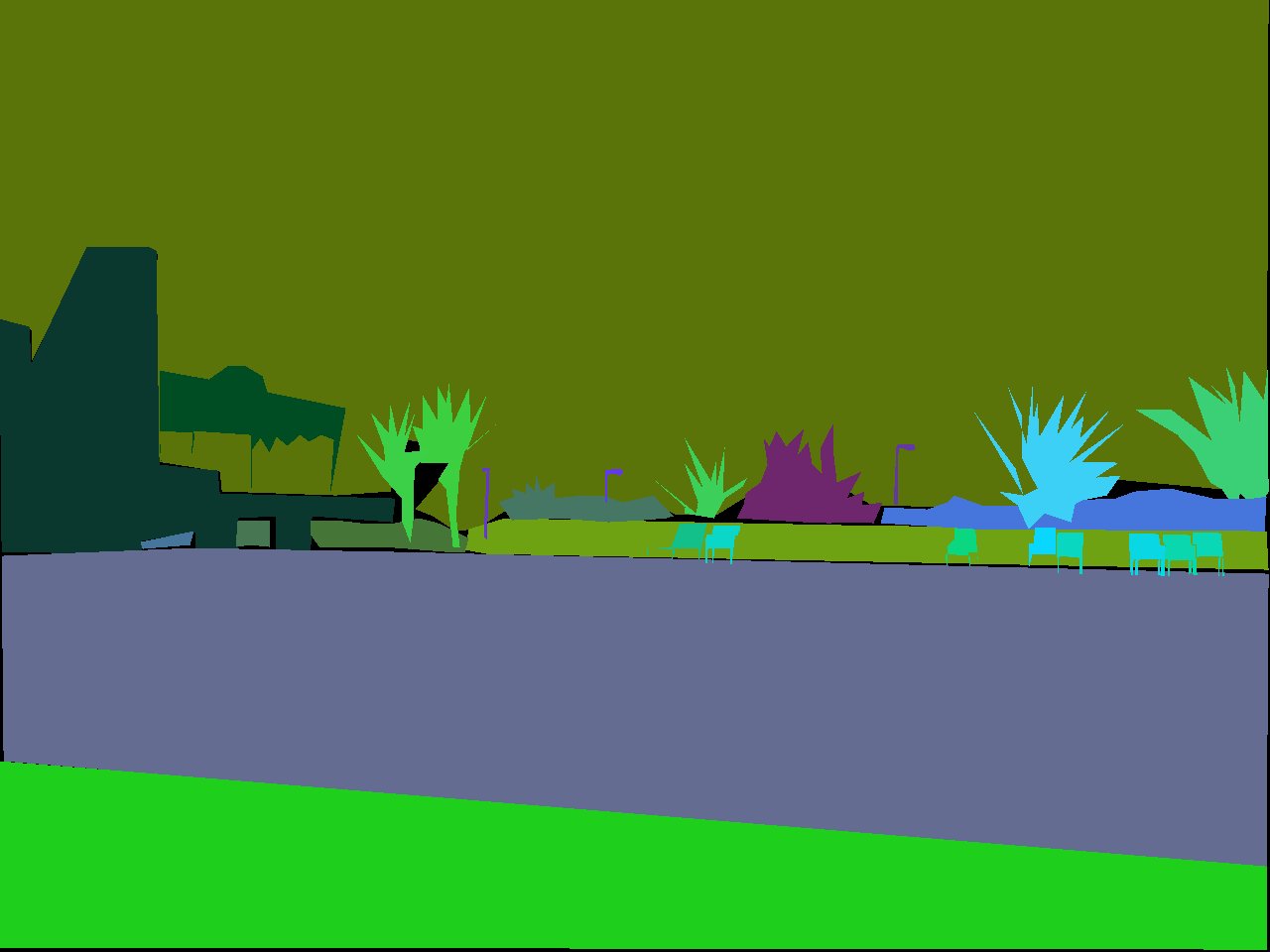} & \includegraphics[width=0.135\textwidth]{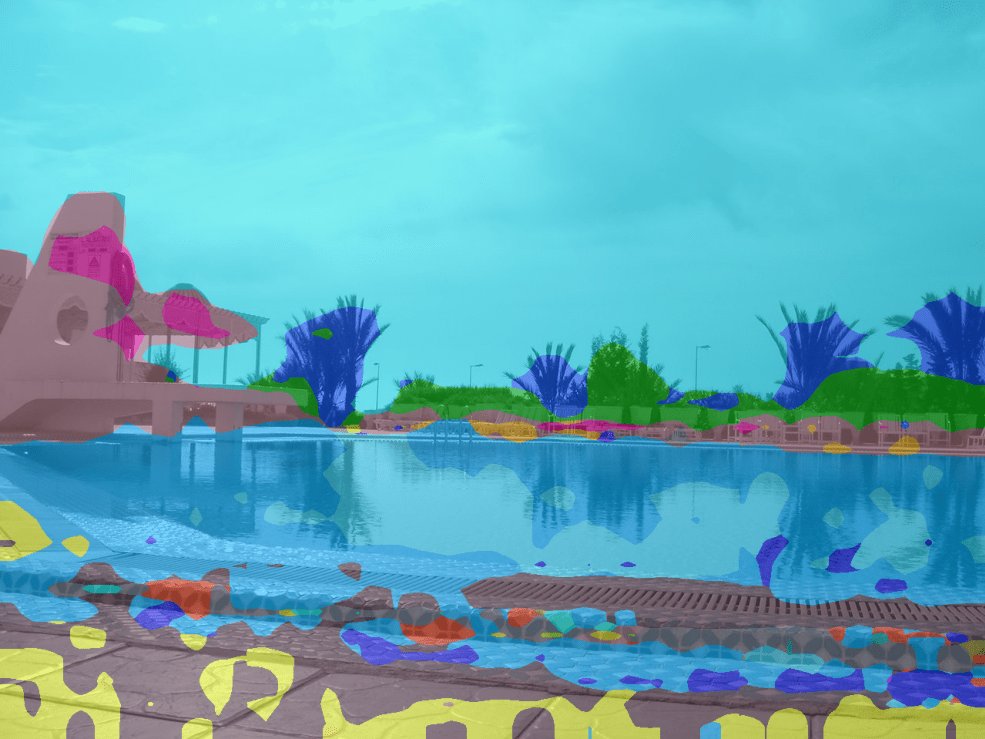} & \includegraphics[width=0.135\textwidth]{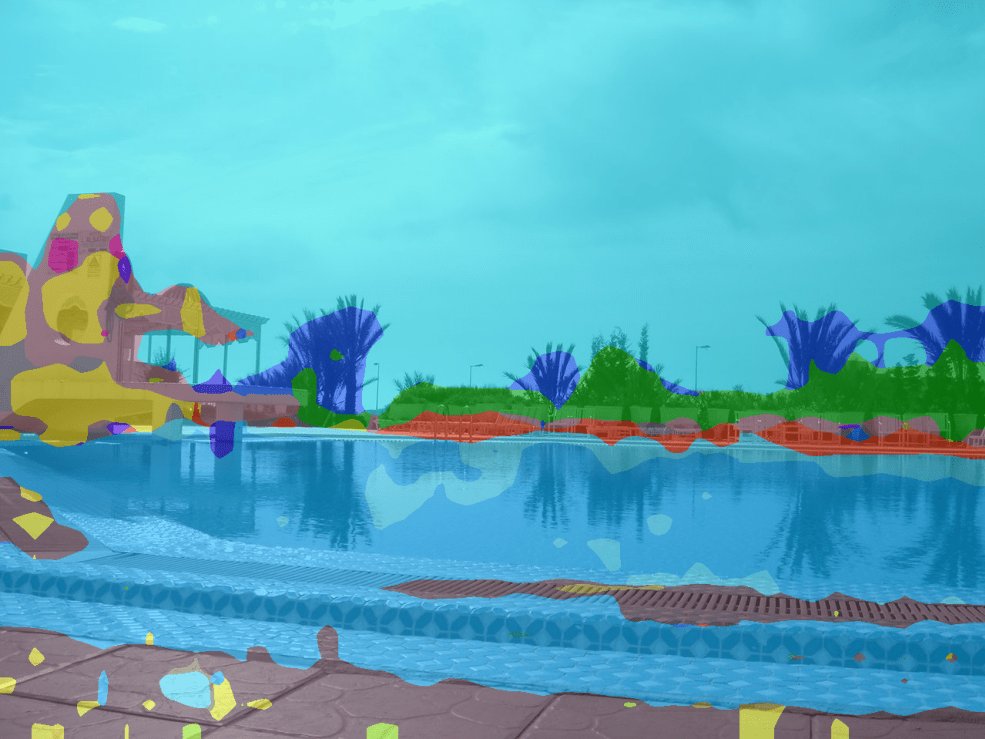}  & \includegraphics[width=0.135\textwidth]{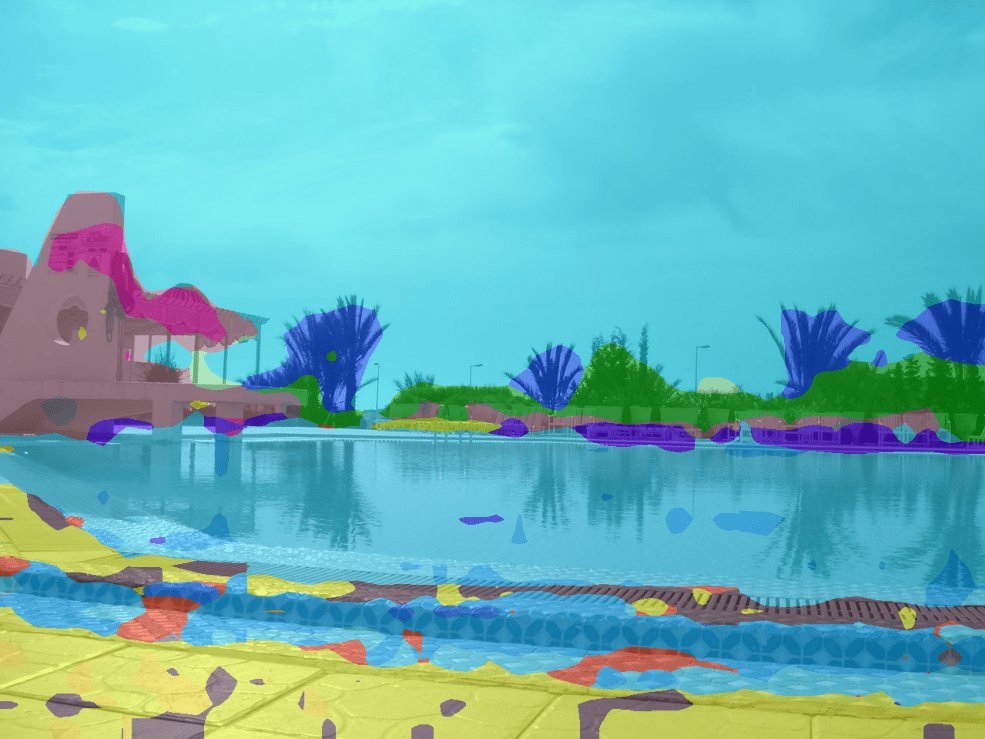}\\

\hline
\hline

\end{tabular}
\caption{ Visual comparison of segmentation results using the Linear decoder across different models. The comparison includes SLIP, Attentive Mask CLIP, and DetailCLIP at 25 epochs.}
\label{fig:seg:linear}
\end{figure*}

\noindent \textbf{Visual Comparative Analysis:} Figure \ref{fig:seg:linear} shows the result of our model compared to the other baselines. Additionally, we invite you to refer to Section~\ref{sec:visual_analysis} of the appendix, where we cover segmentation and object detection tasks.

\subsubsection{Image Classification}

\noindent \textbf{Zero-shot on Text-Image Retrieval:}
We also present the zero-shot text-image retrieval on 3 benchmark datasets: Flickr30K~\citep{young2014image}, MS-COCO~\citep{lin2014microsoft}, and ImageNet-1K~\citep{deng2009imagenet}. Our findings indicate that using plain text, without any added prefixes or suffixes, consistently yields strong performance across all models evaluated. Table \ref{tab:res:retreive} comprehensively evaluates different models on text-image retrieval tasks, explicitly focusing on Flickr30K and MS-COCO datasets, also a zero-shot evaluation on ImageNet-1K.

At 25 epochs, DetailCLIP demonstrates the highest performance across most metrics, achieving the best results in Flickr I2T (62.8), T2I (42.2), COCO I2T (38.3), and zero-shot on IN-1K (43.9), closely matching A-CLIP. For 50 epochs, DetailCLIP maintains its superior performance, particularly in the COCO dataset, where it ties in with A-CLIP for the best I2T score (39.8) and surpasses A-CLIP in T2I with a score of 24.9. The results suggest that DetailCLIP is particularly strong in tasks that require detailed matching between text and images. DetailCLIP outperforms other models in both datasets' I2T and T2I retrieval tasks, indicating its effectiveness in text-image association.

Additionally, in Section~\ref{sec:ablation_weight} of the appendix, we also provide the effect of different hyperparameter settings on ImageNet-1K's zero-shot scores.

\noindent \textbf{Zero-Shot Classification Performance Evaluation on Diverse Benchmarks:} We evaluated DetailCLIP’s zero-shot classification performance on 13 tasks, following SLIP’s~\citep{mu2022slip} evaluation protocols with identical prompts for fair comparison. The results in the table \ref{tab:res:classification} demonstrate that DetailCLIP consistently outperforms its competitors. Specifically, DetailCLIP achieved the highest performance in $12$ out of $26$ evaluation scenarios and secured the second-best position in $5$ additional cases. DetailCLIP also achieved the highest average scores in both the 25-epoch ($37.3$) and 50-epoch ($37.6$) settings.

\begin{table}[h]
\centering
\small
\begin{tabular}{c|c|ccccc}
\hline
                          &                         & \multicolumn{2}{c}{Flickr30K}    & \multicolumn{2}{c}{COCO}      & IN-1K         \\ \cline{3-7} 
\multirow{-2}{*}{Methods} & \multirow{-2}{*}{E} & I2T           & T2I           & I2T           & T2I           & 0-Shot        \\ \hline
CLIP                      & 25                      & 51.4          & 32.6          & 27.9          & 17.6          & 37.6          \\
SLIP                      & 25                      & 57.2          & 41.2          & 33.6          & 21.9          & {\ul 42.8}    \\
MaskCLIP                  & 25                      & 60.0          & 38.8          & 34.1          & 21.2          & 42.7          \\
A-CLIP                    & 25                      & {\ul 62.7}    & {\ul 42.1}    & {\ul 38.0}    & \textbf{23.2} & \textbf{43.9} \\
\rowcolor[HTML]{ECF4FF} 
DetailCLIP                & 25                      & \textbf{62.8} & \textbf{42.2} & \textbf{38.3} & {\ul 22.9}    & \textbf{43.9} \\ \hline
CLIP                      & 50                      & 53.9          & 35.8          & 30.2          & 19.2          & 39.4          \\
SLIP                      & 50                      & 60.6          & 41.1          & {\ul 33.2}    & 22.3          & 44.1          \\
A-CLIP                    & 50                      & \textbf{66.7} & {\ul 43.2}    & \textbf{39.8} & {\ul 24.4}    & \textbf{46.3} \\
\rowcolor[HTML]{ECF4FF} 
DetailCLIP                & 50                      & {\ul 65.9}    & \textbf{44.7} & \textbf{39.8} & \textbf{24.9} & {\ul 46.2}    \\ \hline \hline
\end{tabular}

\caption{Comparison of model performance on text-image retrieval for Flickr30K, MS COCO, and zero-shot ImageNet-1K (IN-1K). Metrics for image-to-text (I2T) and text-to-image (T2I) retrieval are shown after 25 and 50 epochs ($E$), based on results from A-CLIP \cite{yang2023attentive}.}
\label{tab:res:retreive}
\end{table}

\begin{table*}[ht]
\centering
\scriptsize
\begin{tabular}{c|c|ccccccccccccc|c}
\hline
\rotatebox{90}{Epochs}               & \rotatebox{90}{Methods}                                     & \rotatebox{90}{SUN397}                             & \rotatebox{90}{Aircraft}                              & \rotatebox{90}{DTD}                                   & \rotatebox{90}{Pets}                                  & \rotatebox{90}{Flowers}                               & \rotatebox{90}{GTSRB}                       & \rotatebox{90}{MNIST}                                 & \rotatebox{90}{STL-10}                                & \rotatebox{90}{KITTI}                                 & \rotatebox{90}{Country211 }                         & \rotatebox{90}{PCAM}                                  & \rotatebox{90}{CLEVR}                        & \rotatebox{90}{SST2}                               & \rotatebox{90}{Average}                  \\ \hline
                     & CLIP                                        & 51.1                               & 5.4                                   & 21.2                                  & 28.5                                  & 53.3                                  & 6.1                         & 8.4                                   & 90.5                                  & 35.1                                  & 10.5                               & 53.5                                  & 10.8                         & {\ul 50.7}                         &             32.7             \\
                     & SLIP                                        & 53.4                               & 5.7                                   & {\ul 26.1}                            & 31.1                                  & 56.6                                  & \textbf{14.5}               & 9.8                                   & {\ul 94.4}                            & 34.0                                  & 11.6                               & {\ul 55.4}                            & \textbf{17.5}                & \textbf{51.1}                      &    35.5                      \\
                     & MaskCLIP                                    & 54.0                               & {\ul 8.2}                             & 25.5                                  & \textbf{36.8}                         & 53.6                                  & 10.1                        & {\ul 11.2}                            & 93.9                                  & 30.5                                  & {\ul 12.5}                         & 51.2                                  & 12.9                         & 50.0                               &     34.6                     \\
                     & A-CLIP                                      & \textbf{57.0}                      & 7.6                                   & 26.0                                  & 32.0                                  & {\ul 57.7}                            & {\ul 13.1}                  & 9.8                                   & \textbf{95.4}                         & {\ul 35.2}                            & \textbf{13.5}                      & 51.6                                  & {\ul 14.1}                   & 49.9                               &     {\ul 35.6}                     \\
\multirow{-5}{*}{25} & \cellcolor[HTML]{ECF4FF}\textbf{DetailCLIP} & \cellcolor[HTML]{ECF4FF}{\ul 54.9} & \cellcolor[HTML]{ECF4FF}\textbf{8.7}  & \cellcolor[HTML]{ECF4FF}\textbf{30.1} & \cellcolor[HTML]{ECF4FF}{\ul 35.4}    & \cellcolor[HTML]{ECF4FF}\textbf{60.7} & \cellcolor[HTML]{ECF4FF}9.3 & \cellcolor[HTML]{ECF4FF}\textbf{11.4} & \cellcolor[HTML]{ECF4FF}94.1          & \cellcolor[HTML]{ECF4FF}\textbf{42.2} & \cellcolor[HTML]{ECF4FF}11.9       & \cellcolor[HTML]{ECF4FF}\textbf{62.0} & \cellcolor[HTML]{ECF4FF}13.4 & \cellcolor[HTML]{ECF4FF}50.2       & \cellcolor[HTML]{ECF4FF}\textbf{37.3} \\ \hline
                     & CLIP                                        & 49.0                               & 6.3                                   & 23.5                                  & 27.2                                  & 56.2                                  & 7.4                         & \textbf{12.5}                         & 92.1                                  & {\ul 33.6}                            & 10.9                               & 50.8                                  & {\ul 14.0}                   & {\ul 50.1}                         &         33.4                 \\
                     & SLIP                                        & 54.8                               & 9.0                                   & \textbf{29.8}                         & 31.9                                  & 57.7                                  & {\ul 9.0}                   & 9.8                                   & {\ul 95.6}                            & \textbf{35.1}                         & 12.7                               & 54.4                                  & 13.8                         & 49.9                               &      35.7                    \\
                     & A-CLIP                                      & \textbf{58.7}                      & {\ul 10.2}                            & {\ul 27.7}                            & {\ul 40.5}                            & {\ul 61.0}                            & \textbf{9.4}                & {\ul 11.3}                            & 95.5                                  & 23.3                                  & \textbf{14.4}                      & {\ul 63.7}                            & \textbf{19.6}                & \textbf{52.3}                      &    {\ul 37.5}                      \\
\multirow{-4}{*}{50} & \cellcolor[HTML]{ECF4FF}\textbf{DetailCLIP} & \cellcolor[HTML]{ECF4FF}{\ul 57.5} & \cellcolor[HTML]{ECF4FF}\textbf{11.0} & \cellcolor[HTML]{ECF4FF}\textbf{29.8} & \cellcolor[HTML]{ECF4FF}\textbf{40.8} & \cellcolor[HTML]{ECF4FF}\textbf{64.1} & \cellcolor[HTML]{ECF4FF}7.3 & \cellcolor[HTML]{ECF4FF}10.8          & \cellcolor[HTML]{ECF4FF}\textbf{95.8} & \cellcolor[HTML]{ECF4FF}30.9          & \cellcolor[HTML]{ECF4FF}{\ul 13.3} & \cellcolor[HTML]{ECF4FF}\textbf{63.9} & \cellcolor[HTML]{ECF4FF}12.7 & \cellcolor[HTML]{ECF4FF}{\ul 50.1} & \cellcolor[HTML]{ECF4FF}\textbf{37.6} \\ \hline \hline
\end{tabular}
\caption{Zero-shot performance evaluation across multiple classification benchmarks. Epochs indicate the total number of training epochs. The results were collected in the A-CLIP paper \cite{yang2023attentive}.}
\label{tab:res:classification}
\end{table*}

\section{Conclusion}

In this paper, we introduce DetailCLIP, a self-improving vision-language framework designed to overcome the limitations of traditional CLIP-based models in both fine- and coarse-grained tasks. While CLIP excels at aligning global image-text representations, it often misses the fine-grained details needed for tasks like segmentation and object detection. To address this, DetailCLIP incorporates self-curated learning objectives—patch-level self-distillation, pixel-level reconstruction, and attention-based token filtering—to iteratively refine both high-level semantics and detailed features. Experiments show that DetailCLIP generalizes well across tasks, surpassing state-of-the-art models on detail-focused benchmarks like ADE20K and MS COCO, while also delivering strong performance on coarse-grained tasks such as zero-shot classification.


\section*{Acknowledgments}

This project was made possible, in part, by support from the National Science Foundation grant number OAC-2018627 and from the Ohio Supercomputer Center.

\bibliographystyle{iclr2025_conference}  
\bibliography{iclr2025_conference}  

\appendix
\newpage
\section{Appendix}

\subsection{Attention Token Removal}
\label{sec:attn_removal}
\begin{figure*}[ht]
\centering
\begin{tabular}{cccccc}
\includegraphics[width=0.135\textwidth]{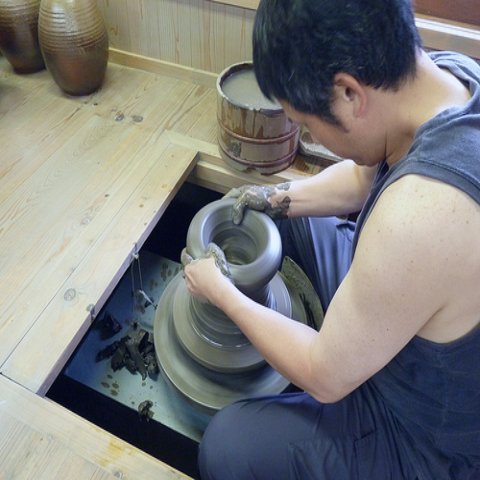} & \includegraphics[width=0.135\textwidth]{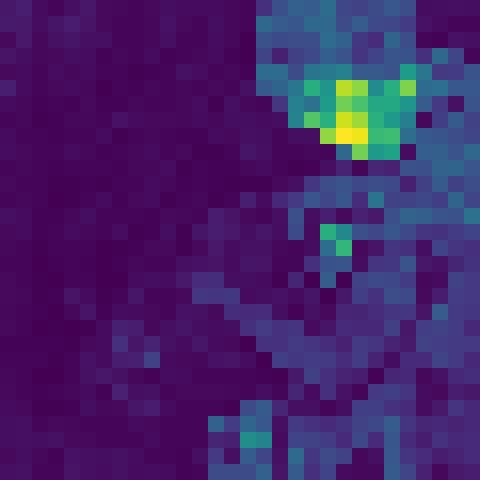} & \includegraphics[width=0.135\textwidth]{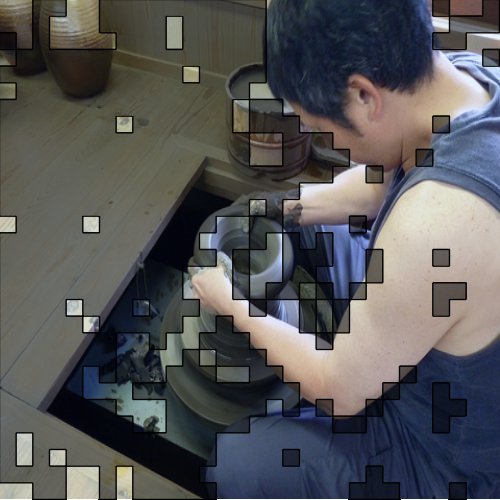} & \includegraphics[width=0.135\textwidth]{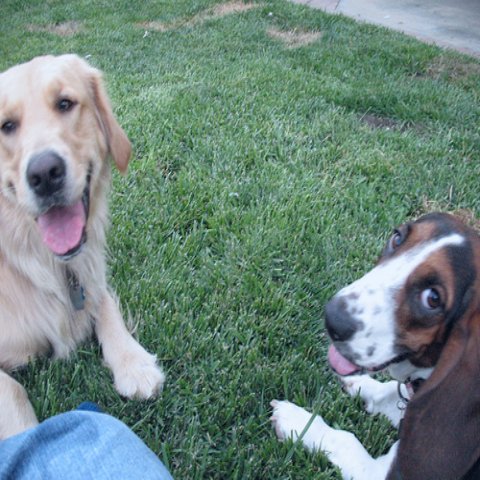}  & \includegraphics[width=0.135\textwidth]{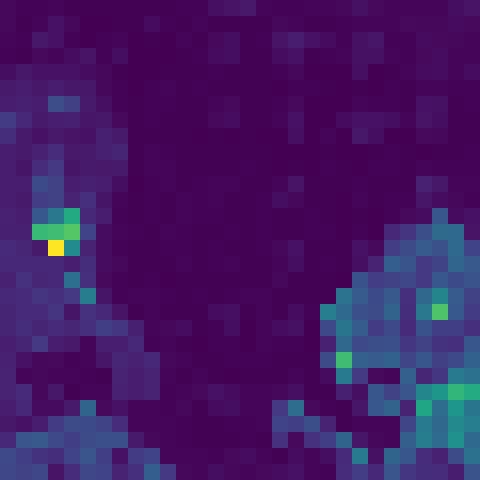} & \includegraphics[width=0.135\textwidth]{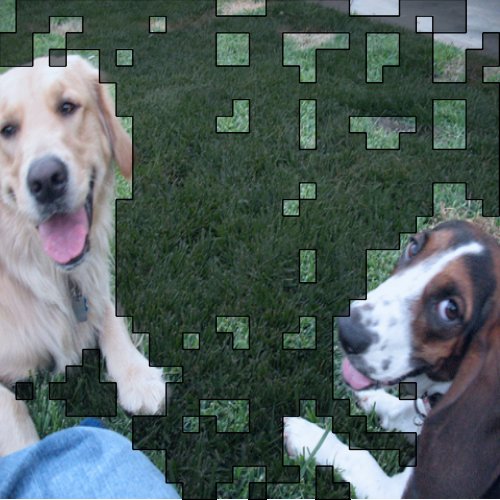} \\

\multicolumn{3}{c}{\parbox{0.4\textwidth}{\centering a person crafting pottery on a potter's wheel, installed in a lowered section of the floor.}} & \multicolumn{3}{c}{\parbox{0.4\textwidth}{\centering Two dogs sitting on the grass; one is a golden retriever, and the other is a basset hound.}}  \\

\includegraphics[width=0.135\textwidth]{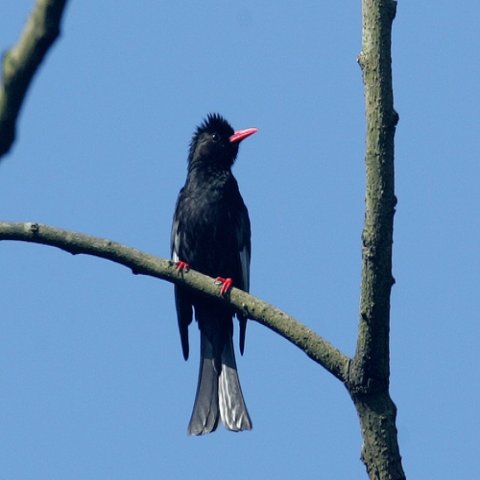} & \includegraphics[width=0.135\textwidth]{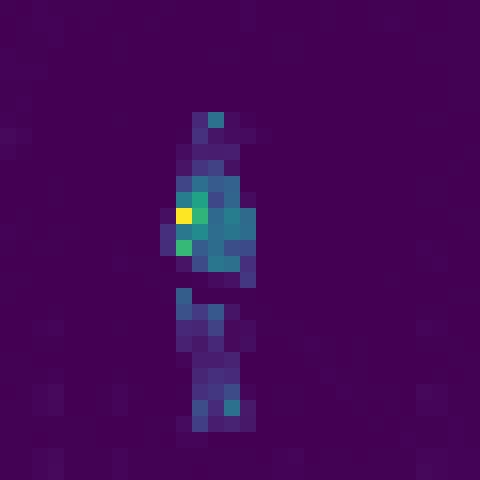} & \includegraphics[width=0.135\textwidth]{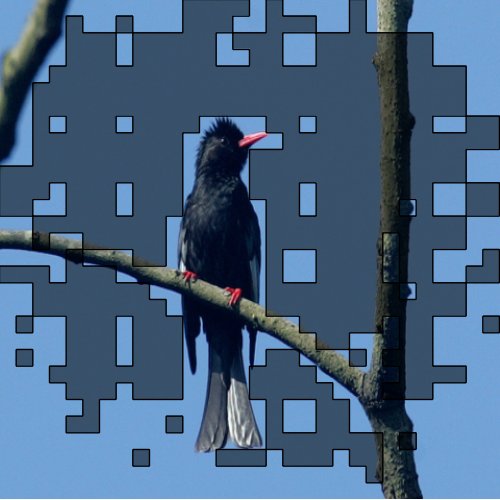} & \includegraphics[width=0.135\textwidth]{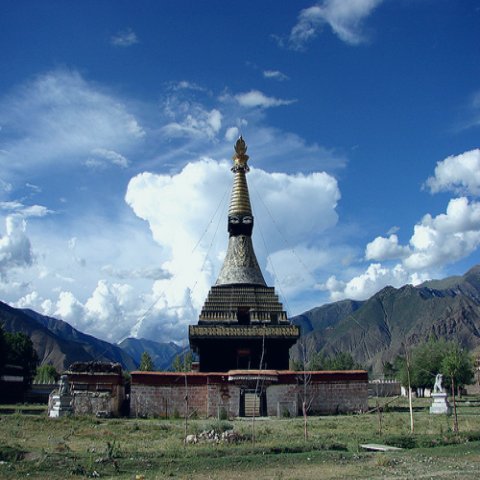}  & \includegraphics[width=0.135\textwidth]{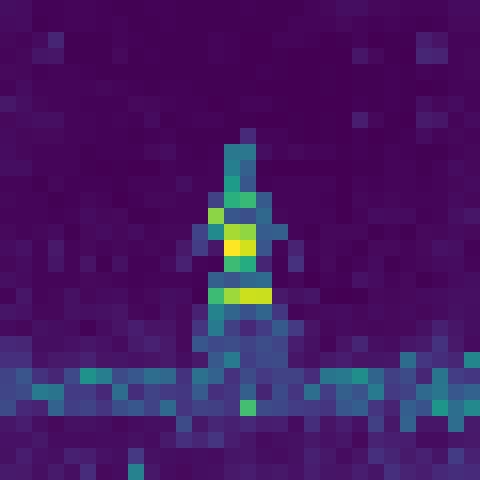} & \includegraphics[width=0.135\textwidth]{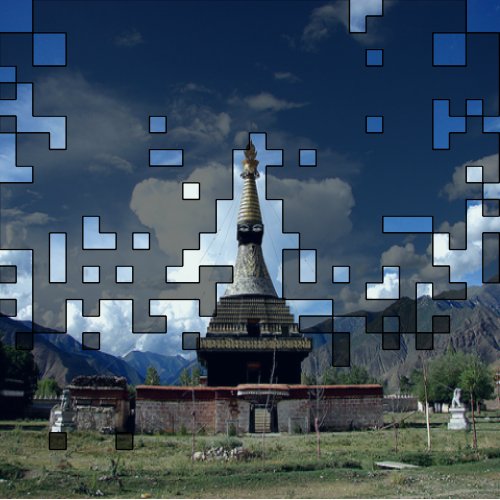} \\

\multicolumn{3}{c}{\parbox{0.4\textwidth}{\centering A black bird with a red beak perched on a tree branch against a blue sky.}} & \multicolumn{3}{c}{\parbox{0.4\textwidth}{\centering A large stupa with a golden spire in a mountainous landscape under a blue sky.}}  \\ 


\end{tabular}

\caption{Illustration of the token removal mechanism, which retains semantically significant tokens to focus on critical input aspects while discarding unnecessary details.}

\label{fig:token:removal}
\end{figure*}

In Figure~\ref{fig:token:removal}, the mechanism of token removal is illustrated. The process selectively eliminates less relevant tokens, retaining only those with strong semantic significance. This approach enhances the model's focus on the most important aspects of the input, ensuring that critical information is preserved while unnecessary details are discarded. The current settings of the model use a 50\% mask ratio, preserving only the semantically relevant details of the image. For example, in the image with the two dogs, the grass background is removed as it is less important to the overall image. Similarly, in the image of the black bird, the tree and branches are ignored while preserving the bird, as it is central to the image. In the DetailCLIP model, the attention scores are derived from the teacher network and are used to guide the student network during training.

\subsection{Computational Efficiency}
\label{sec:compute}

Our model demonstrates an effective balance between performance and efficiency, as highlighted in Table \ref{tab:res:memory_train} and Table \ref{tab:res:retreive}. It requires only 24G of GPU memory, significantly less than SLIP (30G), while achieving a training time of 1.16x—closely aligned with CLIP (1.00x) and notably faster than SLIP (2.67x) and MaskCLIP (1.56x). Our resource usage and  training speed make DetailCLIP highly suitable for applications with limited GPU resources, while offering strong performance on downstream tasks.

\begin{table}[]
\centering
\begin{tabular}{c|c|c}
\hline
Methods & Training Time & GPU Memory \\ \hline
CLIP             & 1.00x                  & 14G                 \\
SLIP             & 2.67x                  & 30G                 \\
MaskCLIP         & 1.56x                  & 16G                 \\
A-CLIP           & 1.16x                  & 14G                 \\
DetailCLIP       & 1.16x                  & 24G                 \\ \hline
\end{tabular}

\caption{Training time and GPU memory usage of various models, measured consistently on the same device. Training time is reported relative to the baseline CLIP (1.00x), and GPU memory is shown in gigabytes (GB) \cite{yang2023attentive}.}
\label{tab:res:memory_train}
\end{table}

\subsection{Pseudocode}
\label{sec:pseudocode}

Algorithm~\ref{alg:detailclip} presents the pseudocode for the DetailCLIP algorithm, which integrates CLIP, self-distillation, and pixel-level loss. The steps include updating the momentum encoder using EMA, generating teacher embeddings for both views (without gradients), and creating attention-based masks with a 50\% mask ratio. Student embeddings are then obtained after applying the masks and projected using CLIP projections alongside text embeddings. The model produces outputs for student distillation and reconstruction, followed by computing classification, patch-level, and reconstruction losses to calculate the total loss.

\lstdefinelanguage{PythonStyle}{
    language=Python,
    basicstyle=\ttfamily\footnotesize,
    keywordstyle=\bfseries\color{blue},
    commentstyle=\color{gray!80!black},
    stringstyle=\color{red},
    emph={forward, update_momentum_encoder, get_att_mask, compute_clip_loss, compute_distill_loss, compute_recon_loss},
    emphstyle=\bfseries\color{black},
    morekeywords={def, return, range, normalize, CrossEntropy, exp, cat, tokenize, crop, aug, with, no_grad, concat},
}

\begin{algorithm}
    \caption{\textbf{Pytorch-like code for DetailCLIP}}
    \label{alg:detailclip}
\begin{lstlisting}[language=PythonStyle]
# u, v: view 1 and view 2 of the image
# s_visual, t_visual: student and teacher image encoders
# t_dist_head, s_dist_head: teacher and student distillation heads
# encode_image, encode_text: students CLIP image and text projections

# reconstr_head: pixel-level reconstruction head
def forward(u, v, text, momentum):
    # Update momentum encoder using EMA
    update_momentum_encoder(momentum)

    # Teacher embeddings for both views (no gradient)
    with no_grad():
        t_embed_u, attn_u = t_visual(u)
        t_embed_v, attn_v = t_visual(v)
        t_dist_out = t_dist_head(concat(t_embed_u, t_embed_v))

    # Generate masks from attention using 50% mask ratio
    mask_u, mask_v = get_att_mask(attn_u), get_att_mask(attn_v)

    # Get student embeddings after applying masks
    s_embed_u = s_visual(u, mask=mask_u)
    s_embed_v = s_visual(v, mask=mask_v)

    # Project student embeddings using CLIP projections
    img_emb_u = encode_image(s_embed_u)
    img_emb_v = encode_image(s_embed_v)

    # Get text embedding
    text_emb = encode_text(tokenize(text))

    # Student distillation output
    s_dist_out = s_dist_head(concat(s_embed_u, s_embed_v))

    # Reconstruction
    s_dec_u = decoder(s_embed_u)
    s_dec_v = decoder(s_embed_v)
    u_rec = h_reconst(s_dec_u)
    v_rec = h_reconst(s_dec_v)

    # Compute losses
    clip_loss = compute_clip_loss(img_emb_u, img_emb_v, text_emb)
    distill_loss = compute_distill_loss(s_dist_out, t_dist_out)
    recon_loss = compute_recon_loss(u_rec, v_rec, mask_u, mask_v)

    # Total loss
    total_loss = clip_loss + distill_loss + recon_loss
    return total_loss
\end{lstlisting}
\end{algorithm}

\subsection{Ablation Study on Loss Weight}
\label{sec:ablation_weight}

\begin{table}[ht]
\centering
\begin{tabular}{c|c|c|c|c|c}
\hline
\multirow{3}{*}{} & $\alpha_1=1$ & $\alpha_1=1$ & $\alpha_1=1$ & $\alpha_1=.5$ & $\alpha_1=0$ \\
                  & $\alpha_2=1$ & $\alpha_2=1$ & $\alpha_2=1$ & $\alpha_2=.5$ & $\alpha_2=0$ \\
                  & $\alpha_3=1$ & $\alpha_3=0$ & $\alpha_3=2$ & $\alpha_3=.5$ & $\alpha_3=1$ \\ \hline
   Acc            & \textbf{43.9} & 42.9 & 43.2 & {\ul 43.3} & 42.6 \\ \hline \hline
\end{tabular}
\caption{Impact of varying $\alpha$ values on the zero-shot accuracy of ImageNet-1k after 25 epochs of training with the DetailCLIP.}
\label{tab:abla:weight:sum}
\end{table}

In this ablation study, we investigate the impact of different weight configurations on the comprehensive loss function $\mathcal{L}_{\text{tot}}$ introduced in Eq. \ref{eq:loss_total}. We aim to identify the optimal values for the hyperparameters $\alpha_1$, $\alpha_2$, and $\alpha_3$ that effectively balance the various losses.

Table \ref{tab:abla:weight:sum} presents the results of this exploration, showing how different combinations of these weights influence the zero-shot accuracy on the ImageNet-1k dataset after $25$ epochs of training. The accuracy values reflect the model's performance under varying degrees of emphasis on each loss term.

Analyzing the results, we find that the baseline configuration, where all weights are set to 1, achieves the highest accuracy of $43.9\%$, suggesting that equal emphasis on all loss components provides a robust performance baseline. When $\alpha_3$ is reduced to 0, thereby ignoring the reconstruction loss, the accuracy slightly decreases to $42.9\%$, indicating that while this loss component is not critical, it still positively influences overall model performance. Doubling $\alpha_3$ to 2 results in a marginal accuracy decrease to $43.2\%$, implying that an overemphasis on the reconstruction loss can detract slightly from the model’s effectiveness in other tasks. Lowering the weights of $\alpha_1$ and $\alpha_2$ to $0.5$ yields an accuracy of $43.3\%$, demonstrating that moderate weight reductions do not significantly impact performance. However, eliminating these ($\alpha_1 = 0$ and $\alpha_2 = 0$) reduces accuracy further to $42.6\%$, underscoring their essential role in the model’s learning process.

\subsection{Visual Analysis}
\label{sec:visual_analysis}
\begin{figure*}[ht]
\centering
\begin{tabular}{c|c|cccc}
\hline
\textbf{Image} &\textbf{Ground Truth} & \textbf{A-Clip} & \textbf{SLIP} & \textbf{DetailCLIP-25} & \textbf{DetailCLIP-50} \\ \hline

\includegraphics[width=0.135\textwidth]{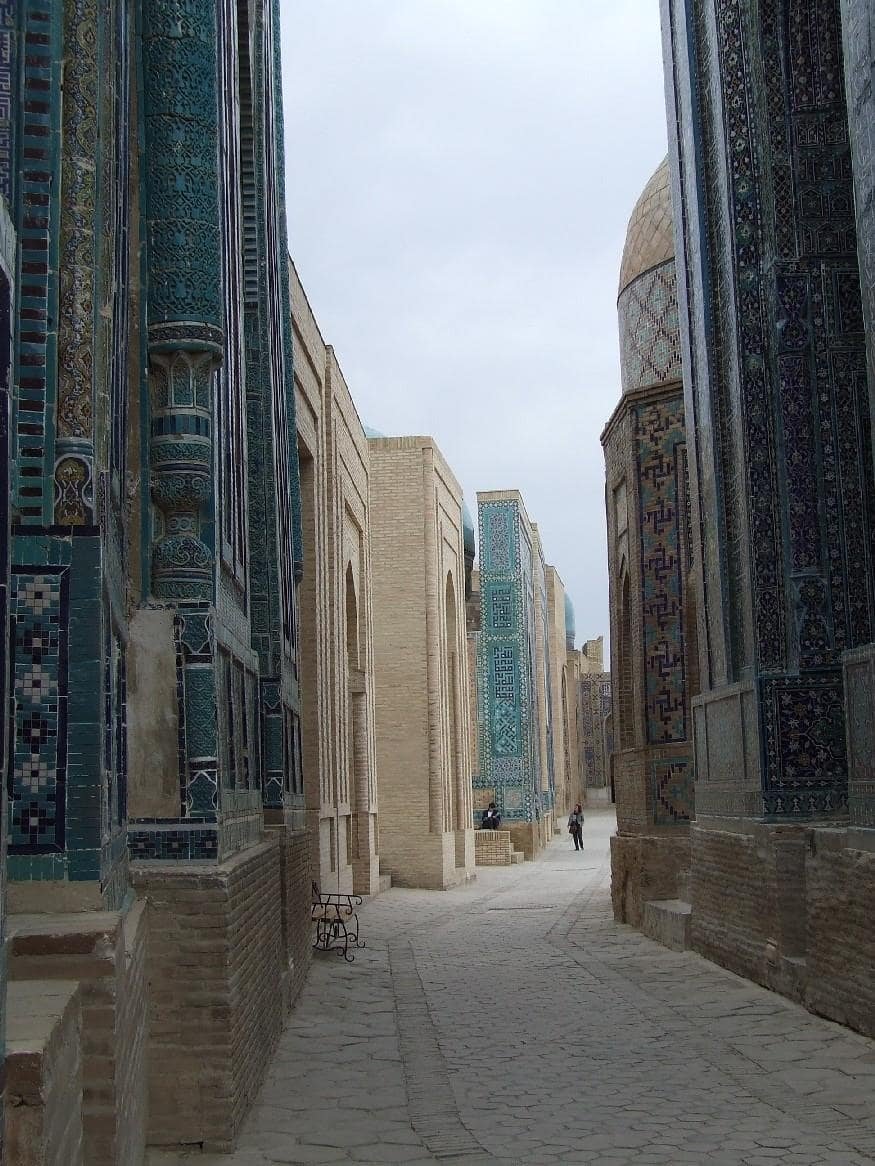} & \includegraphics[width=0.135\textwidth]{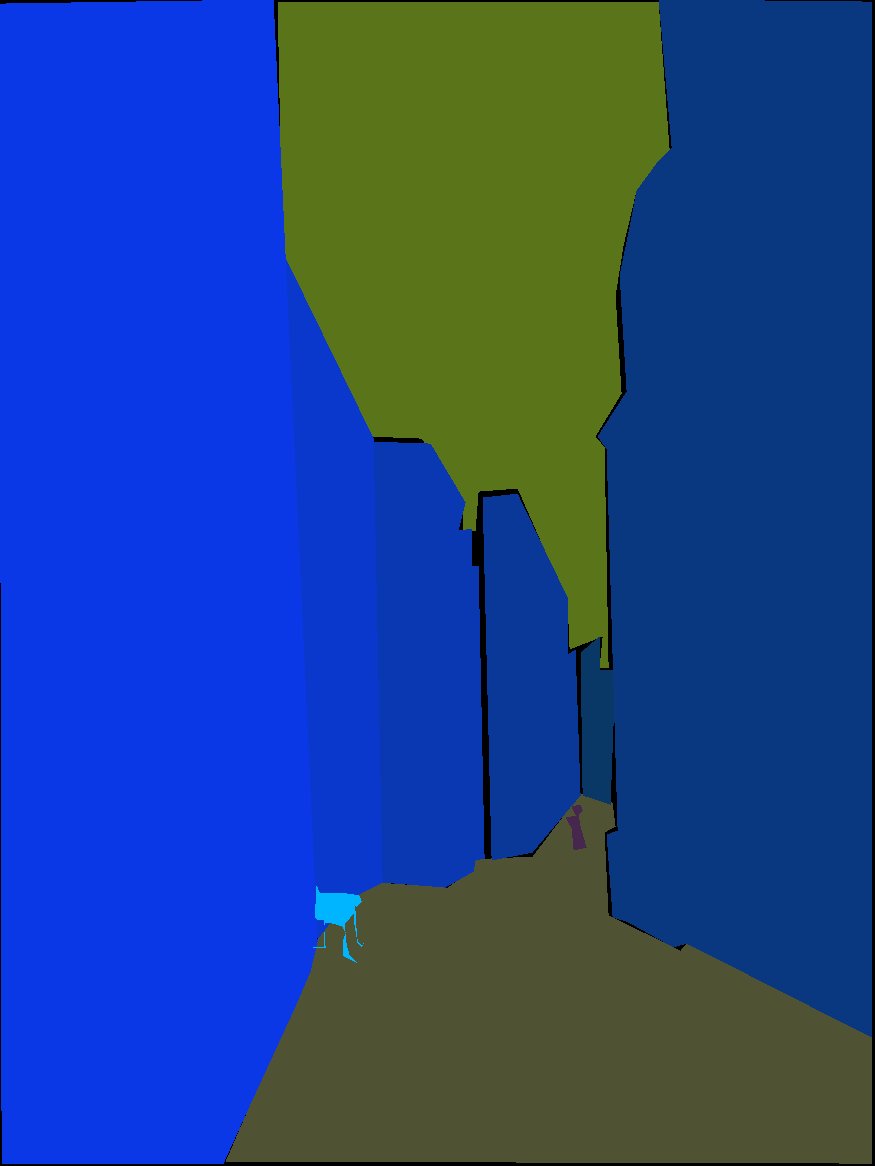} & \includegraphics[width=0.135\textwidth]{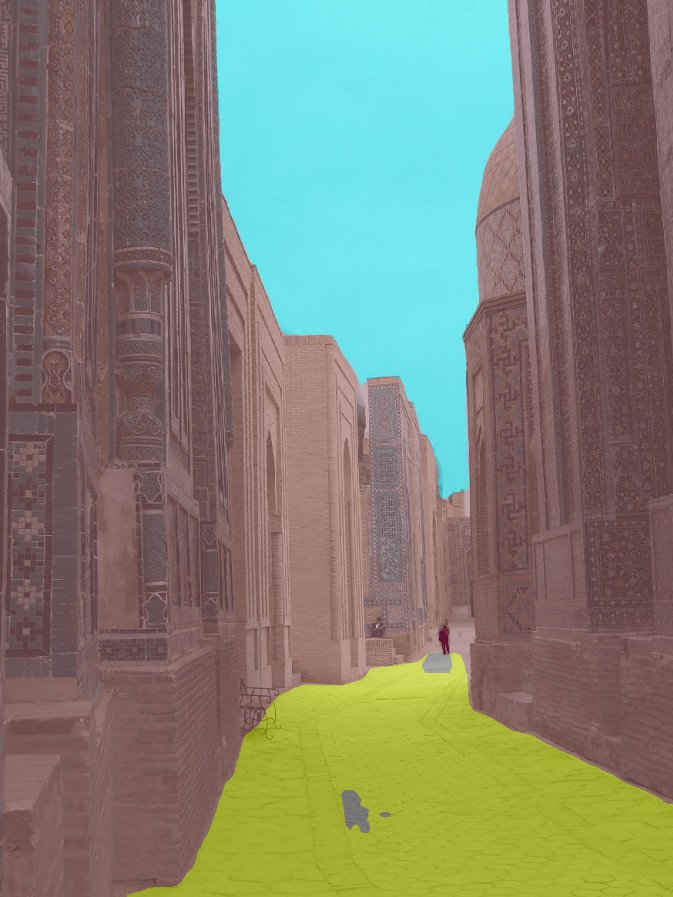} & \includegraphics[width=0.135\textwidth]{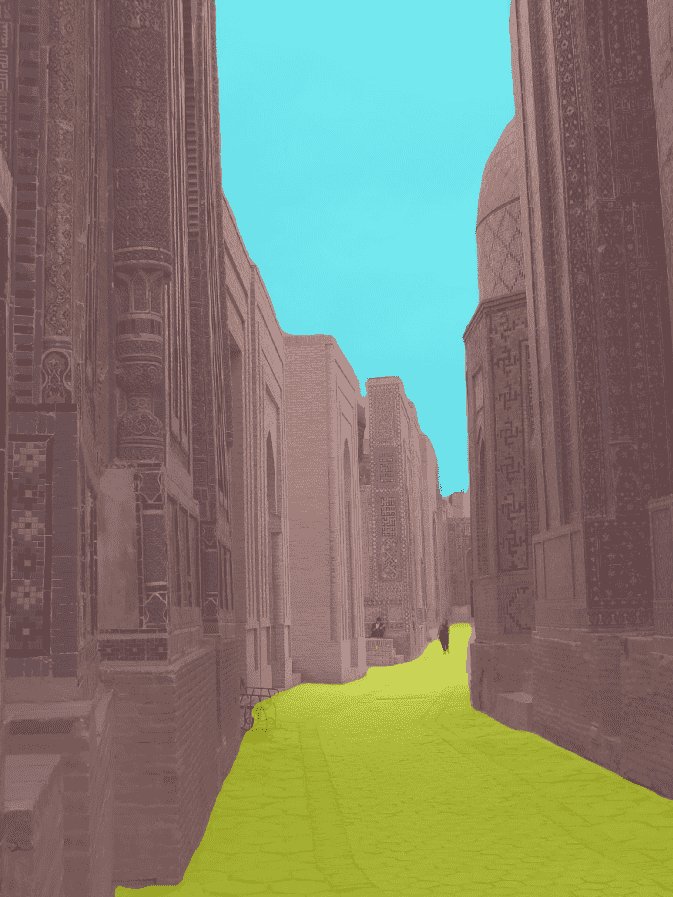}  & \includegraphics[width=0.135\textwidth]{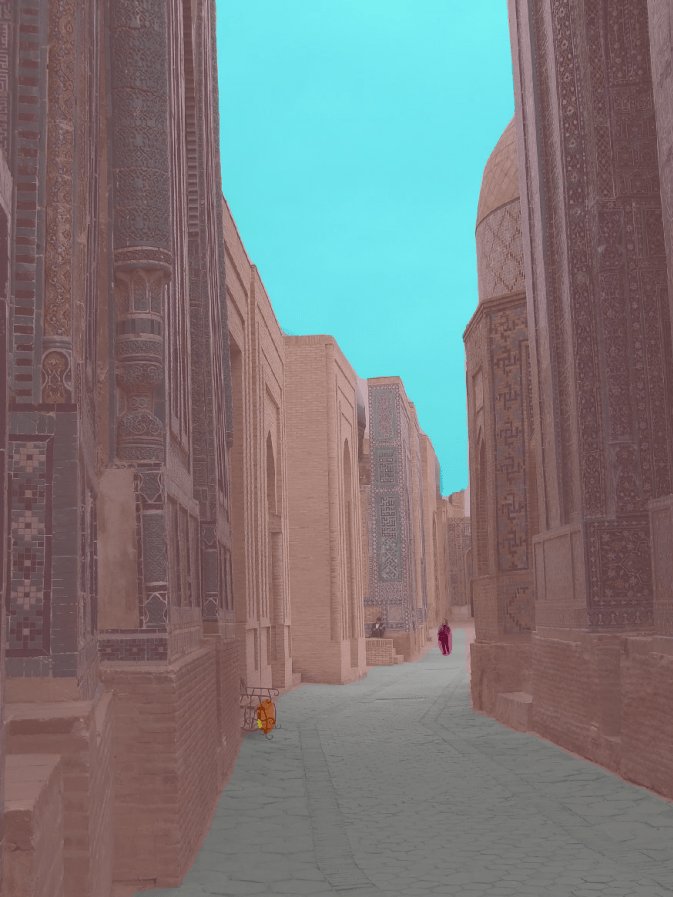} & \includegraphics[width=0.135\textwidth]{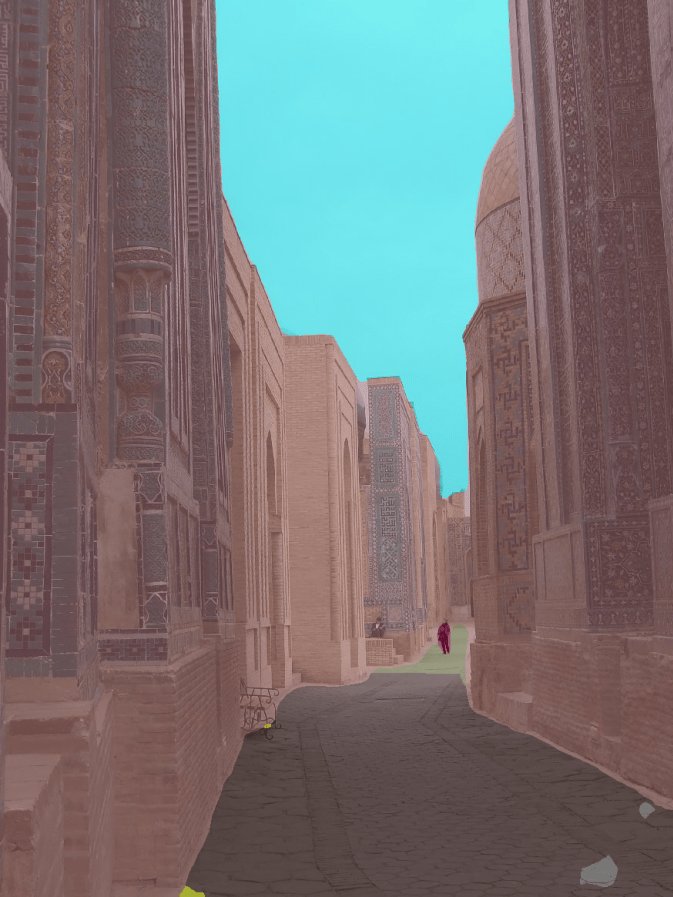} \\

\includegraphics[width=0.135\textwidth]{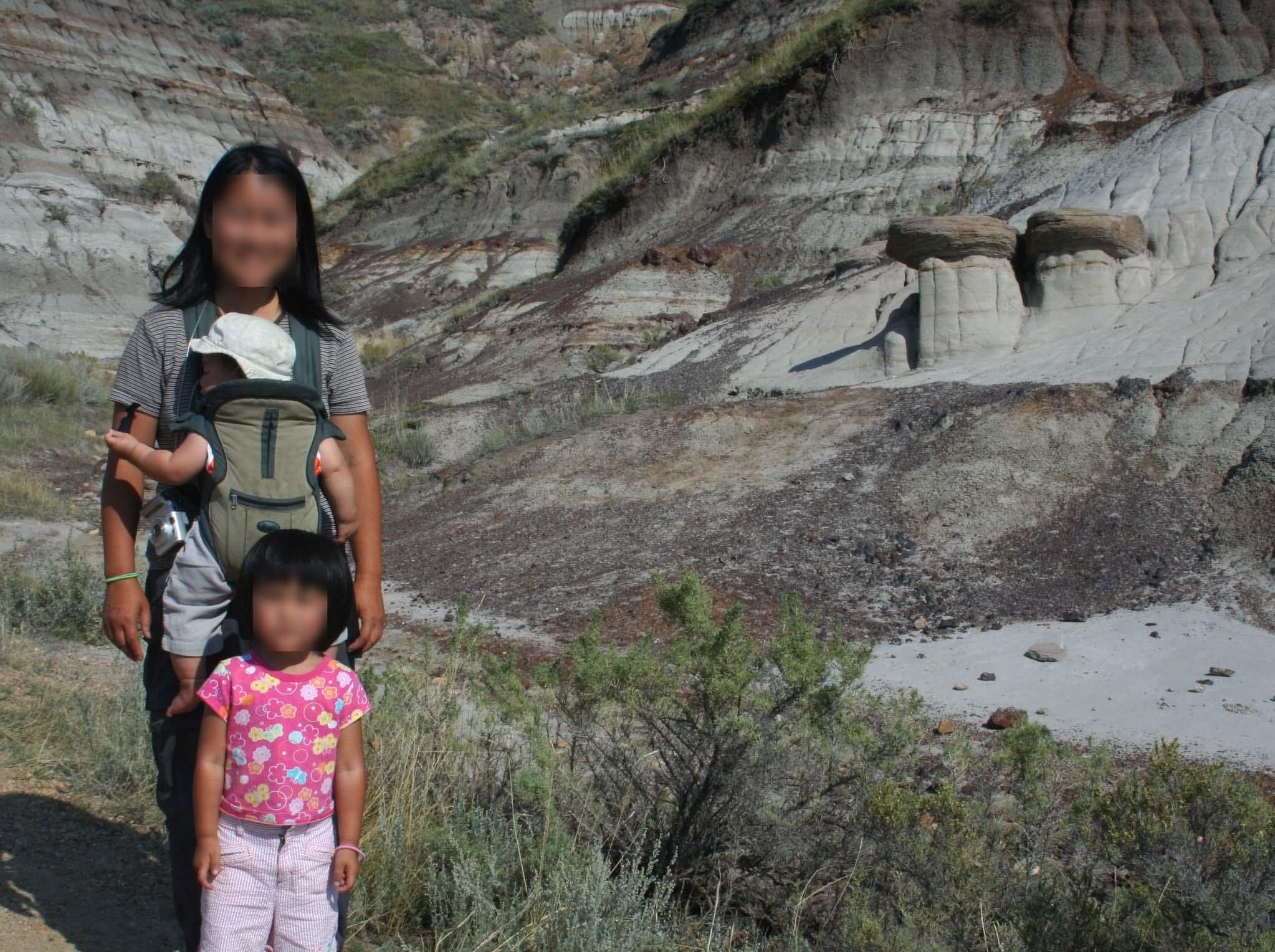} & \includegraphics[width=0.135\textwidth]{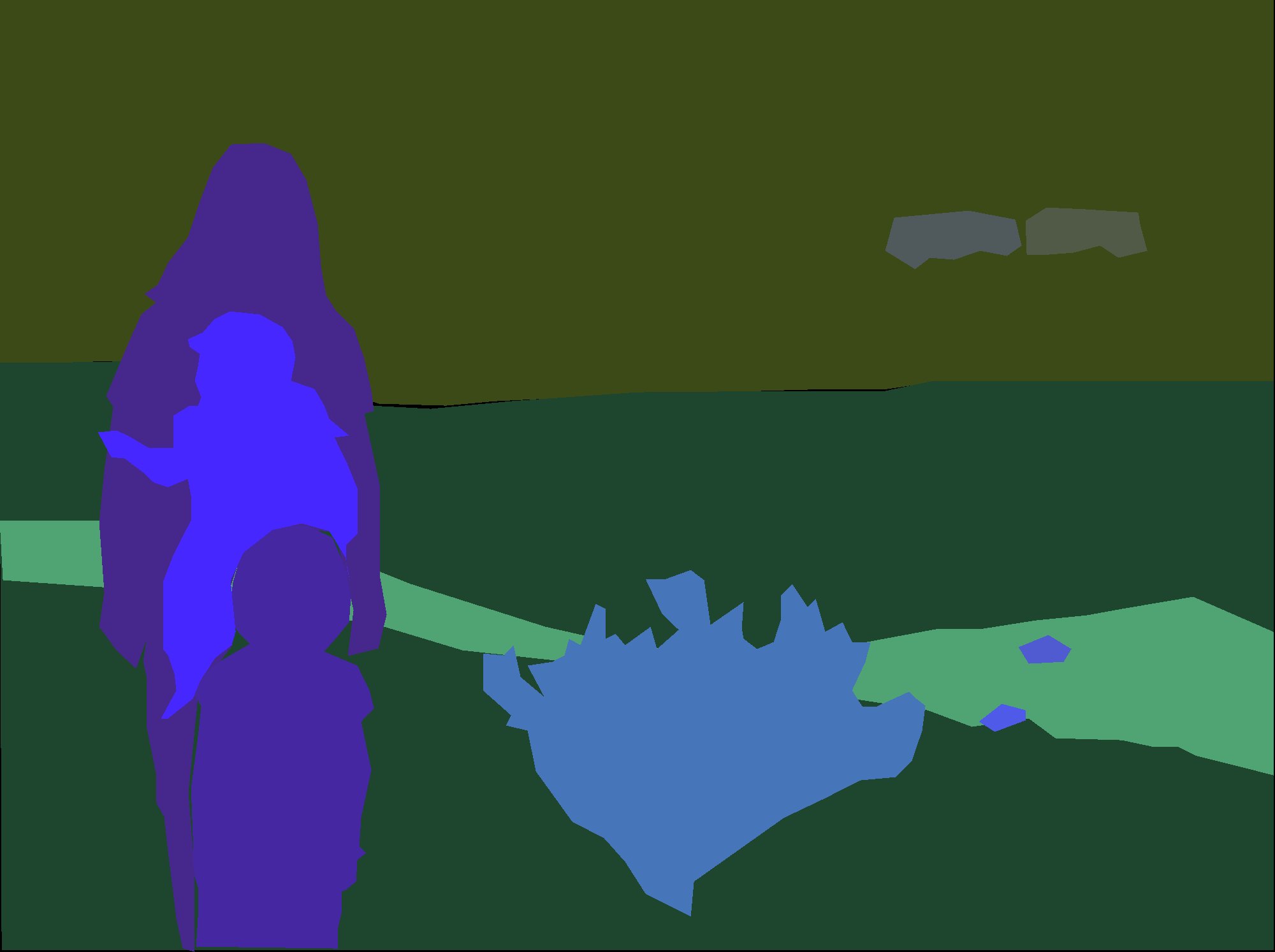} & \includegraphics[width=0.135\textwidth]{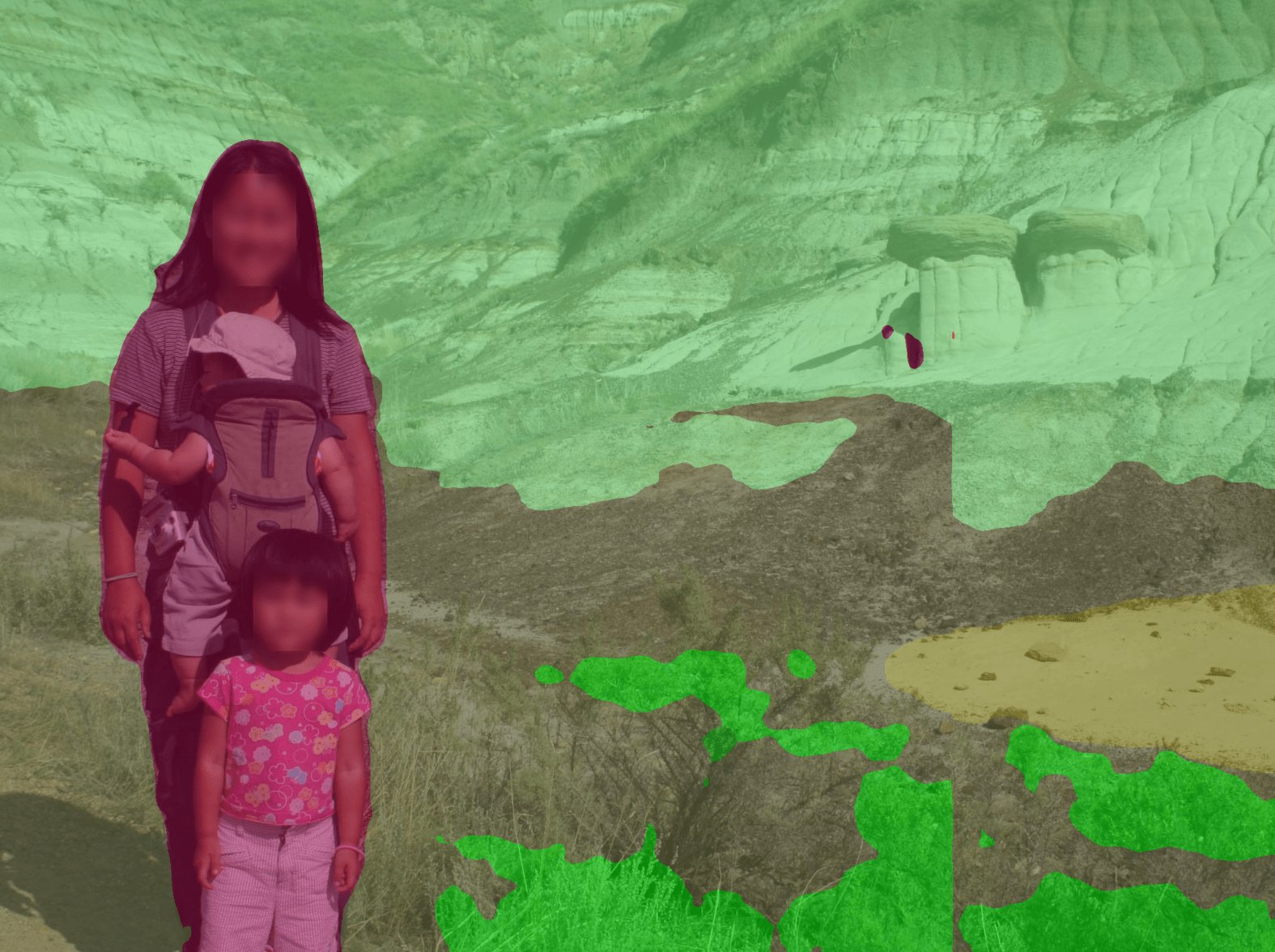} & \includegraphics[width=0.135\textwidth]{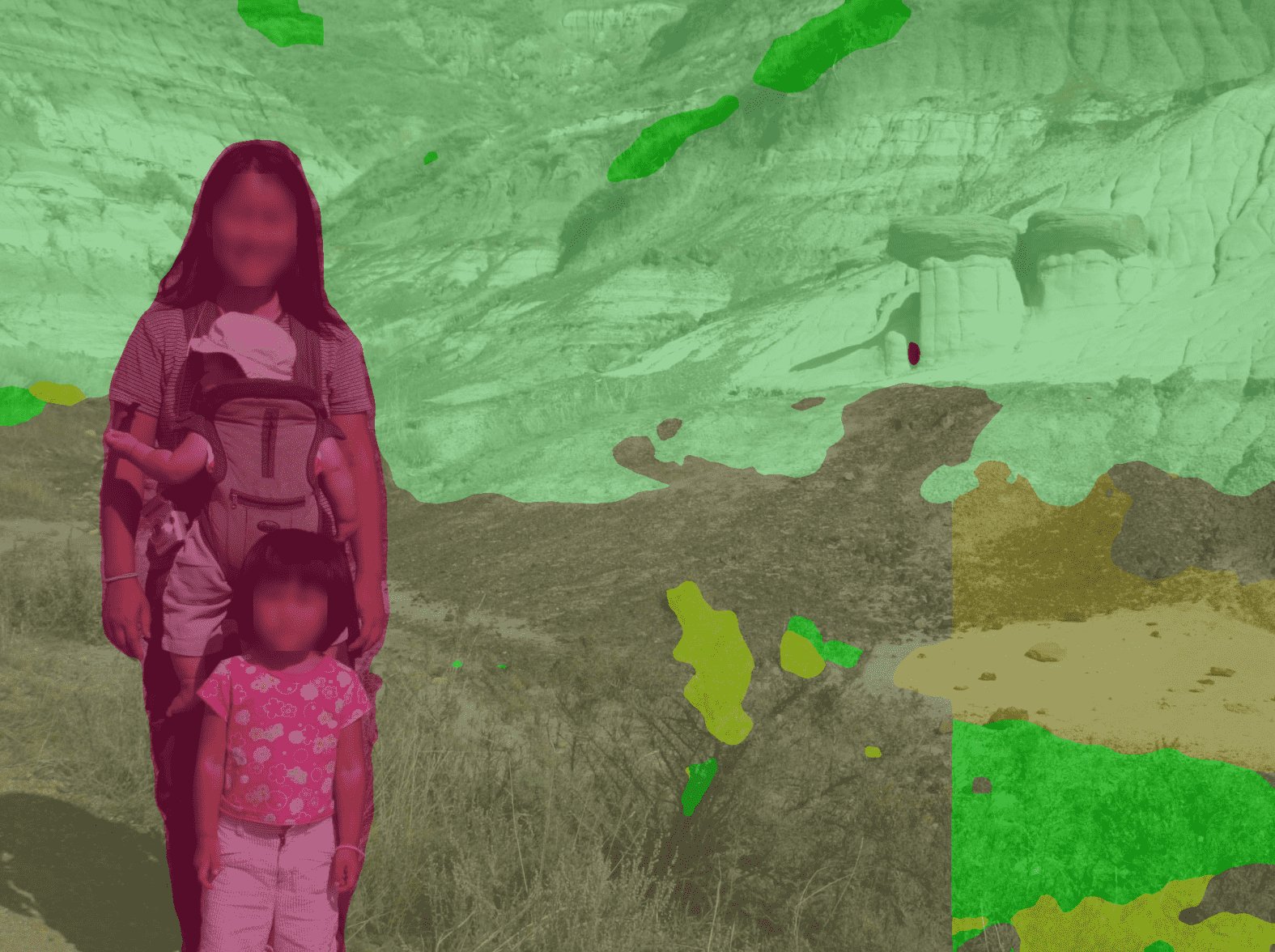}  & \includegraphics[width=0.135\textwidth]{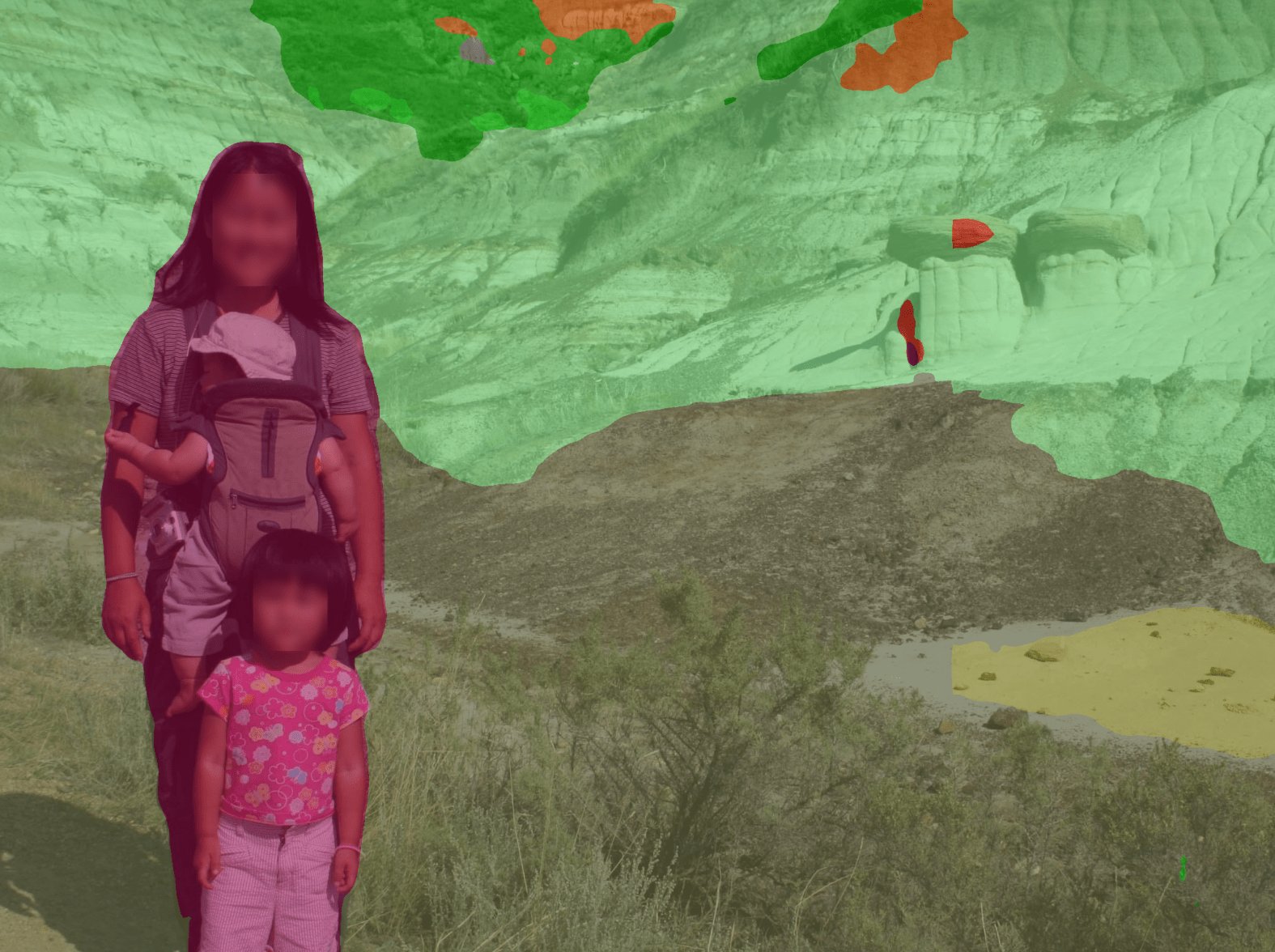} & \includegraphics[width=0.135\textwidth]{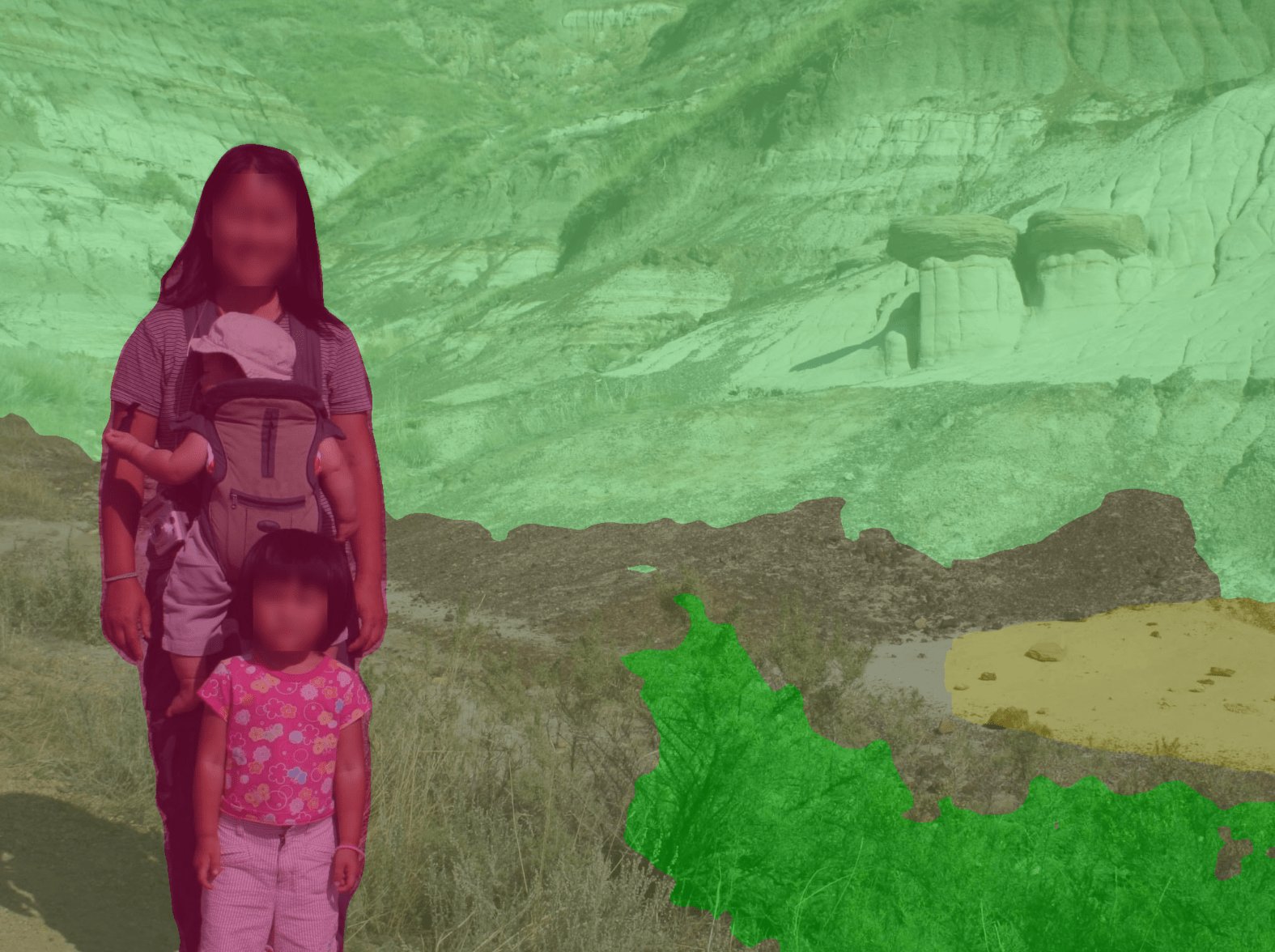} \\

\includegraphics[width=0.135\textwidth]{image_files/semantic_segmentation/raw/ADE_val_00001073.jpg} & \includegraphics[width=0.135\textwidth]{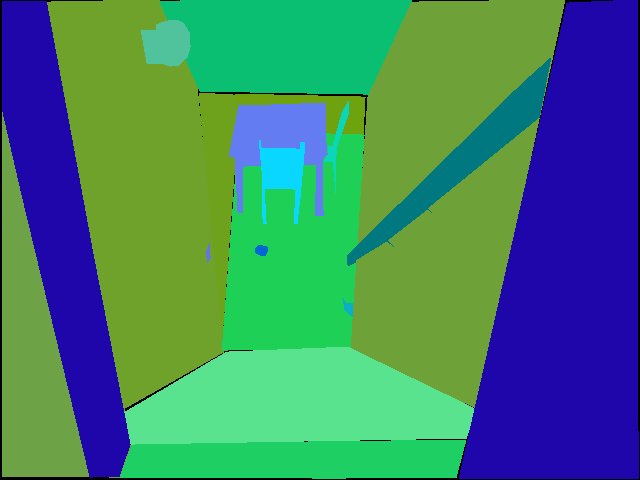} & \includegraphics[width=0.135\textwidth]{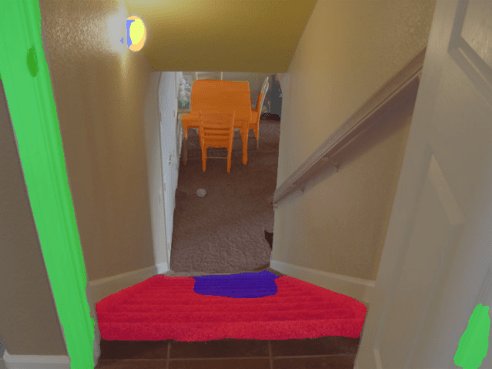} & \includegraphics[width=0.135\textwidth]{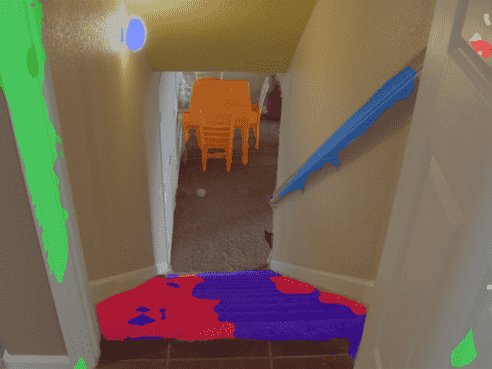}  & \includegraphics[width=0.135\textwidth]{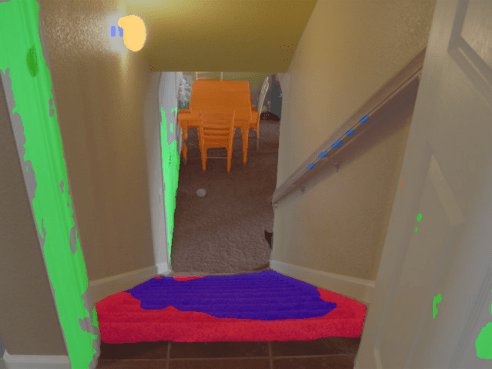} & \includegraphics[width=0.135\textwidth]{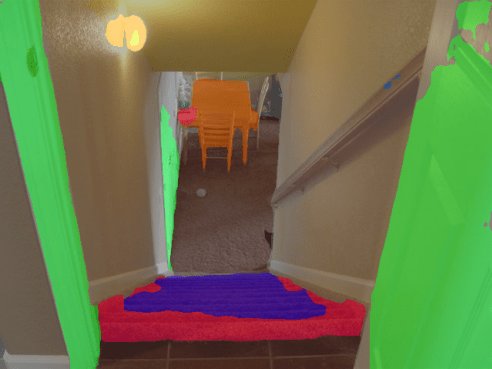} \\

\includegraphics[width=0.135\textwidth]{image_files/semantic_segmentation/raw/ADE_val_00001083.jpg} & \includegraphics[width=0.135\textwidth]{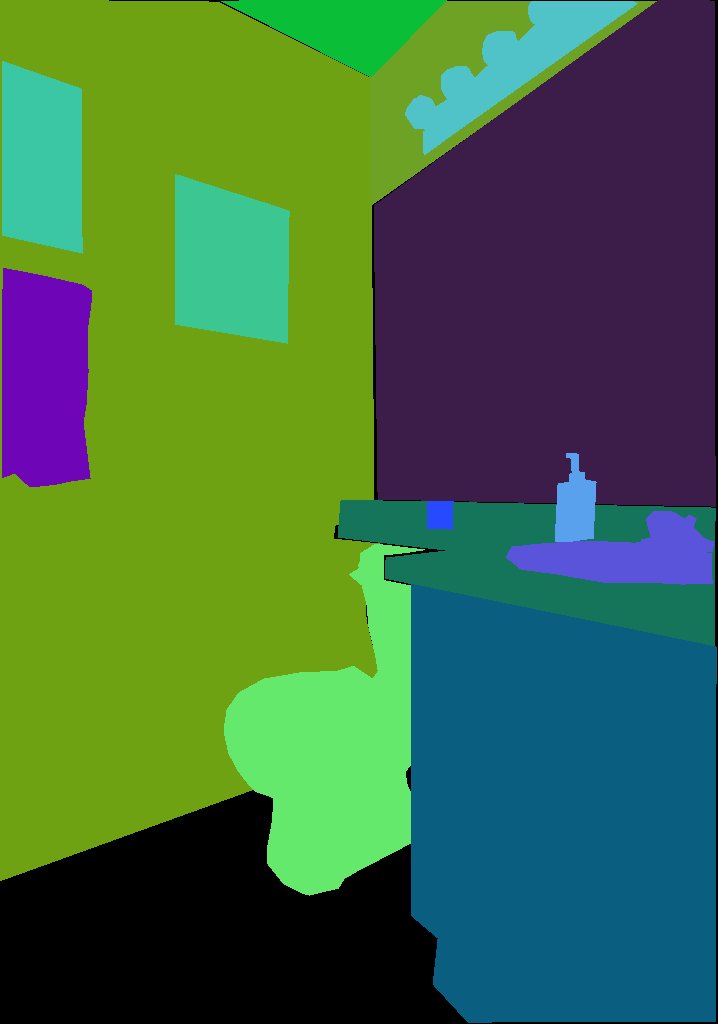} & \includegraphics[width=0.135\textwidth]{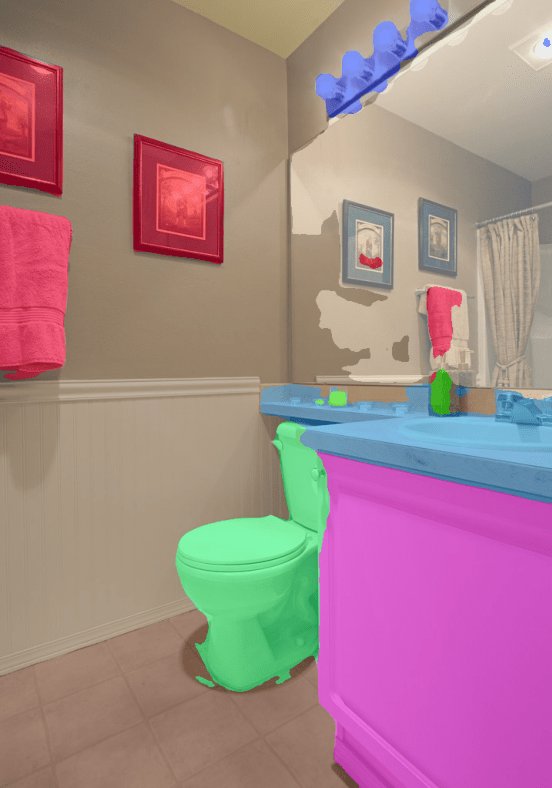} & \includegraphics[width=0.135\textwidth]{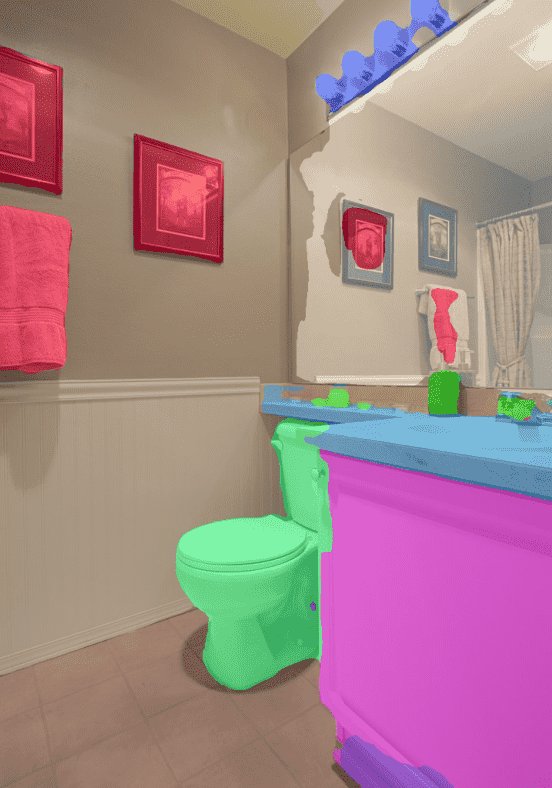}  & \includegraphics[width=0.135\textwidth]{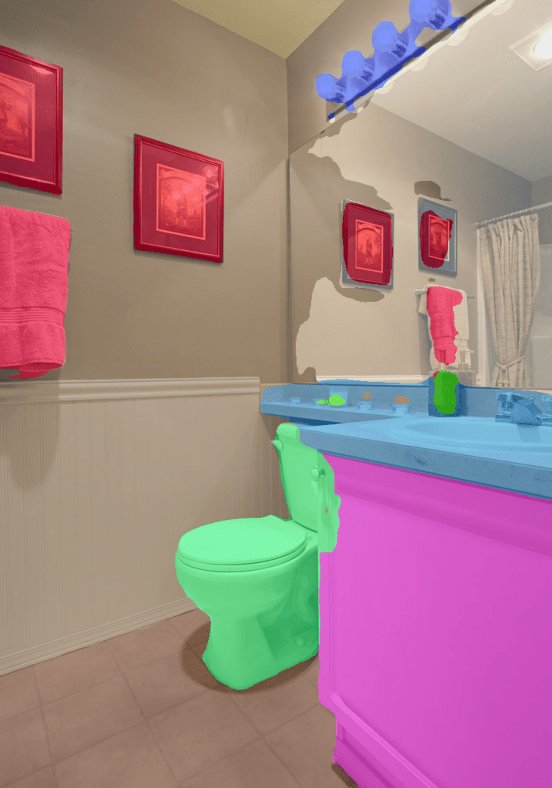} & \includegraphics[width=0.135\textwidth]{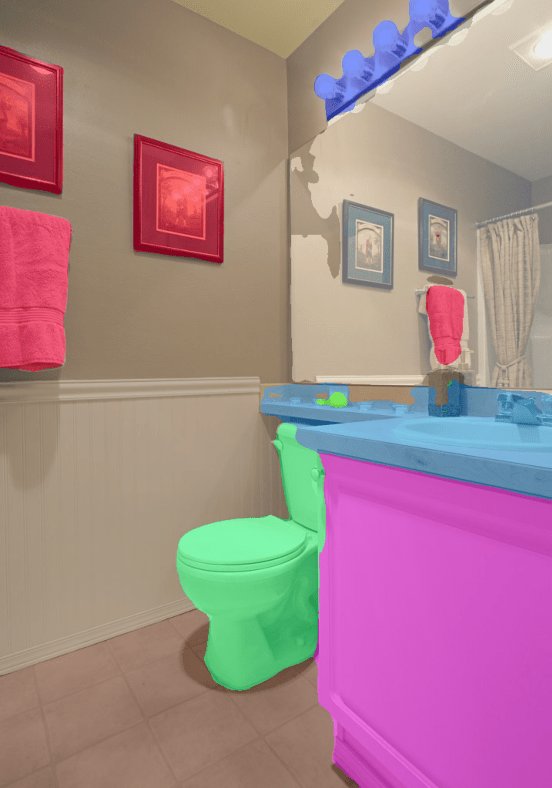} \\

\includegraphics[width=0.135\textwidth]{image_files/semantic_segmentation/raw/ADE_val_00001106.jpg} & \includegraphics[width=0.135\textwidth]{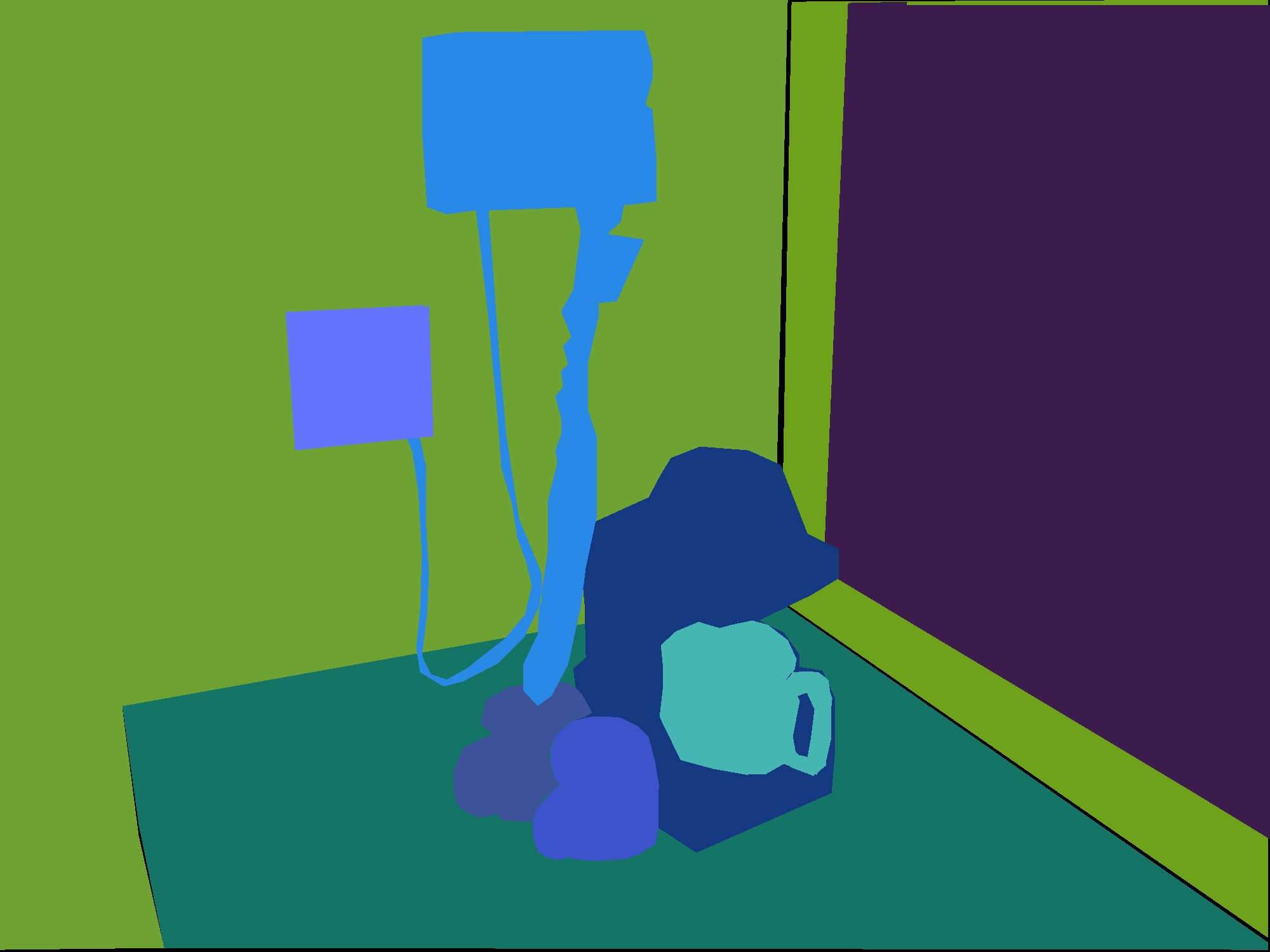} & \includegraphics[width=0.135\textwidth]{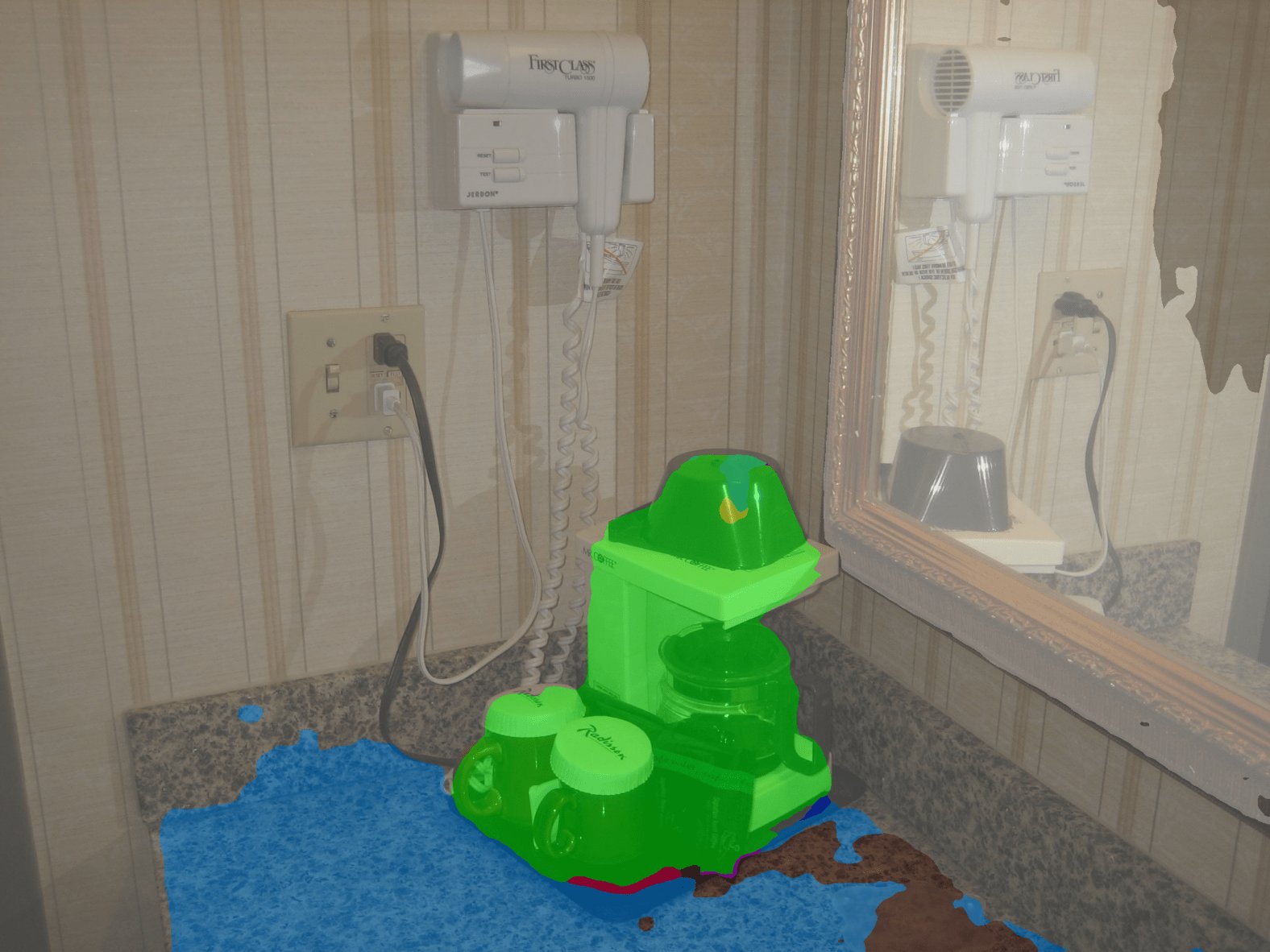} & \includegraphics[width=0.135\textwidth]{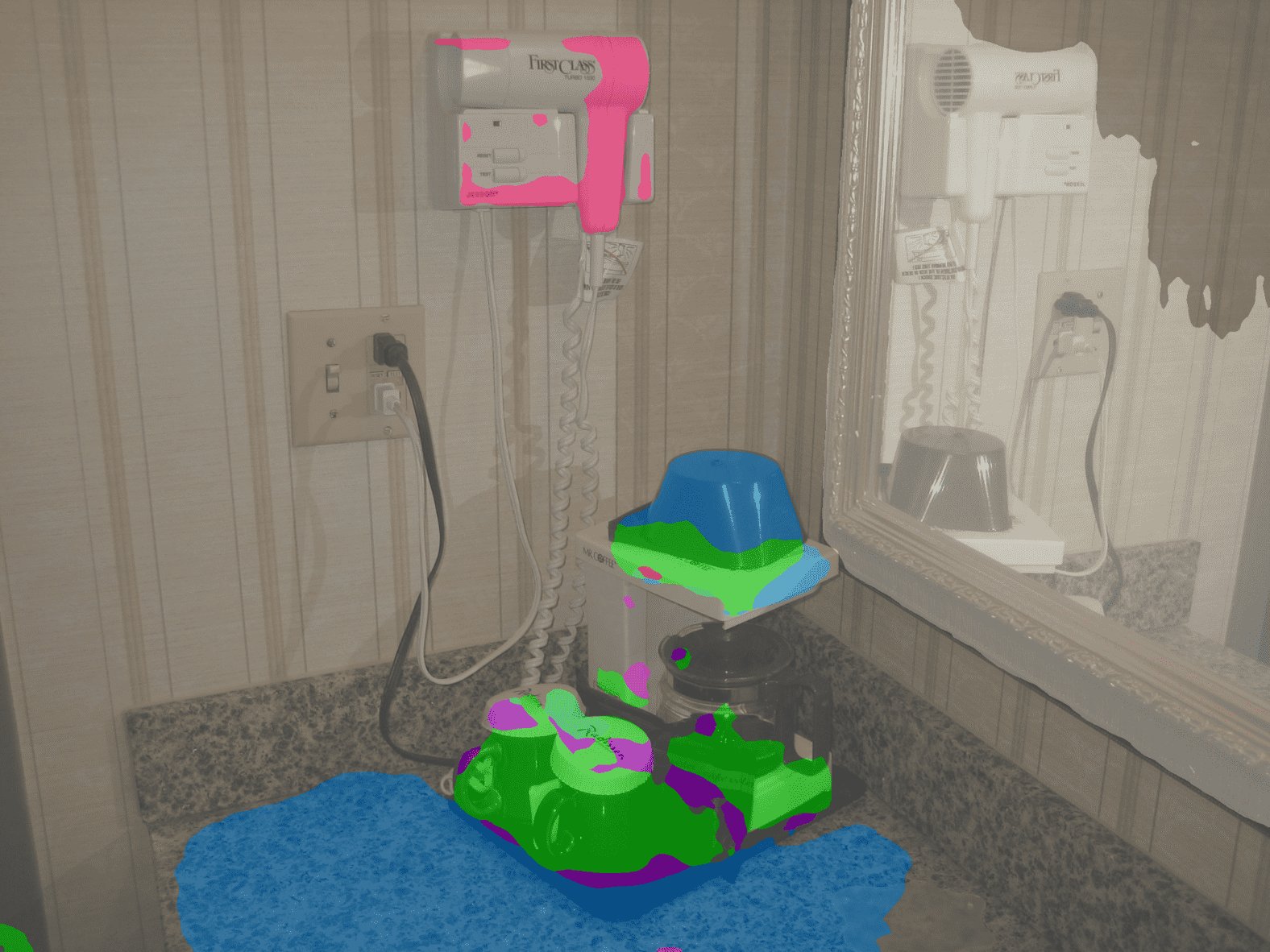}  & \includegraphics[width=0.135\textwidth]{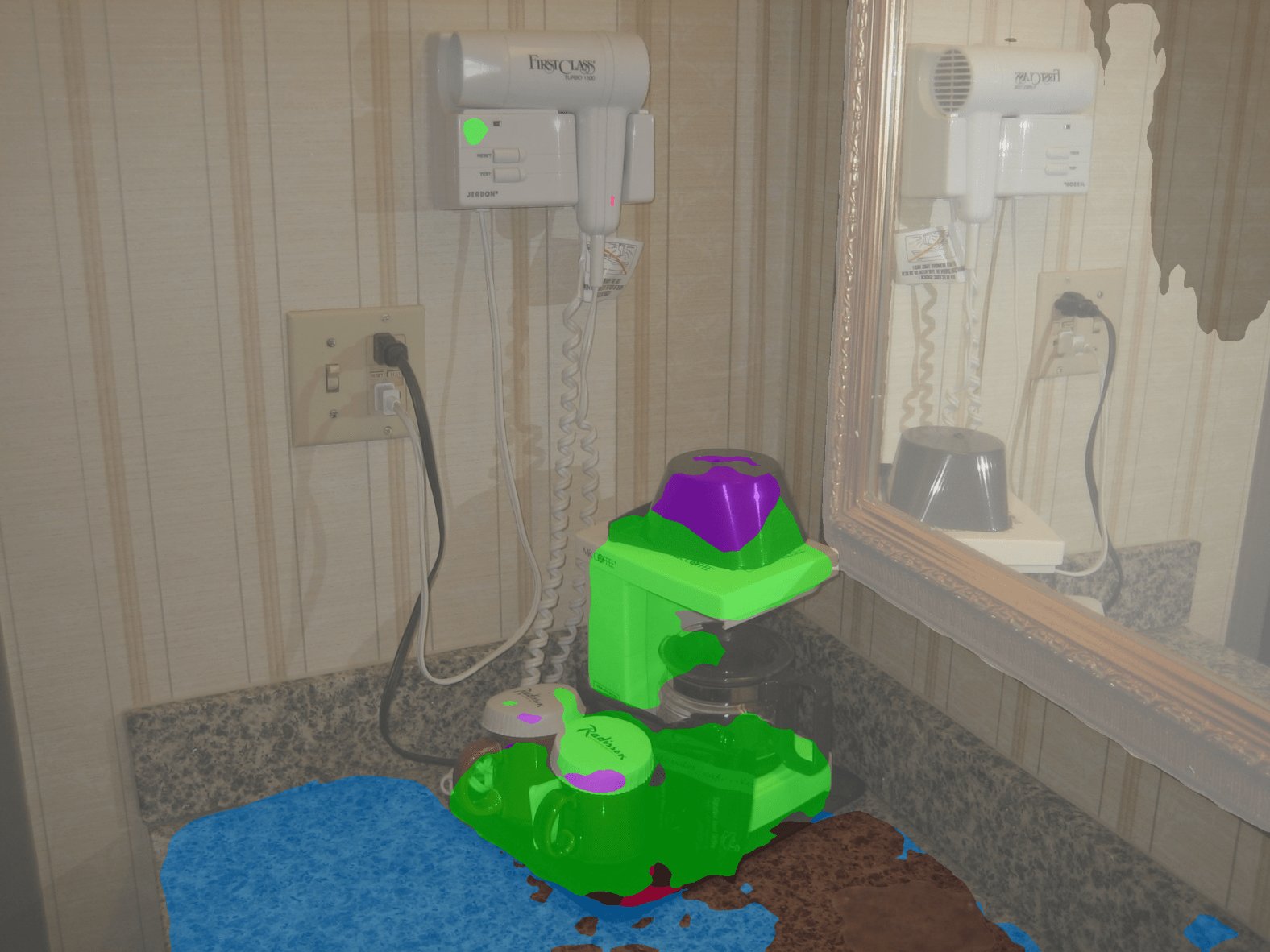} & \includegraphics[width=0.135\textwidth]{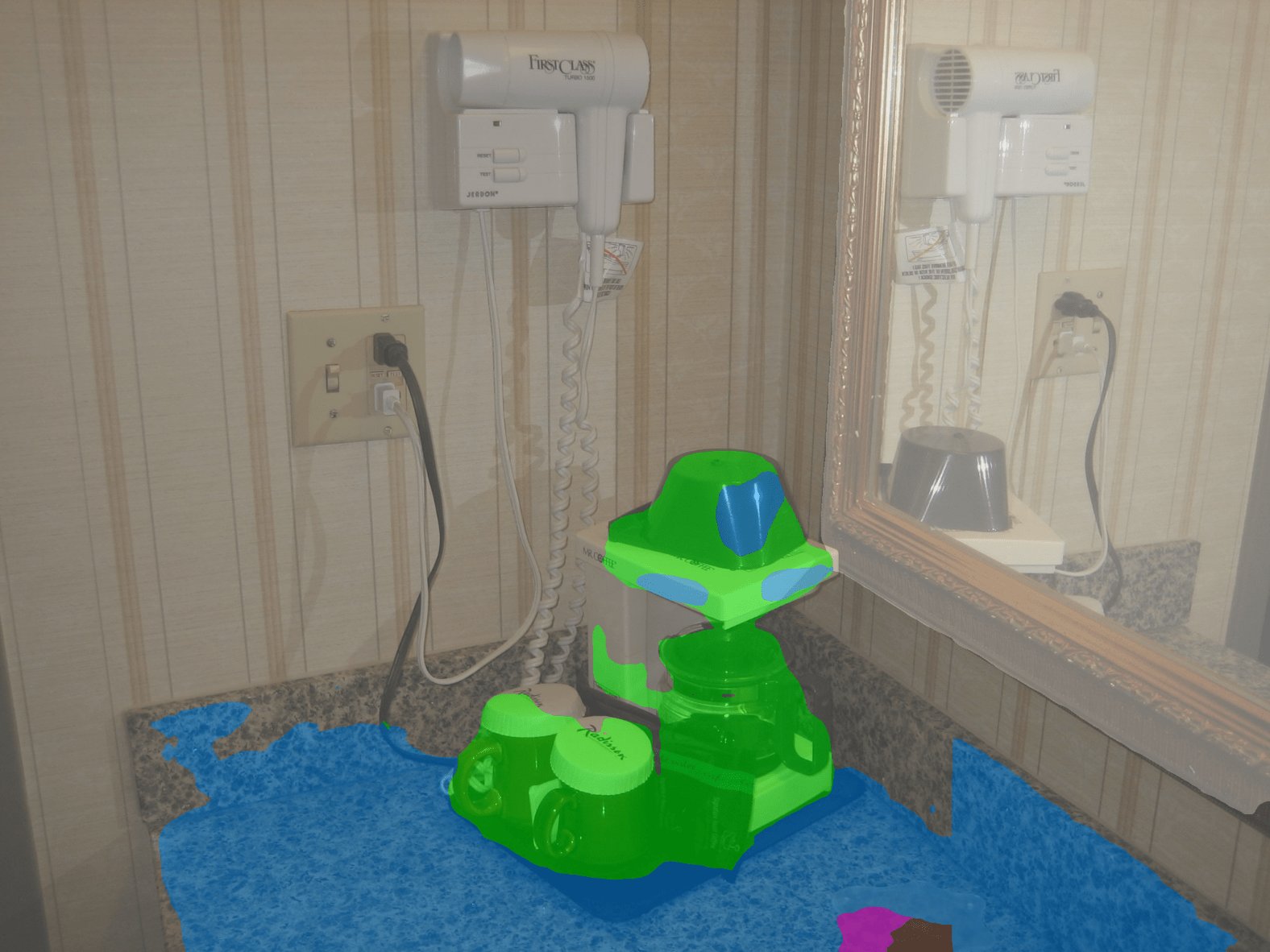} \\

\includegraphics[width=0.135\textwidth]{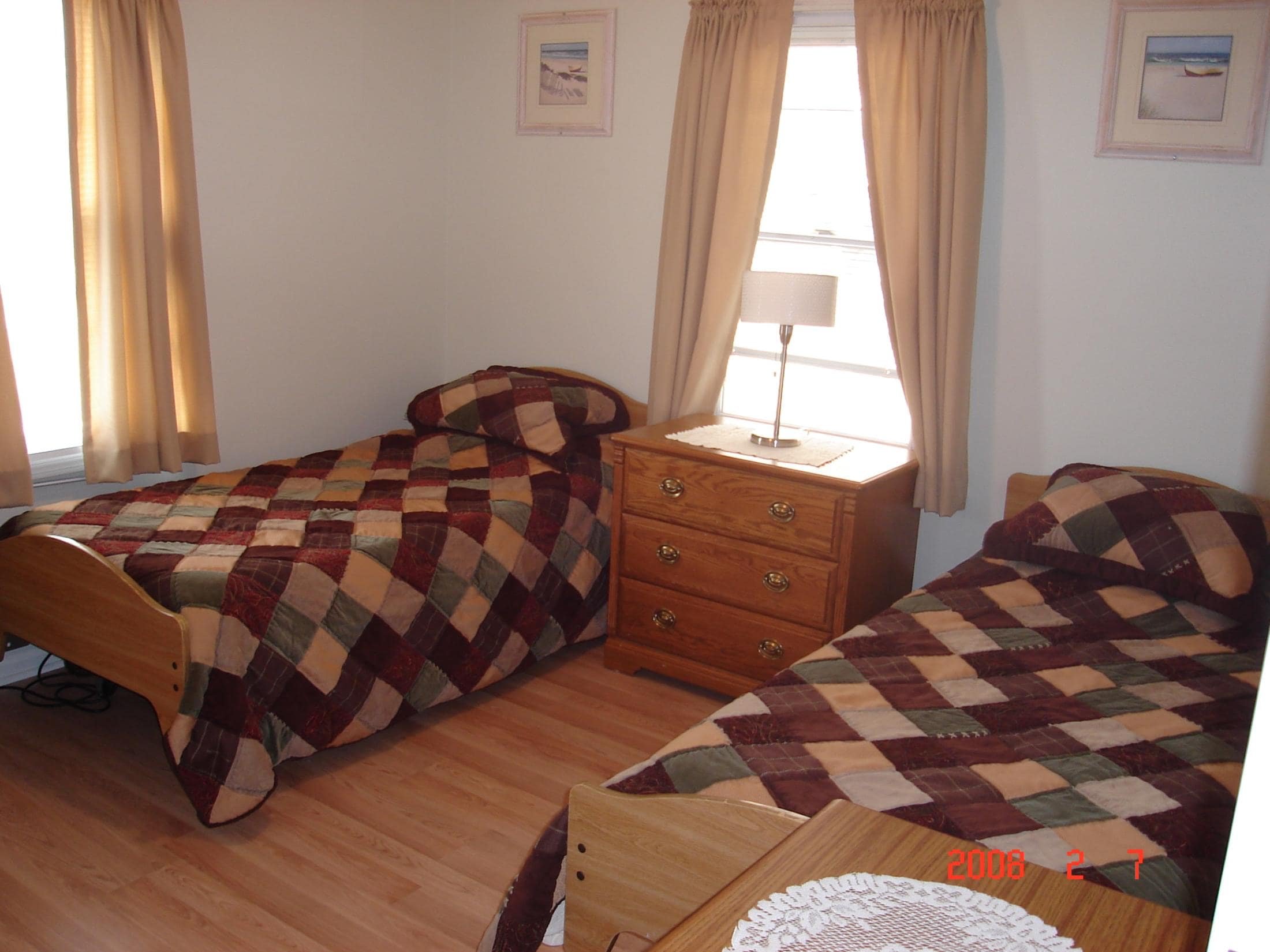} & \includegraphics[width=0.135\textwidth]{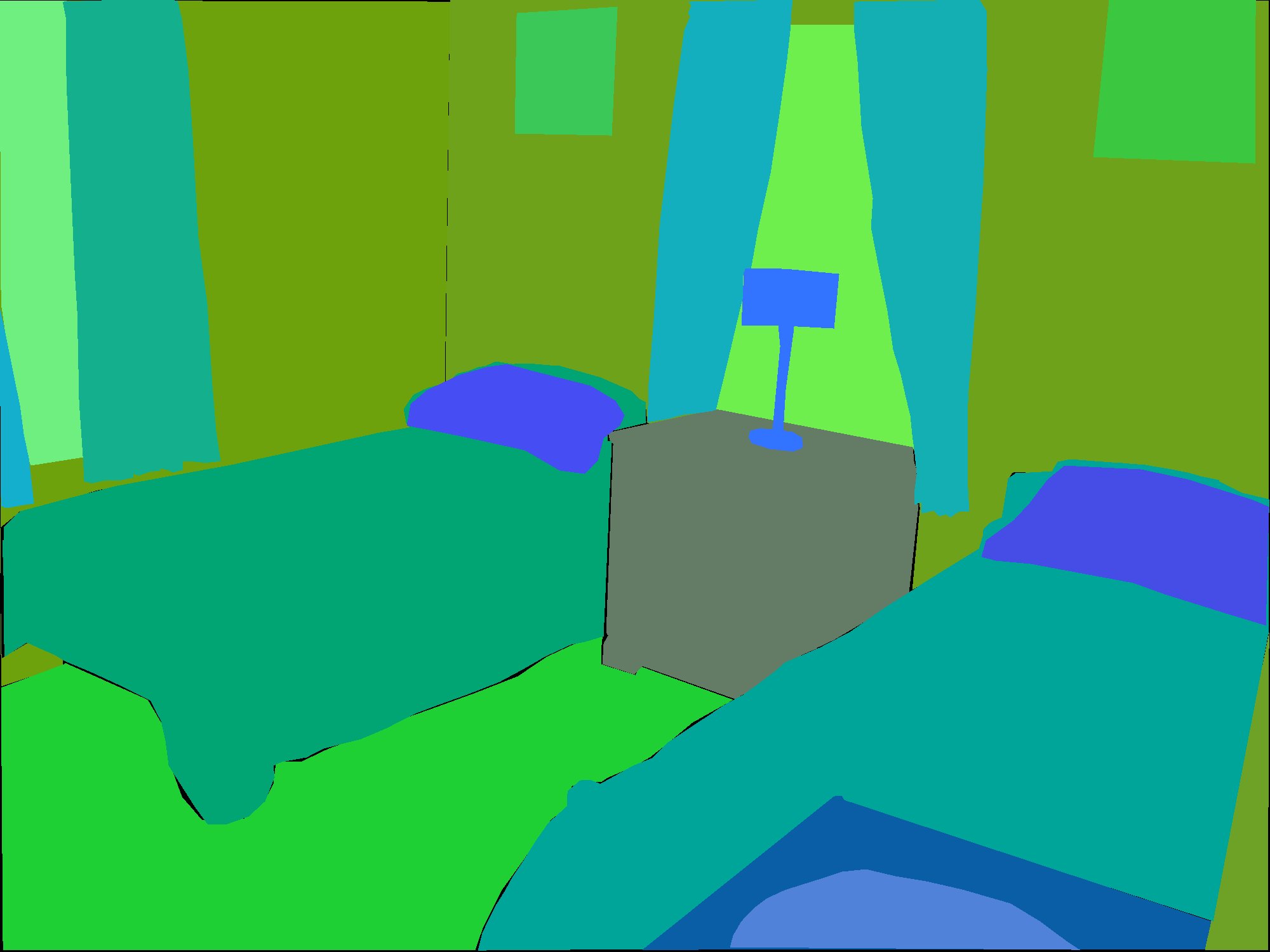} & \includegraphics[width=0.135\textwidth]{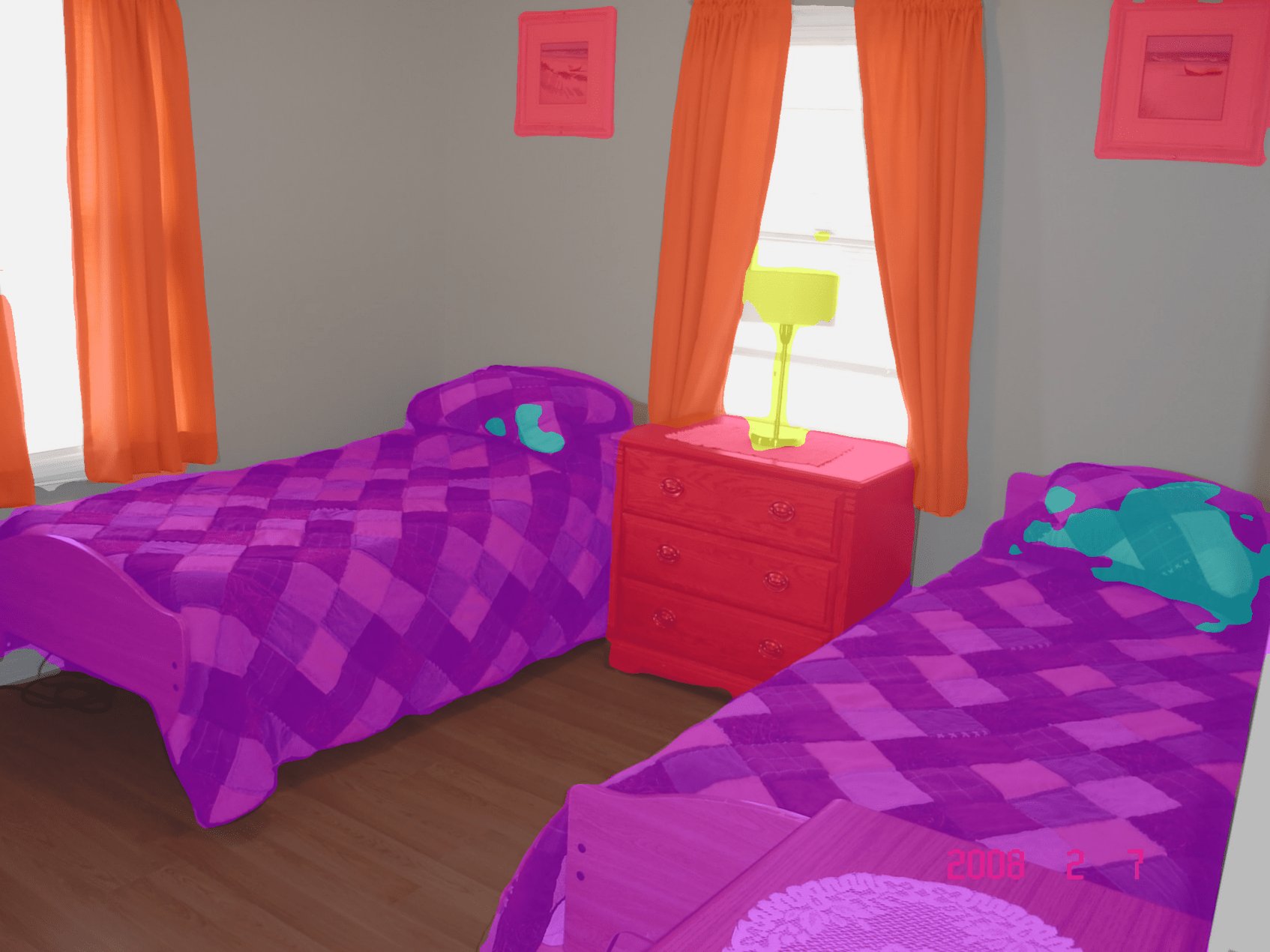} & \includegraphics[width=0.135\textwidth]{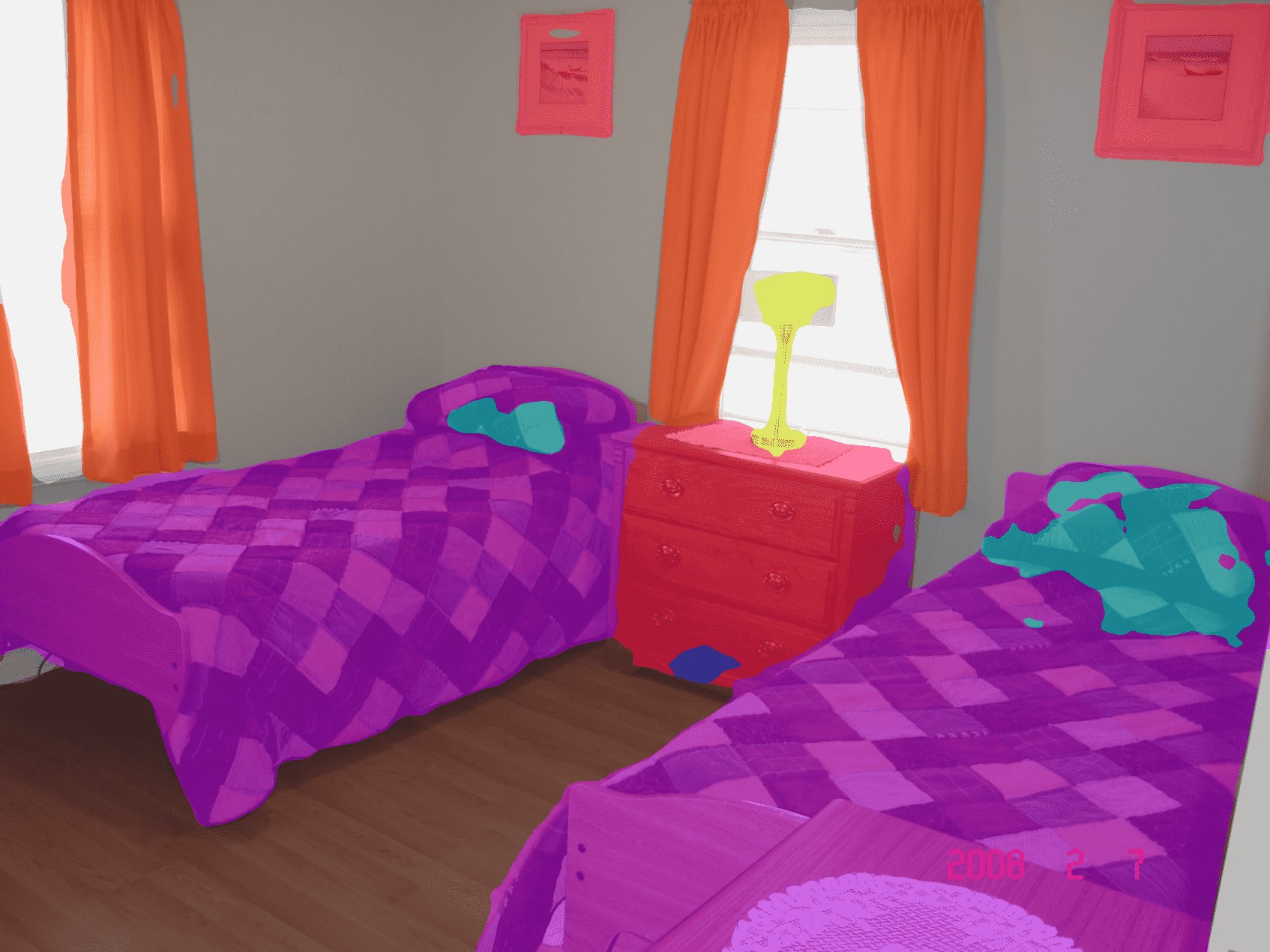}  & \includegraphics[width=0.135\textwidth]{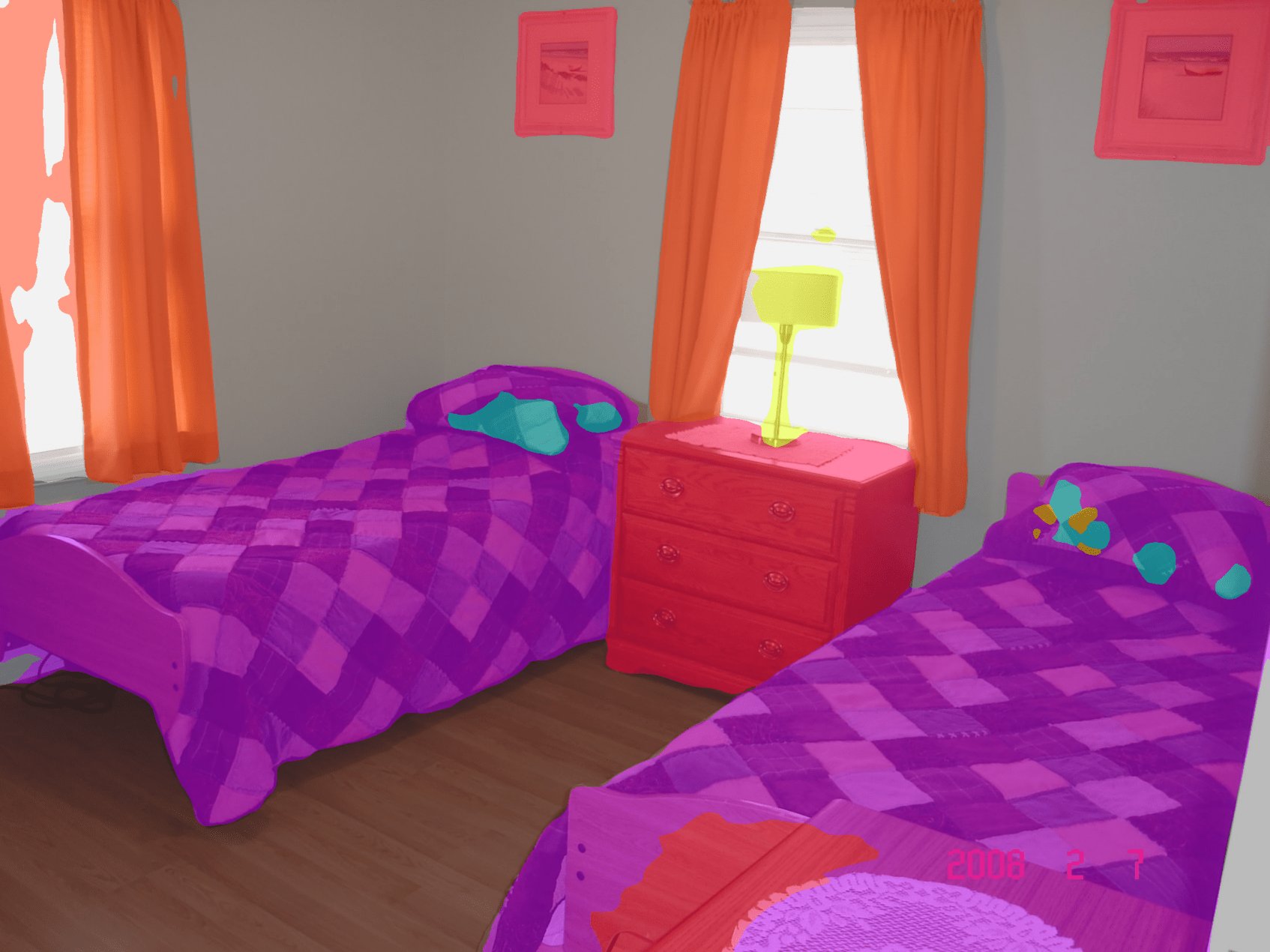} & \includegraphics[width=0.135\textwidth]{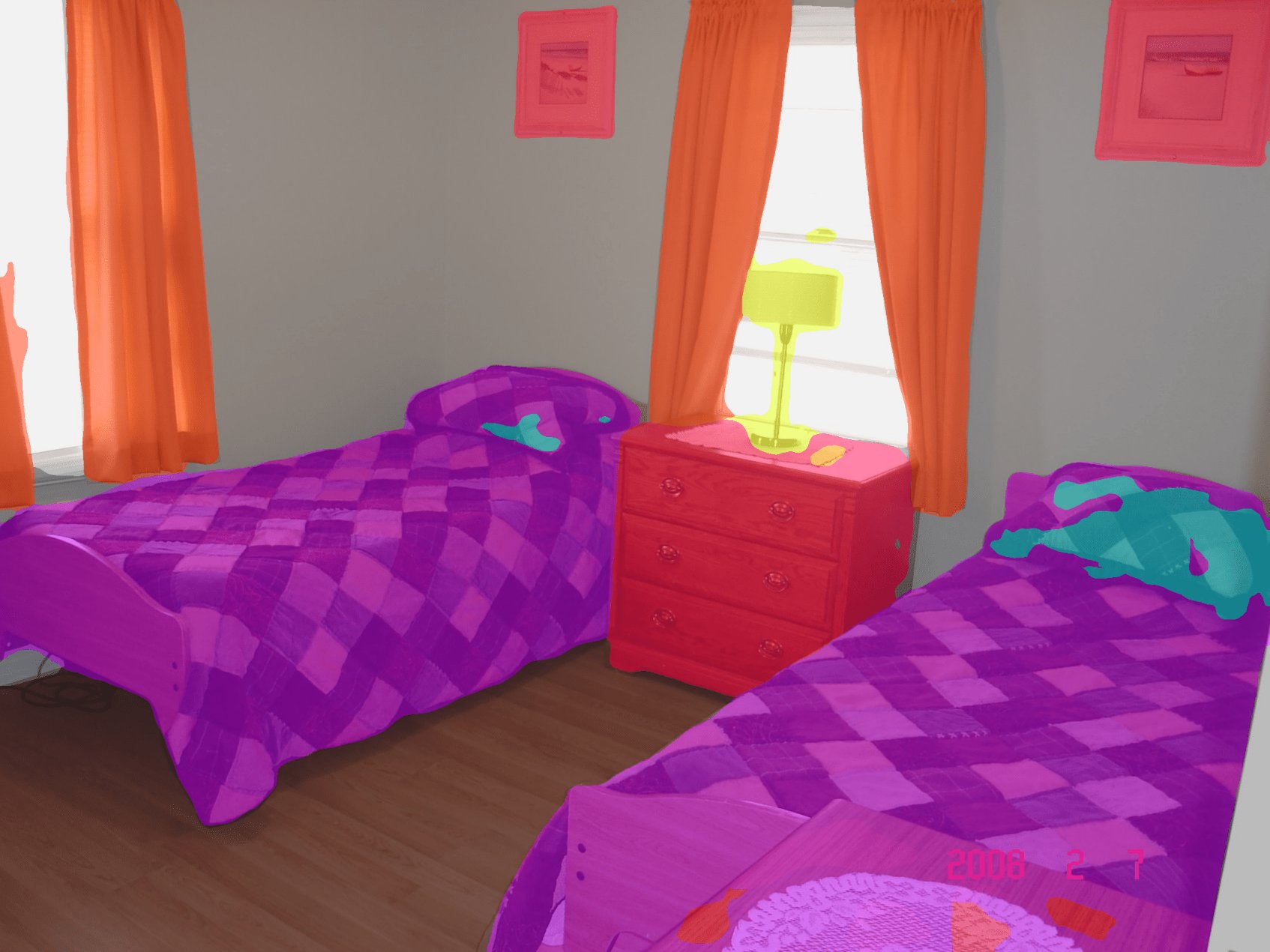} \\

\includegraphics[width=0.135\textwidth]{image_files/semantic_segmentation/raw/ADE_val_00001131.jpg} & \includegraphics[width=0.135\textwidth]{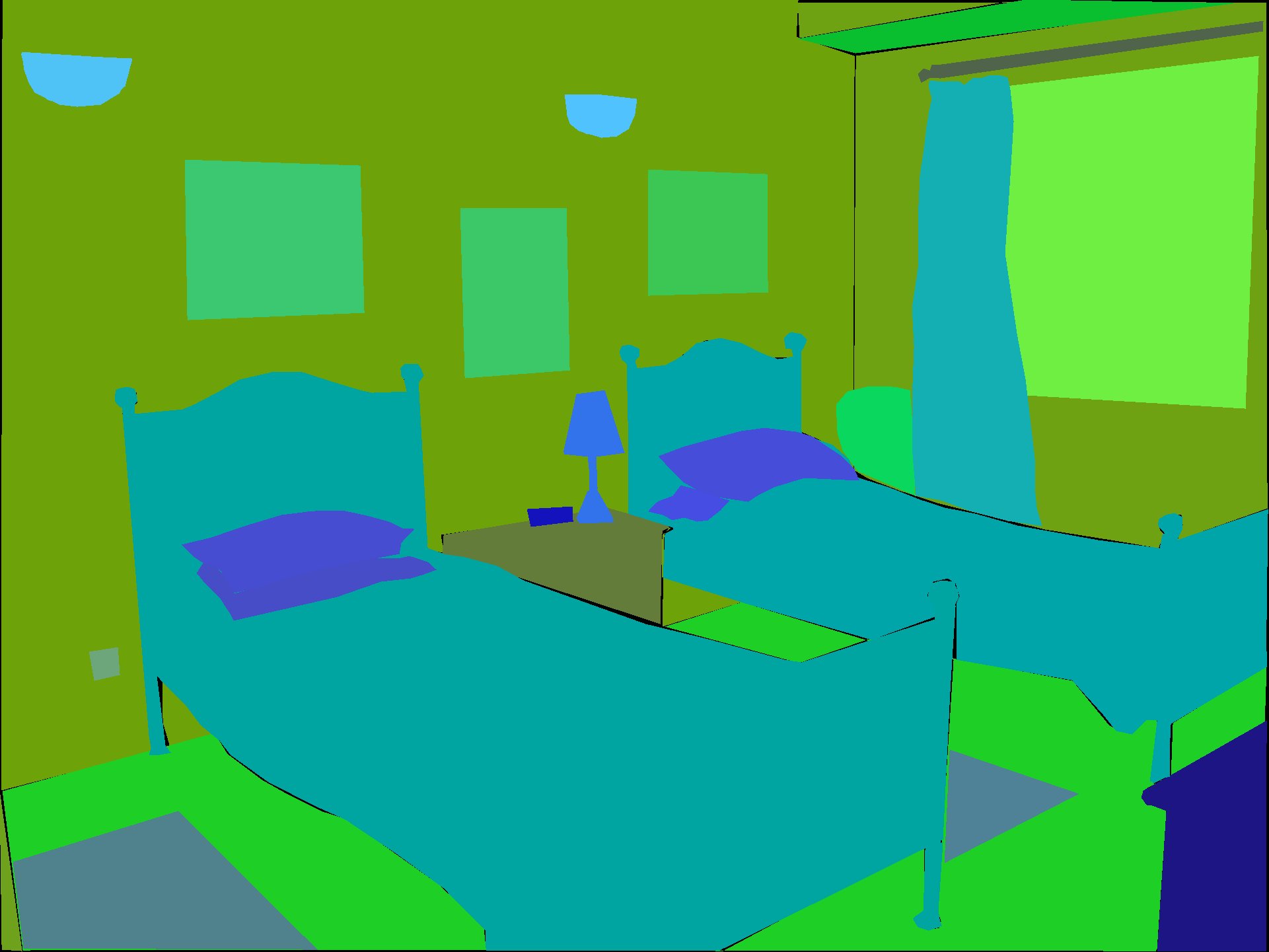} & \includegraphics[width=0.135\textwidth]{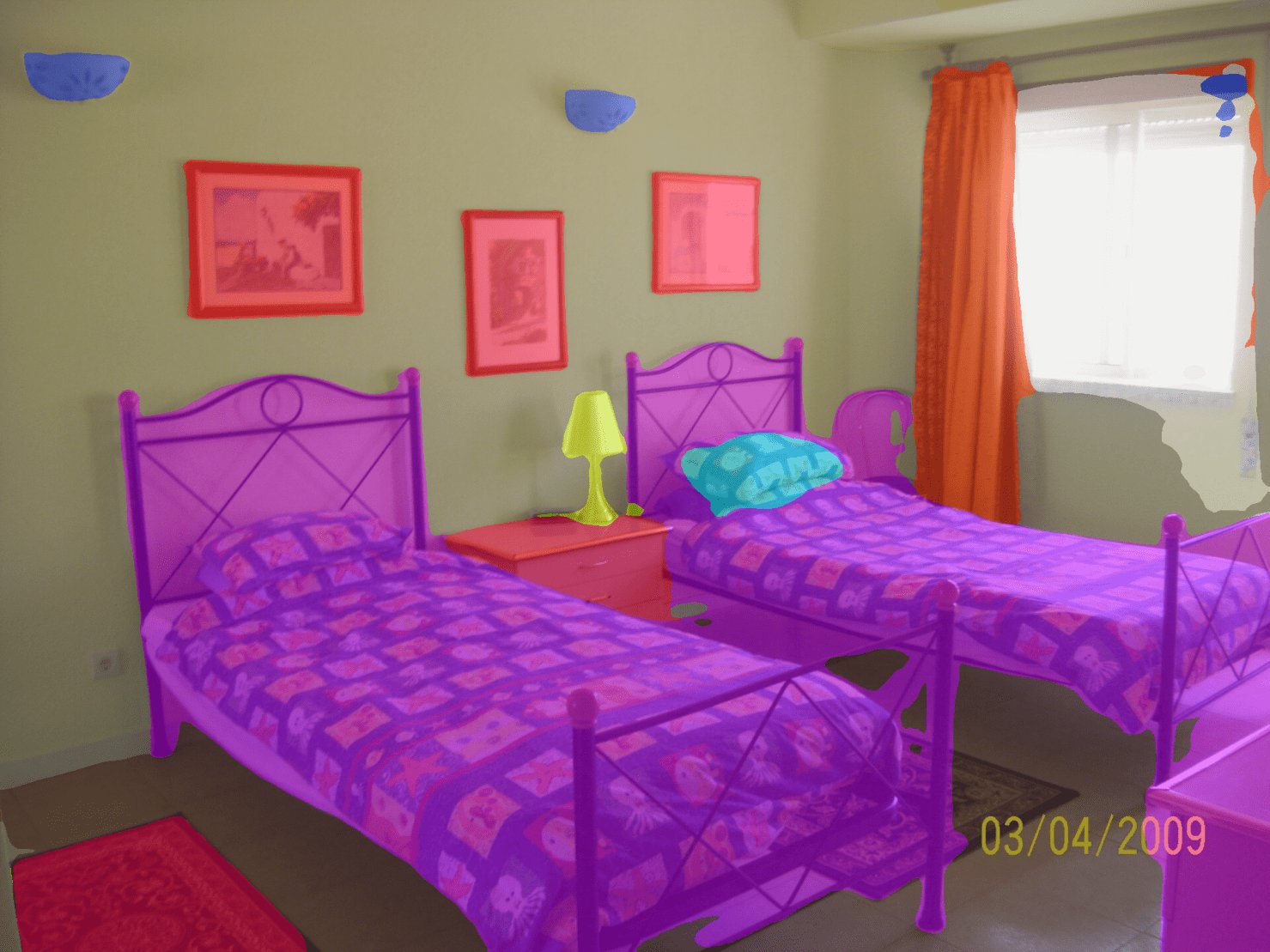} & \includegraphics[width=0.135\textwidth]{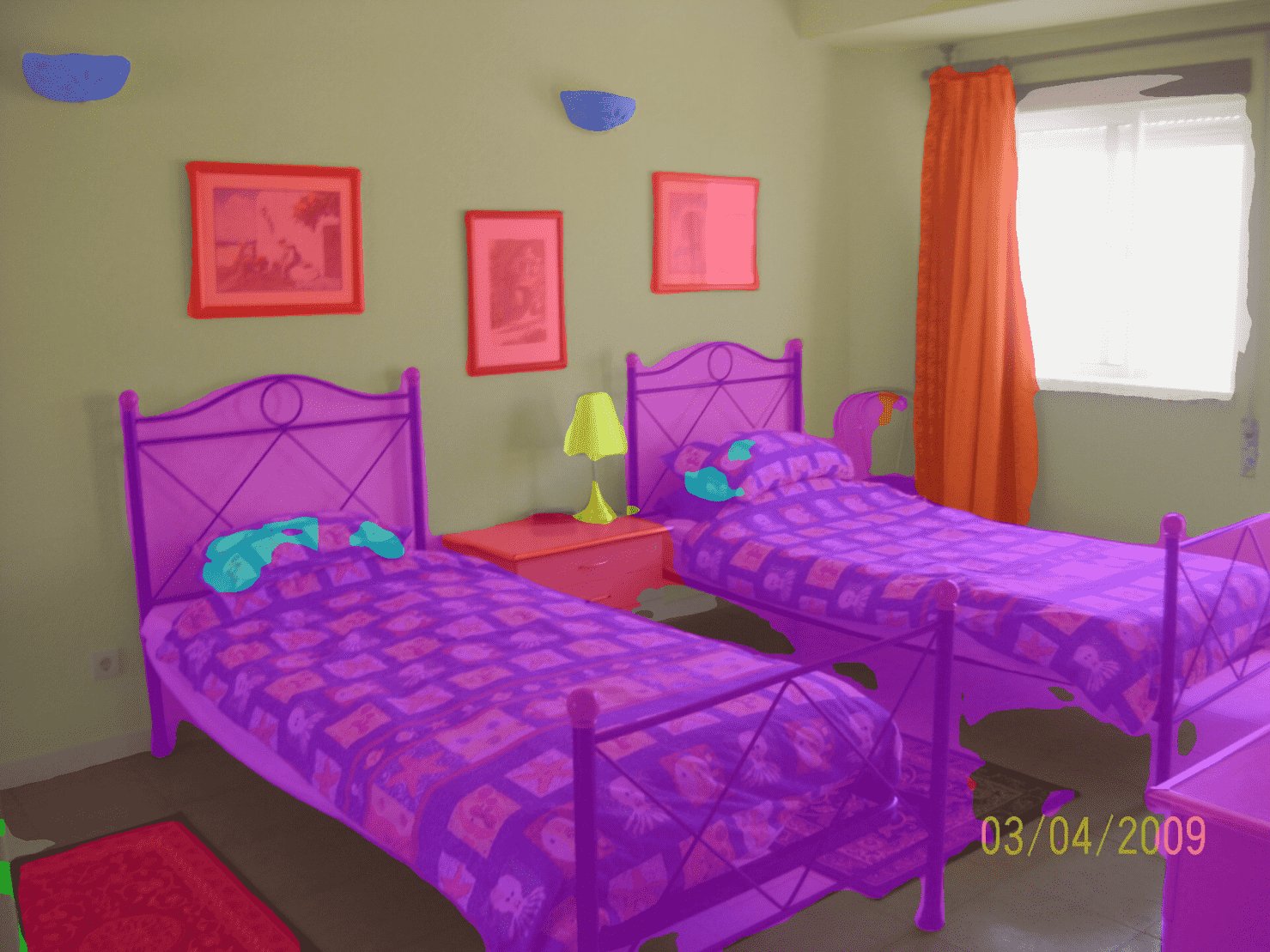}  & \includegraphics[width=0.135\textwidth]{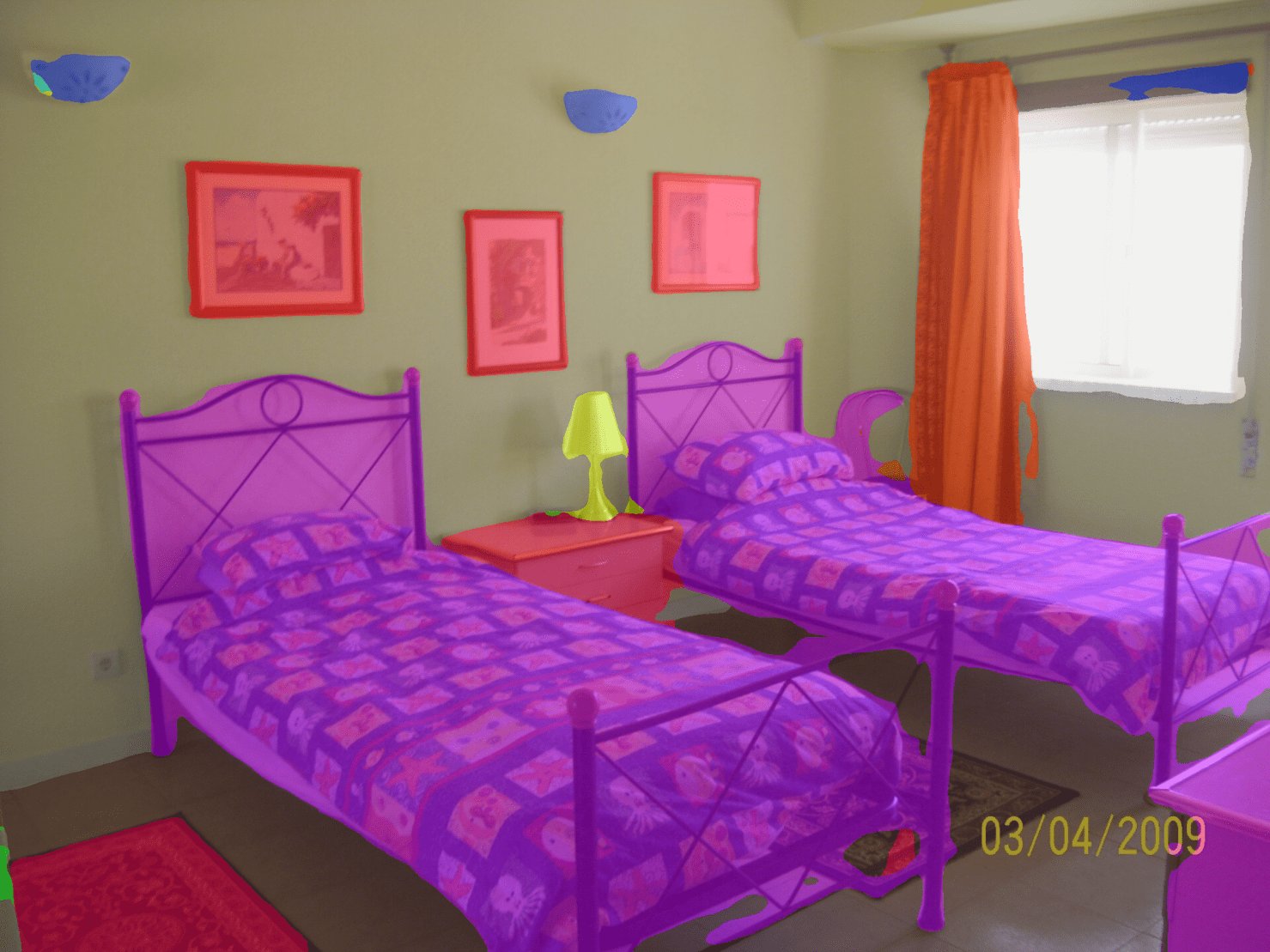} & \includegraphics[width=0.135\textwidth]{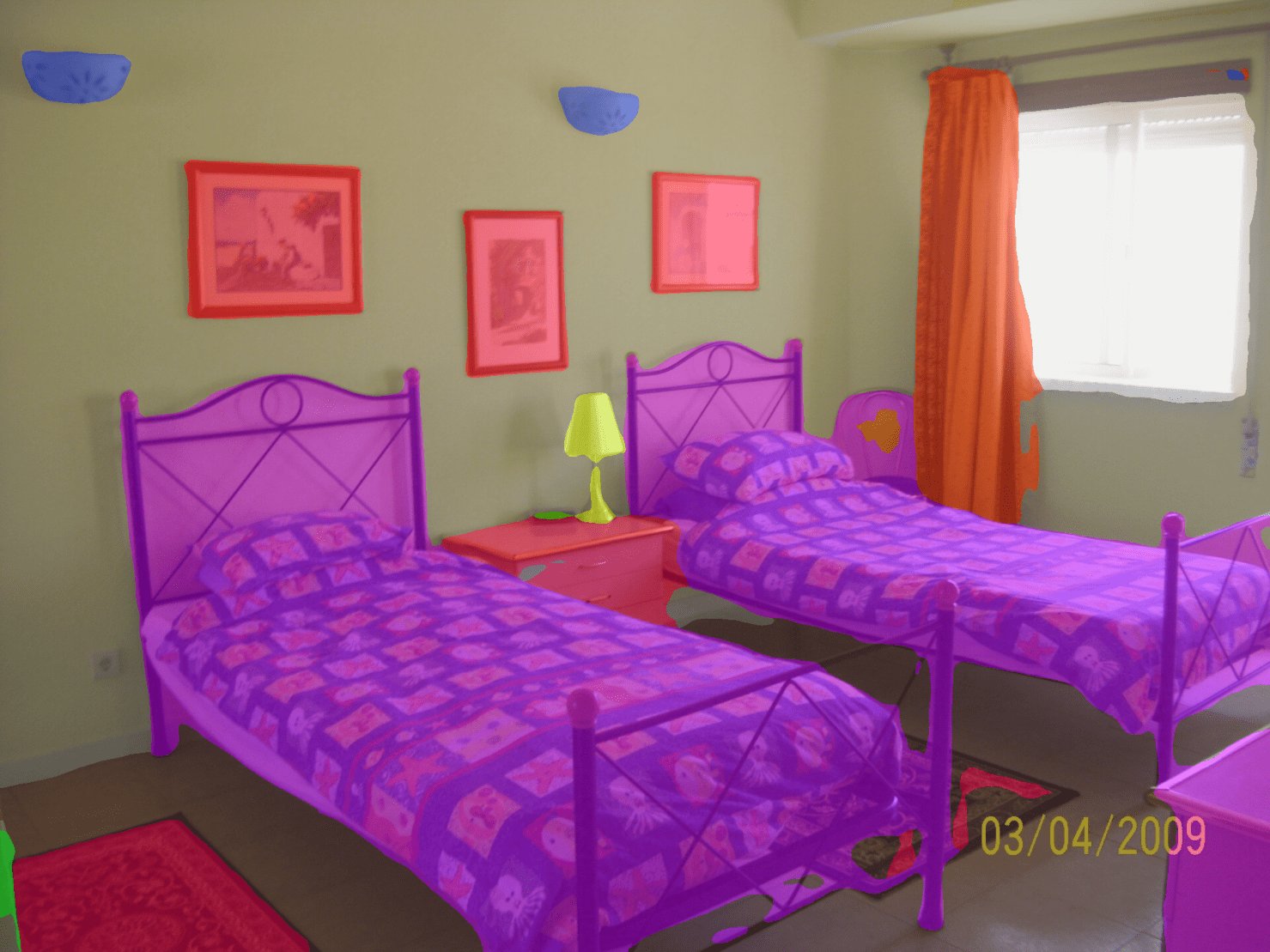} \\

\includegraphics[width=0.135\textwidth]{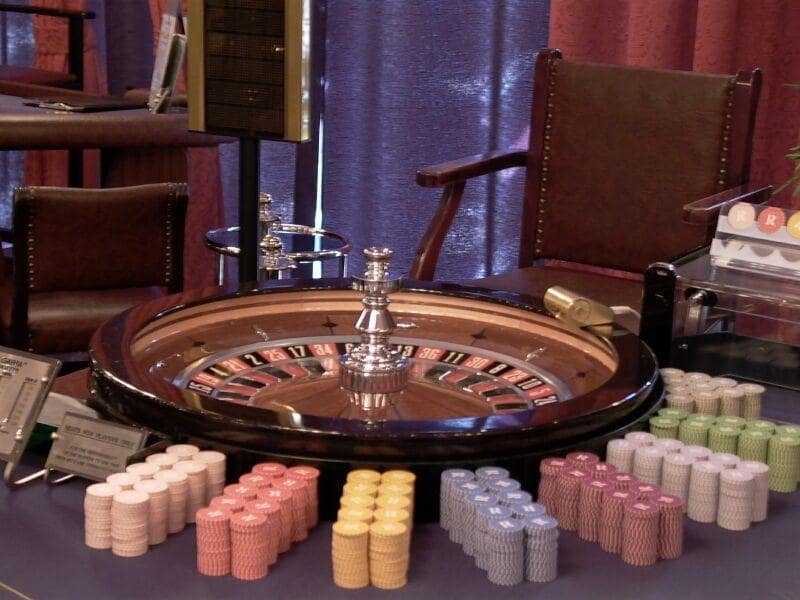} & \includegraphics[width=0.135\textwidth]{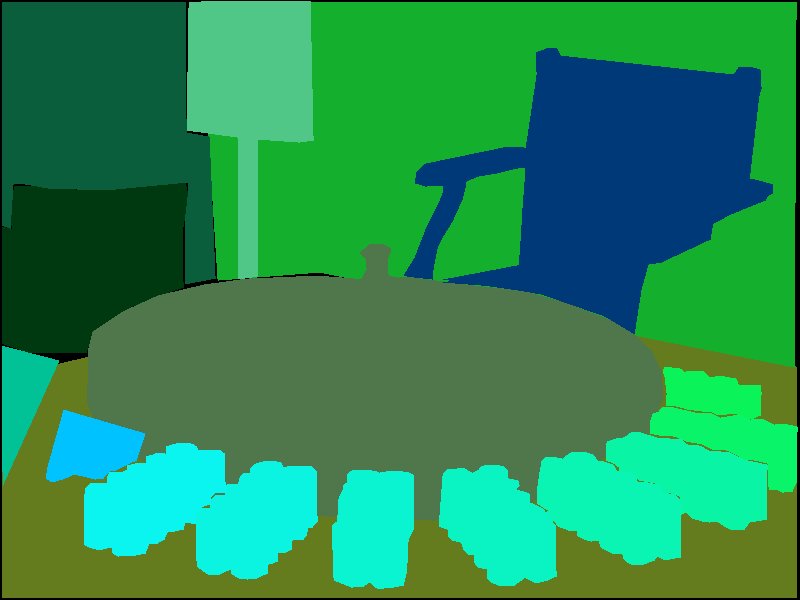} & \includegraphics[width=0.135\textwidth]{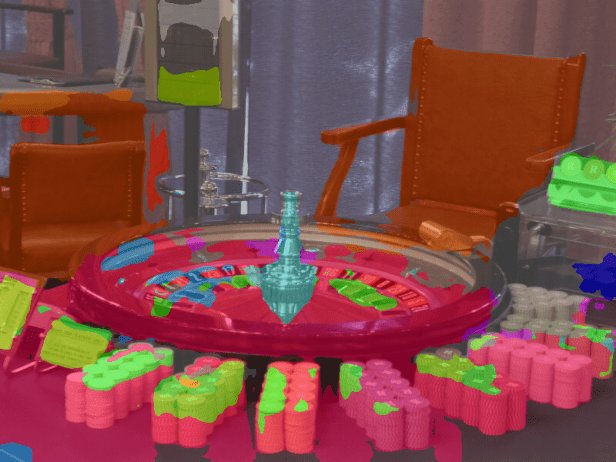} & \includegraphics[width=0.135\textwidth]{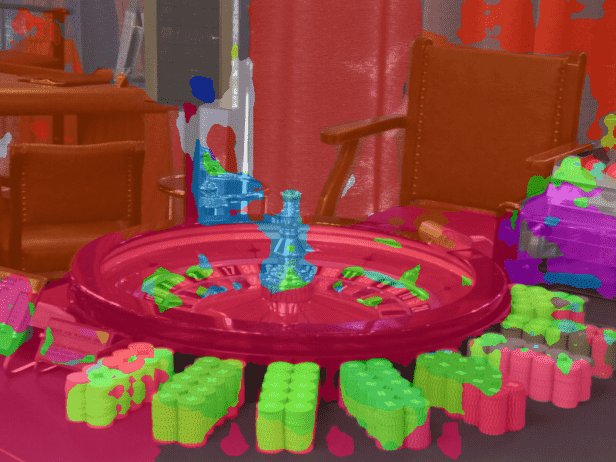}  & \includegraphics[width=0.135\textwidth]{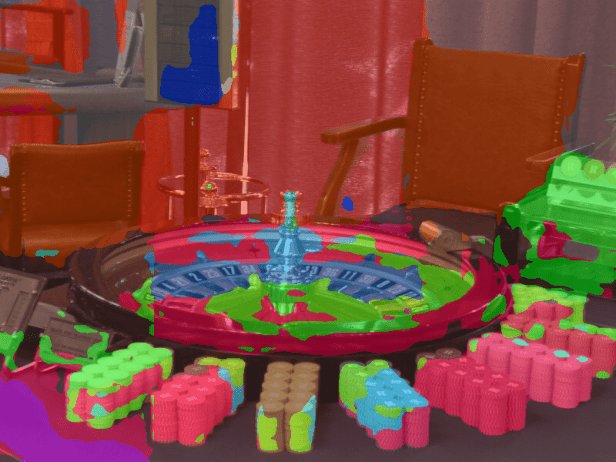} & \includegraphics[width=0.135\textwidth]{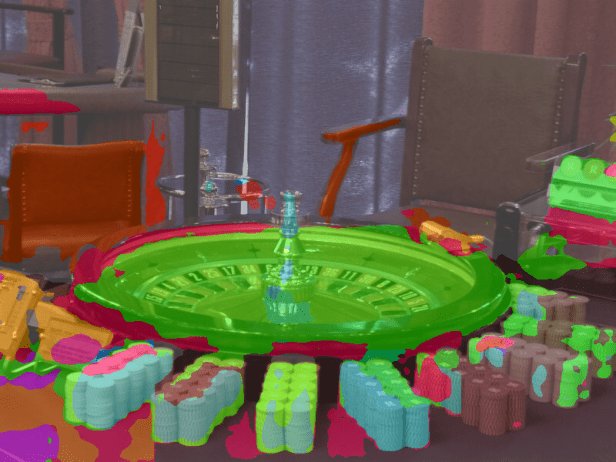} \\

\hline
\hline

\end{tabular}
\caption{Visual comparison of segmentation results using the UPerNet decoder across different models. The comparison includes SLIP, A-CLIP, DetailCLIP at 25 epochs, and DetailCLIP at 50 epochs.}
\label{fig:seg:upernet}
\end{figure*}

\begin{figure*}[ht]
\centering
\begin{tabular}{c|c|cccc}
\hline
\textbf{Image} & \textbf{Ground Truth} & \textbf{A-Clip} & \textbf{SLIP} & \textbf{DetailCLIP} \\ \hline

\includegraphics[width=0.145\textwidth]{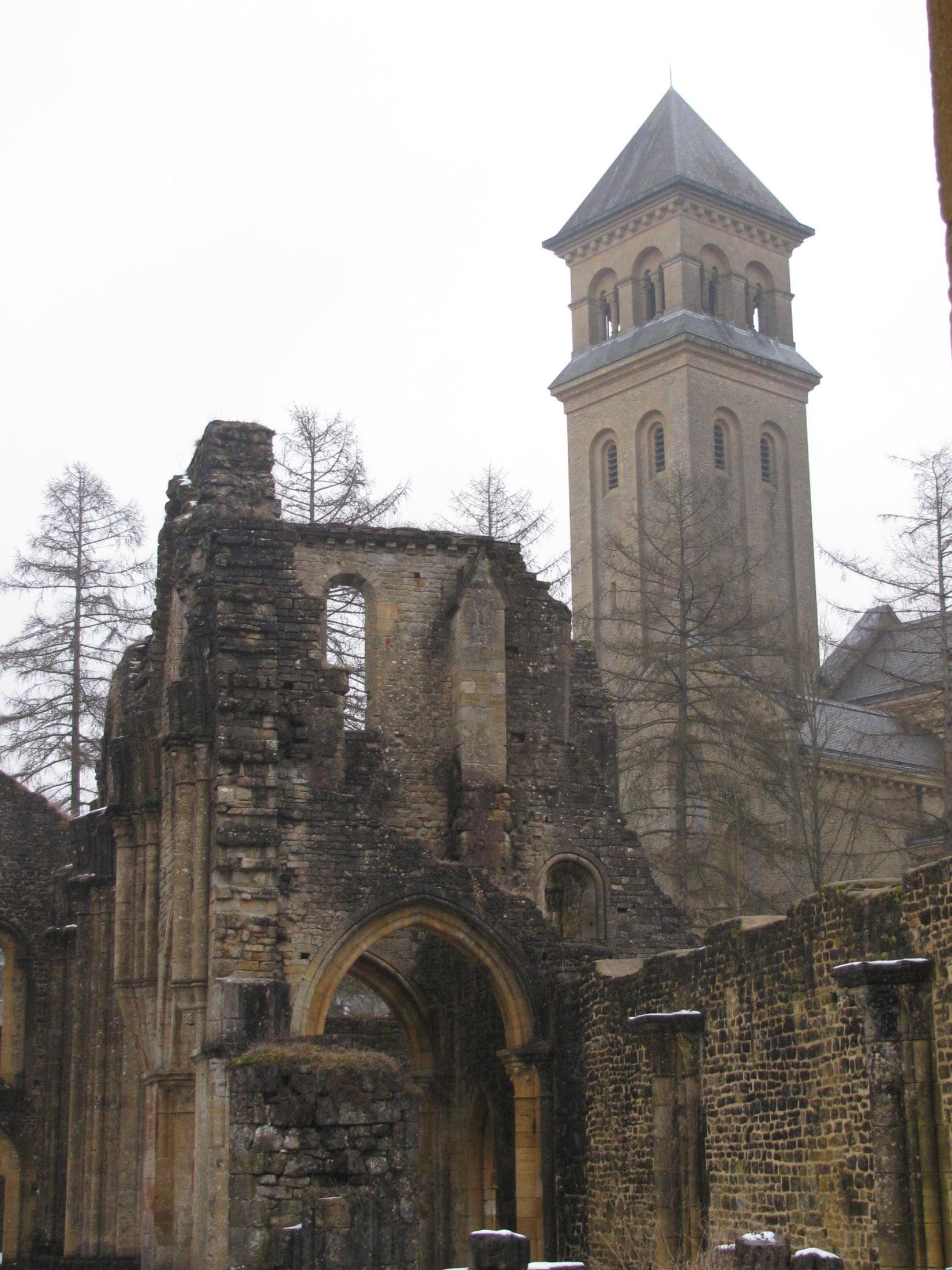} & \includegraphics[width=0.145\textwidth]{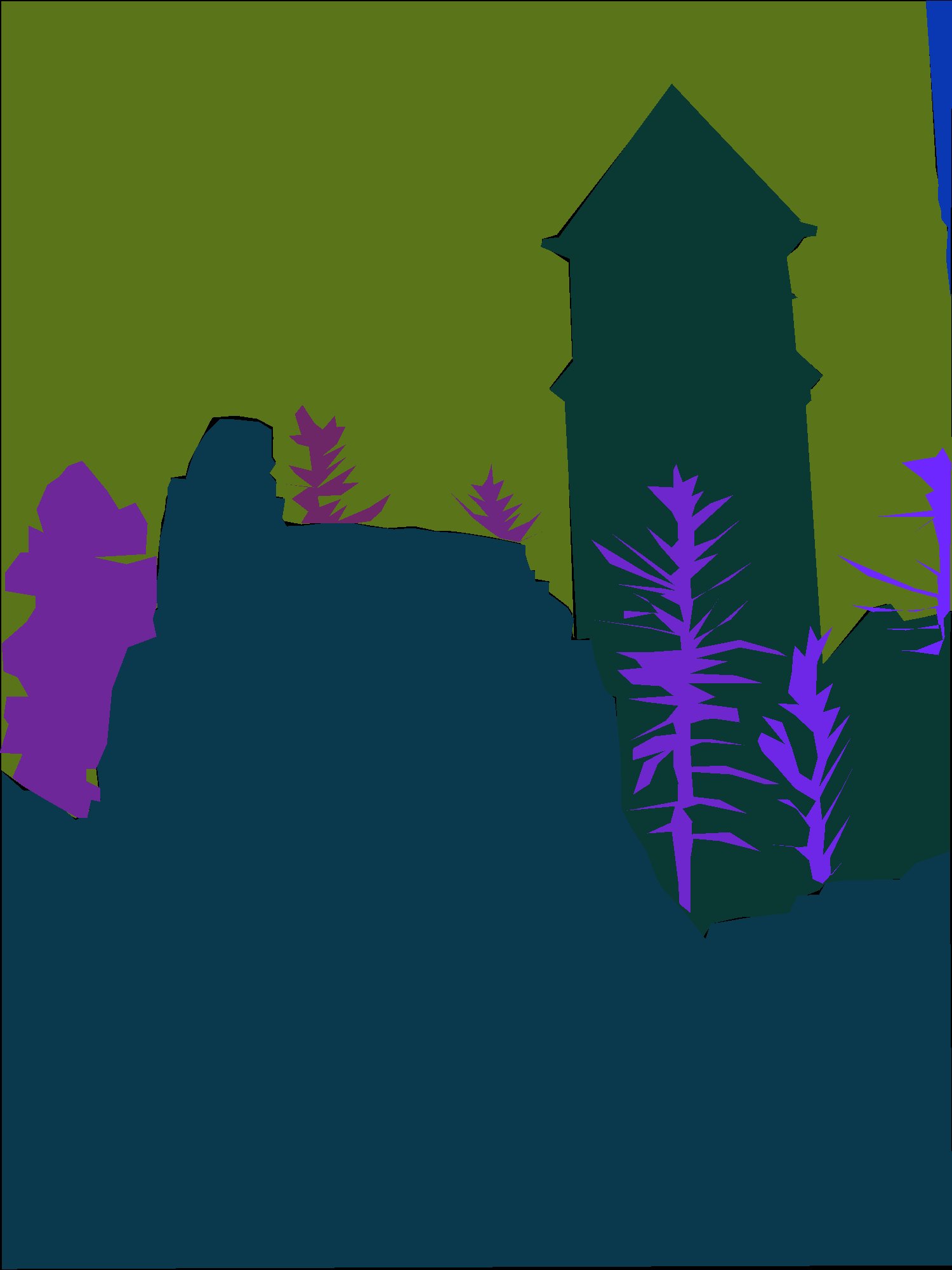} & \includegraphics[width=0.145\textwidth]{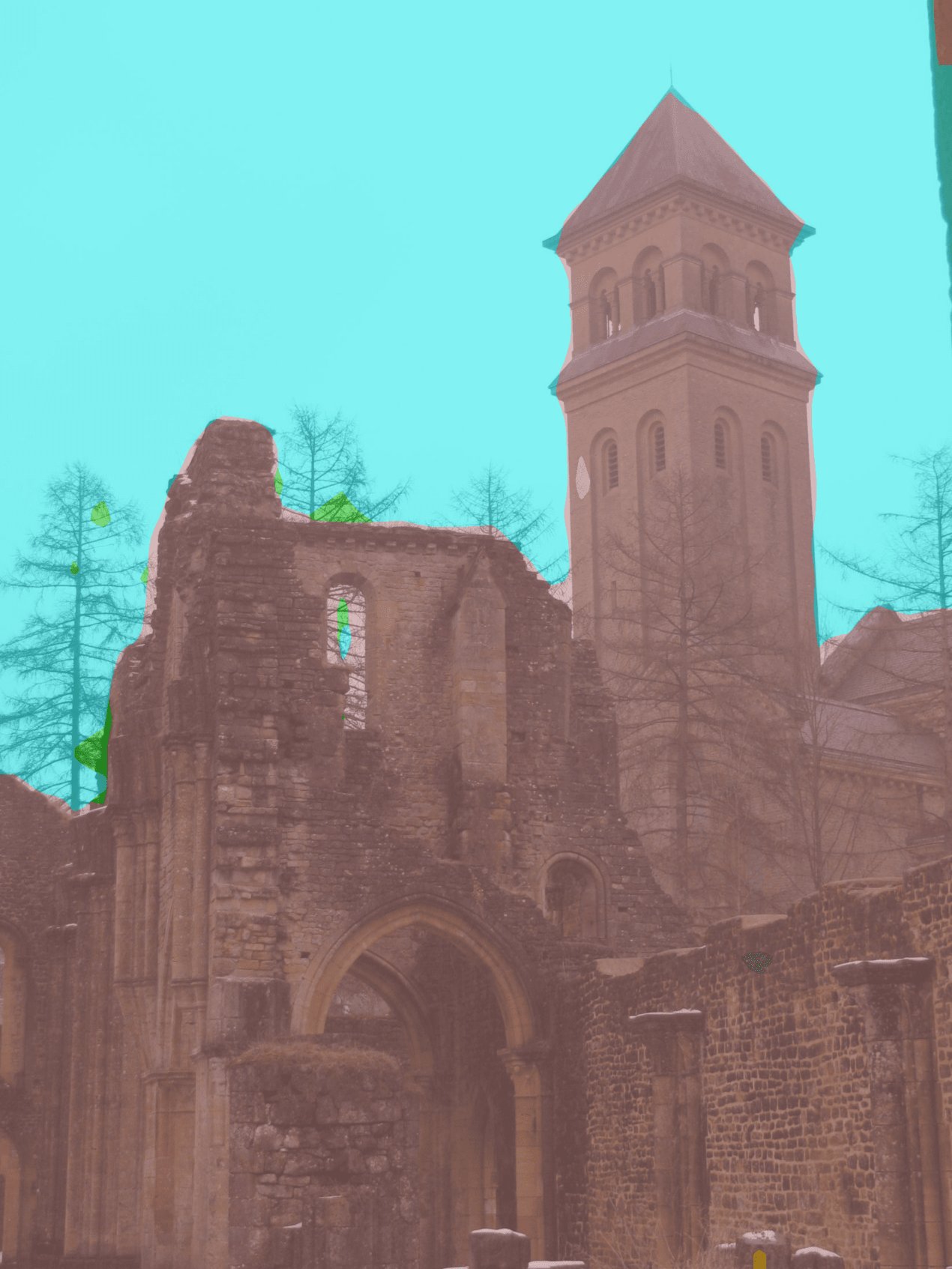} & \includegraphics[width=0.145\textwidth]{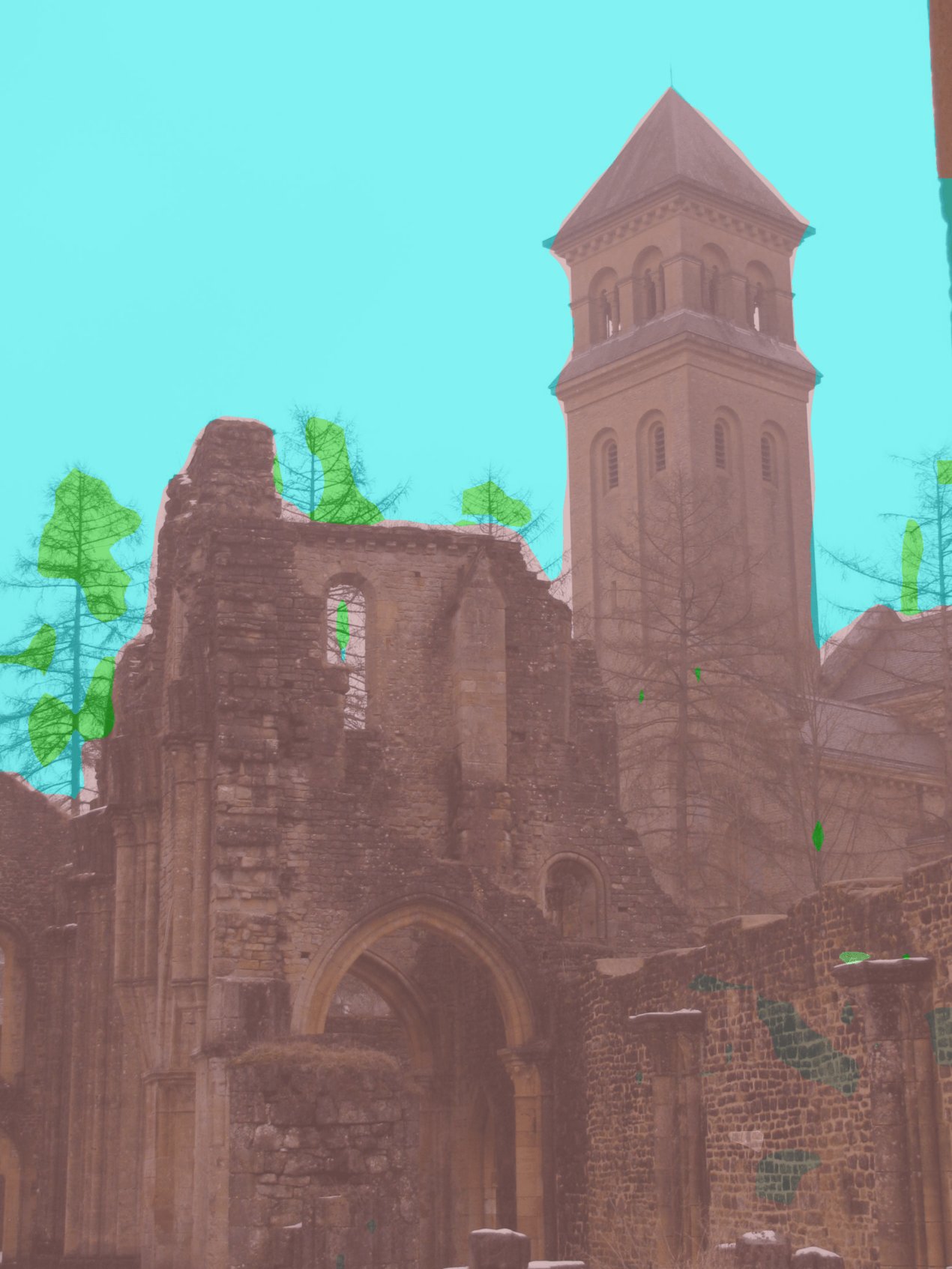}  & \includegraphics[width=0.145\textwidth]{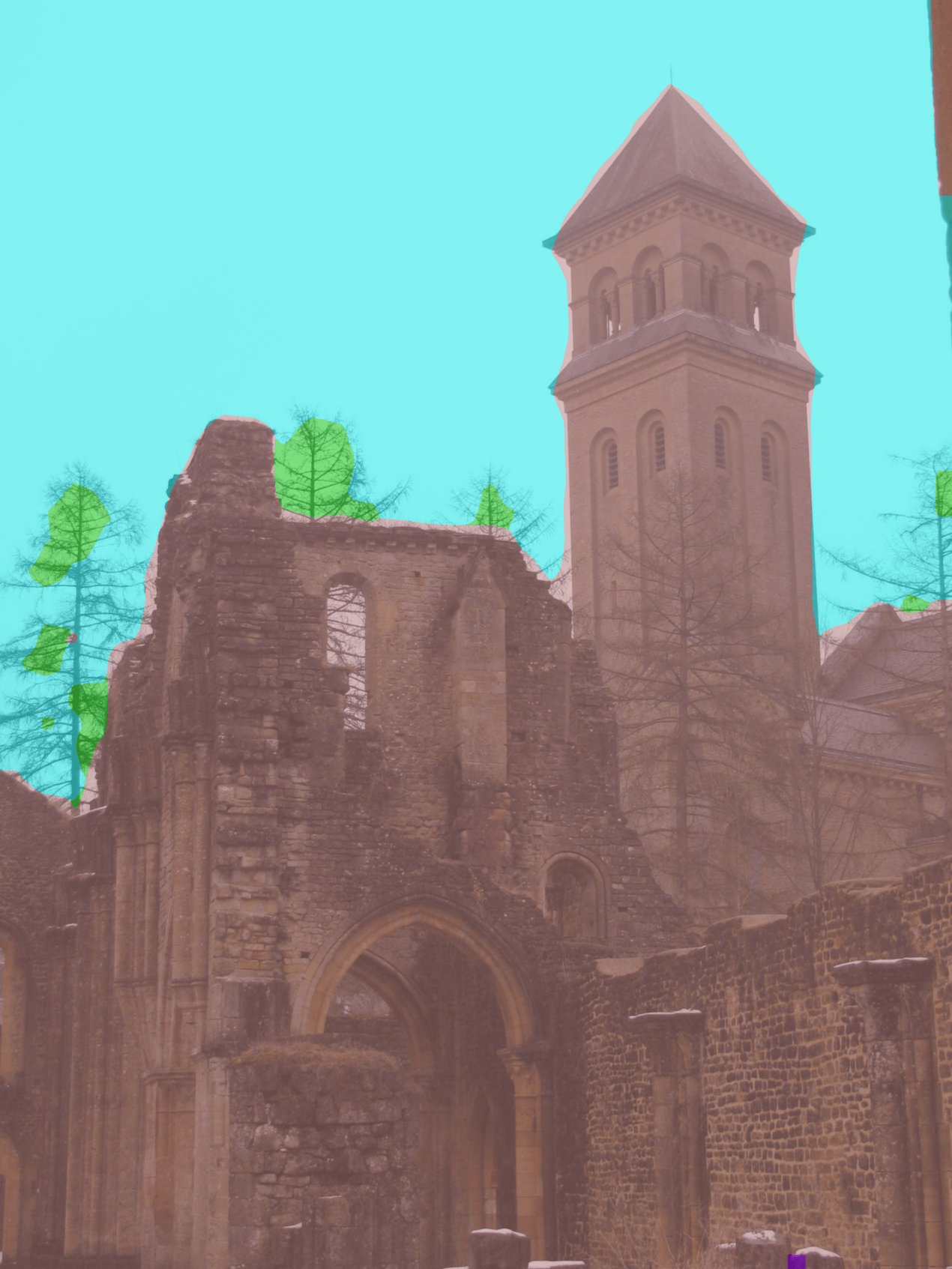}  \\

\includegraphics[width=0.145\textwidth]{image_files/semantic_segmentation/raw/ADE_val_00001005.jpg} & \includegraphics[width=0.145\textwidth]{image_files/semantic_segmentation/ground_truth/ADE_val_00001005_seg.jpg} & \includegraphics[width=0.145\textwidth]{image_files/semantic_segmentation/linear_relevant/aclip_25epoch/ADE_val_00001005_seg.jpg} & \includegraphics[width=0.145\textwidth]{image_files/semantic_segmentation/linear_relevant/slip_25epoch/ADE_val_00001005_seg.jpg}  & \includegraphics[width=0.145\textwidth]{image_files/semantic_segmentation/linear_relevant/iclip_25epoch/ADE_val_00001005_seg.jpg} \\

\includegraphics[width=0.145\textwidth]{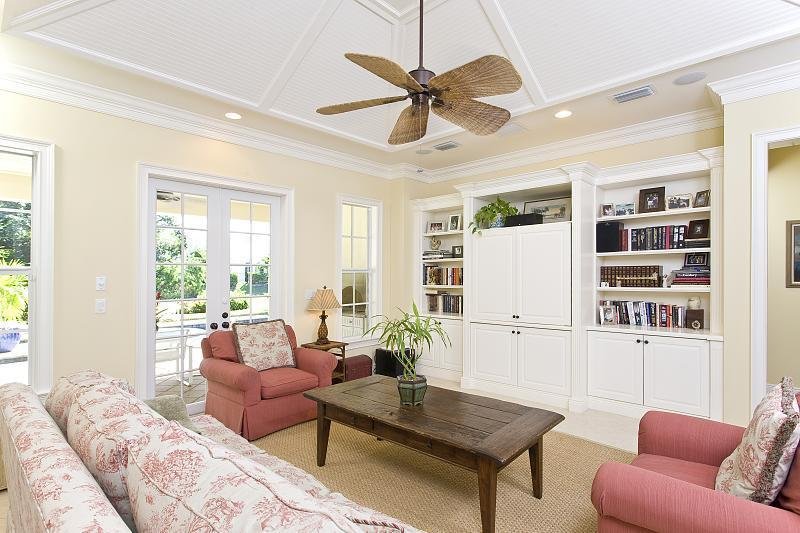} & \includegraphics[width=0.145\textwidth]{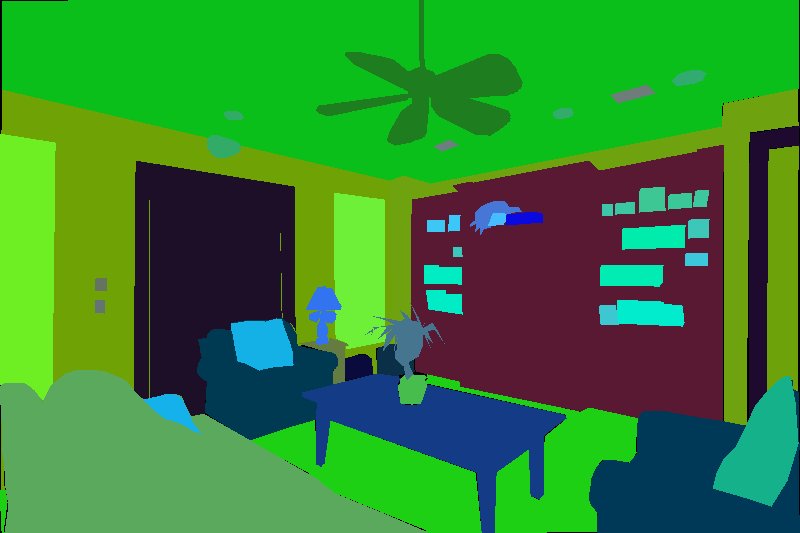} & \includegraphics[width=0.145\textwidth]{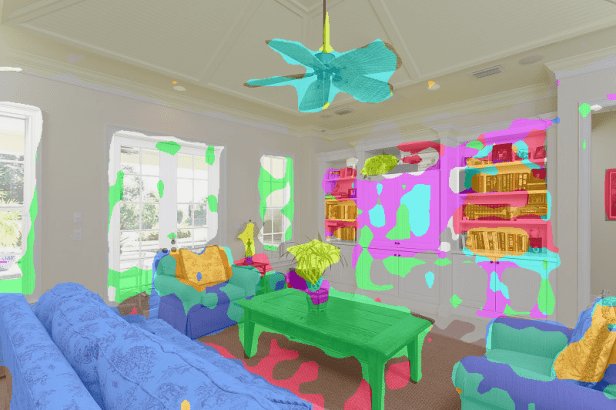} & \includegraphics[width=0.145\textwidth]{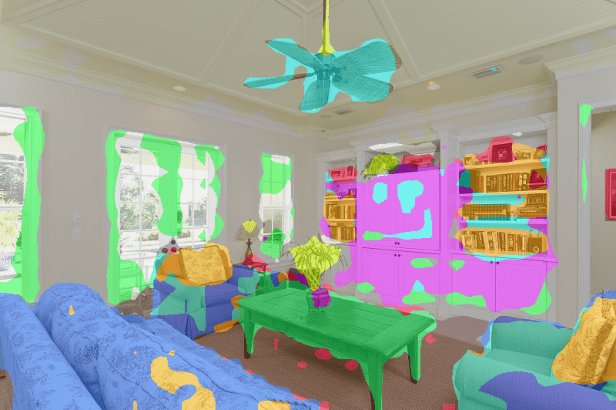}  & \includegraphics[width=0.145\textwidth]{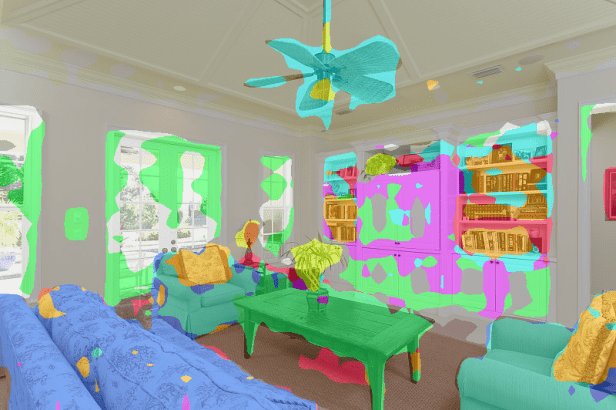} \\

\includegraphics[width=0.145\textwidth]{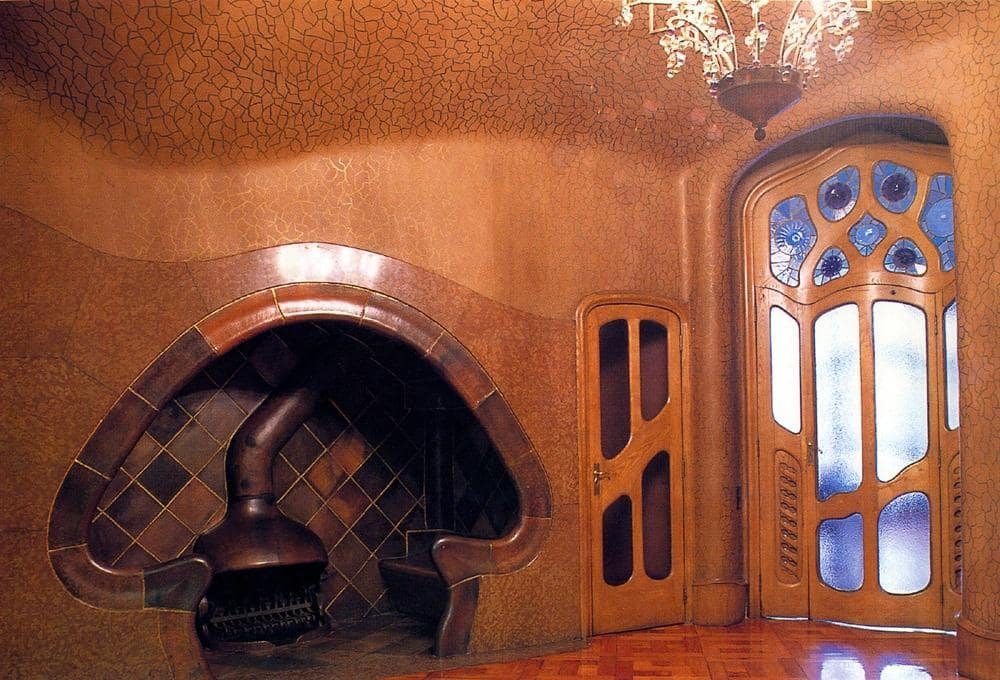} & \includegraphics[width=0.145\textwidth]{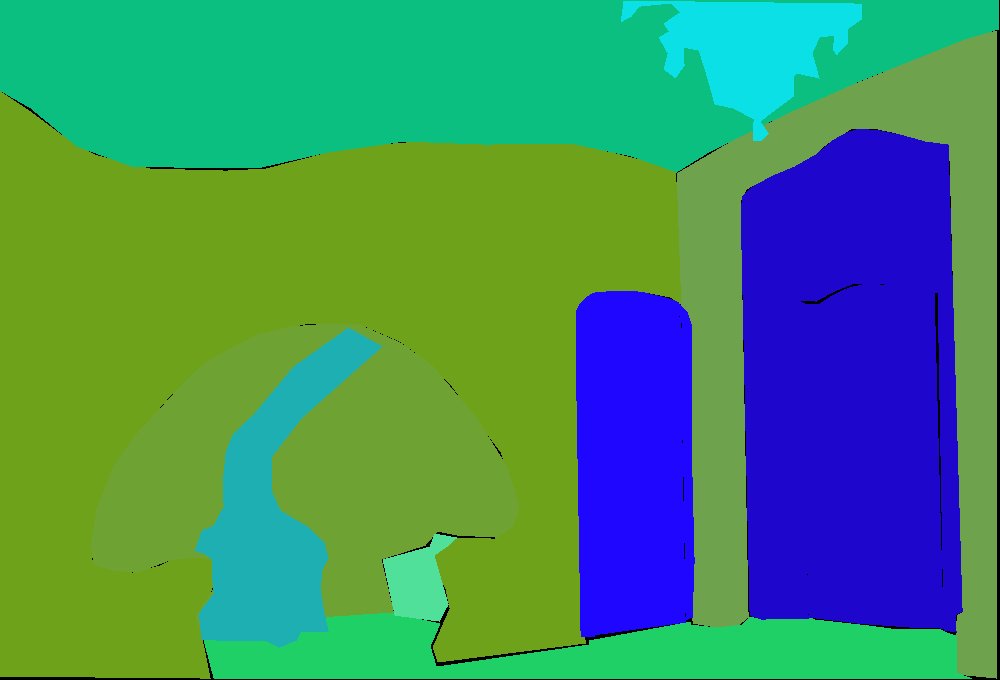} & \includegraphics[width=0.145\textwidth]{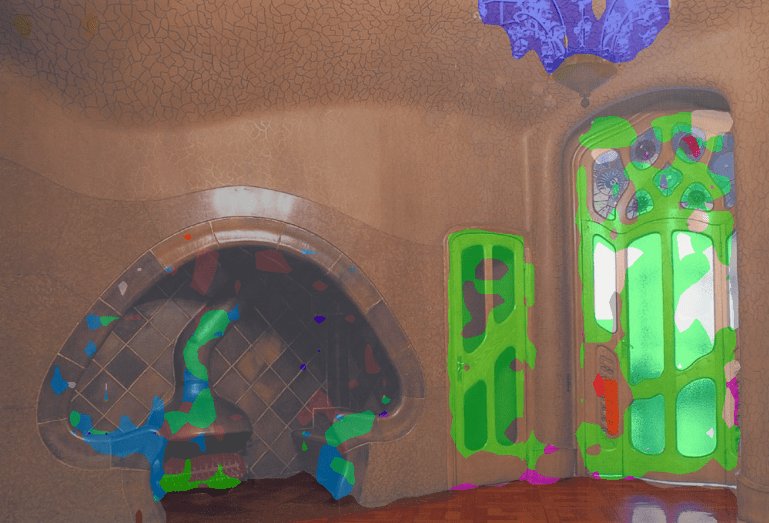} & \includegraphics[width=0.145\textwidth]{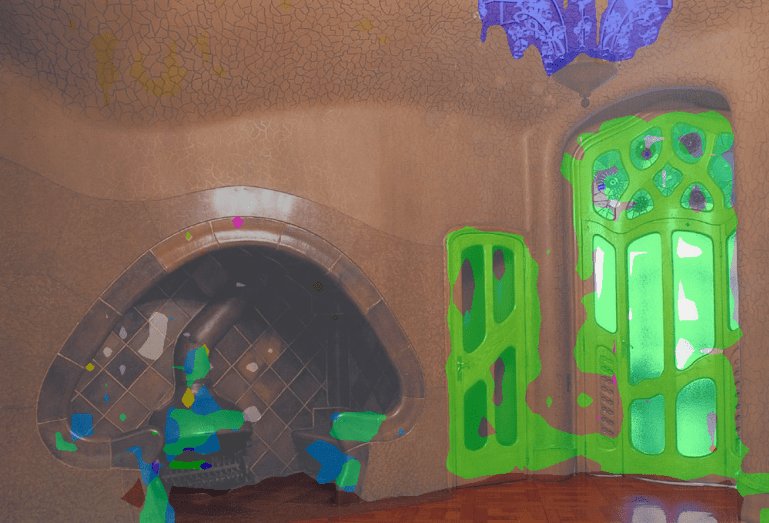}  & \includegraphics[width=0.145\textwidth]{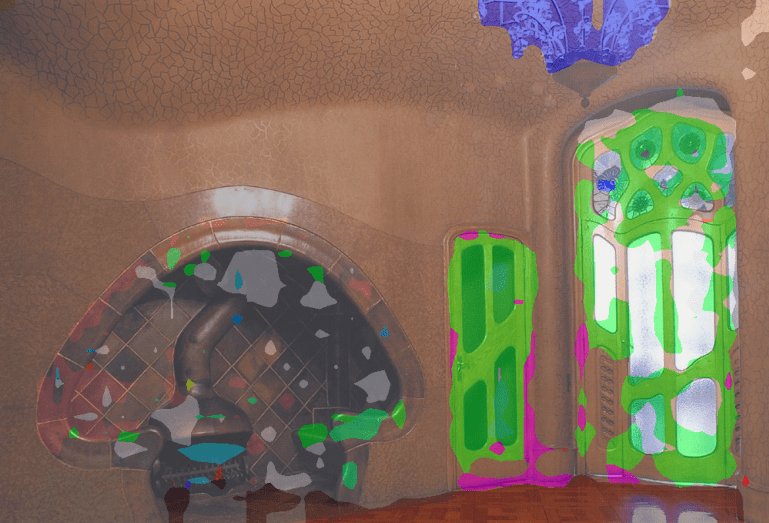}\\

\includegraphics[width=0.145\textwidth]{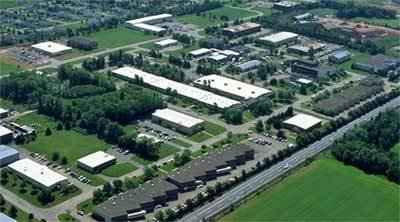} & \includegraphics[width=0.145\textwidth]{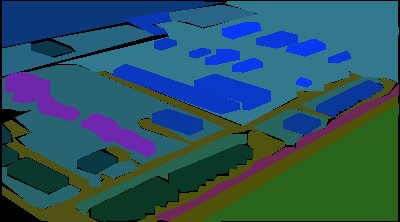} & \includegraphics[width=0.145\textwidth]{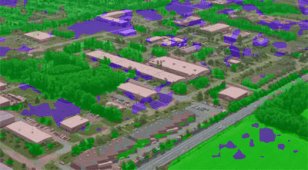} & \includegraphics[width=0.145\textwidth]{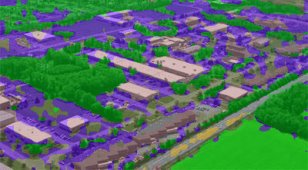}  & \includegraphics[width=0.145\textwidth]{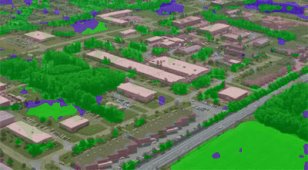}\\

\includegraphics[width=0.145\textwidth]{image_files/semantic_segmentation/raw/ADE_val_00001657.jpg} & \includegraphics[width=0.145\textwidth]{image_files/semantic_segmentation/ground_truth/ADE_val_00001657_seg.jpg} & \includegraphics[width=0.145\textwidth]{image_files/semantic_segmentation/linear_relevant/aclip_25epoch/ADE_val_00001657_seg.jpg} & \includegraphics[width=0.145\textwidth]{image_files/semantic_segmentation/linear_relevant/slip_25epoch/ADE_val_00001657_seg.jpg}  & \includegraphics[width=0.145\textwidth]{image_files/semantic_segmentation/linear_relevant/iclip_25epoch/ADE_val_00001657_seg.jpg}\\

\includegraphics[width=0.145\textwidth]{image_files/semantic_segmentation/raw/ADE_val_00001715.jpg} & \includegraphics[width=0.145\textwidth]{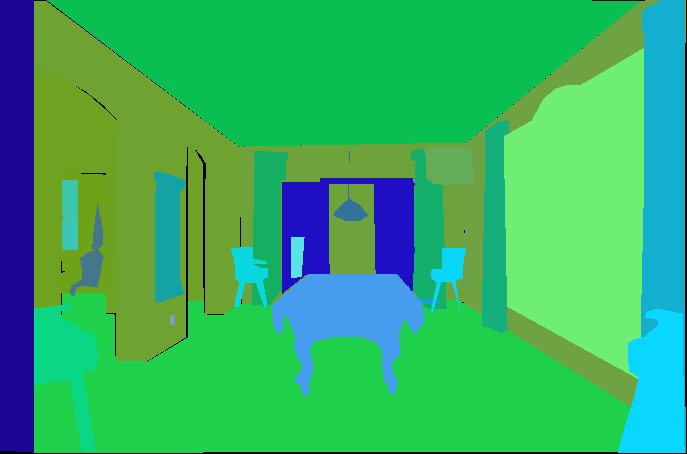} & \includegraphics[width=0.145\textwidth]{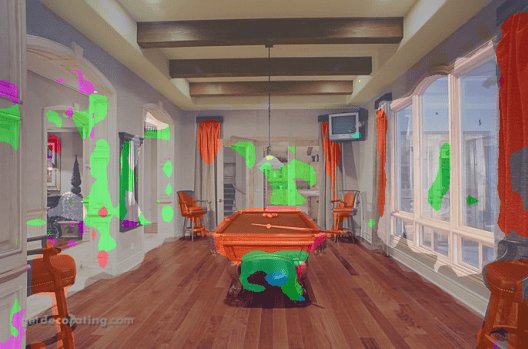} & \includegraphics[width=0.145\textwidth]{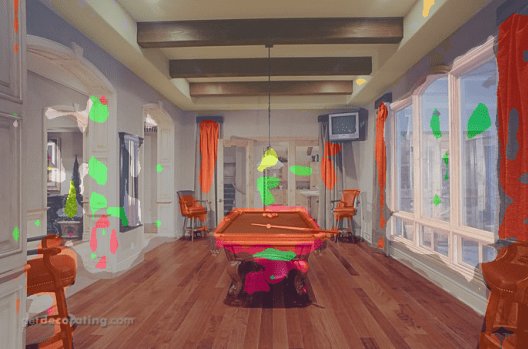}  & \includegraphics[width=0.145\textwidth]{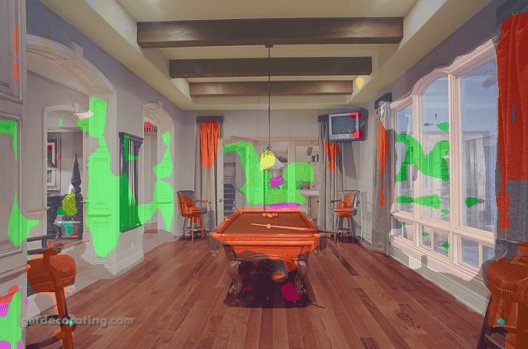}\\

\includegraphics[width=0.145\textwidth]{image_files/semantic_segmentation/raw/ADE_val_00001793.jpg} & \includegraphics[width=0.145\textwidth]{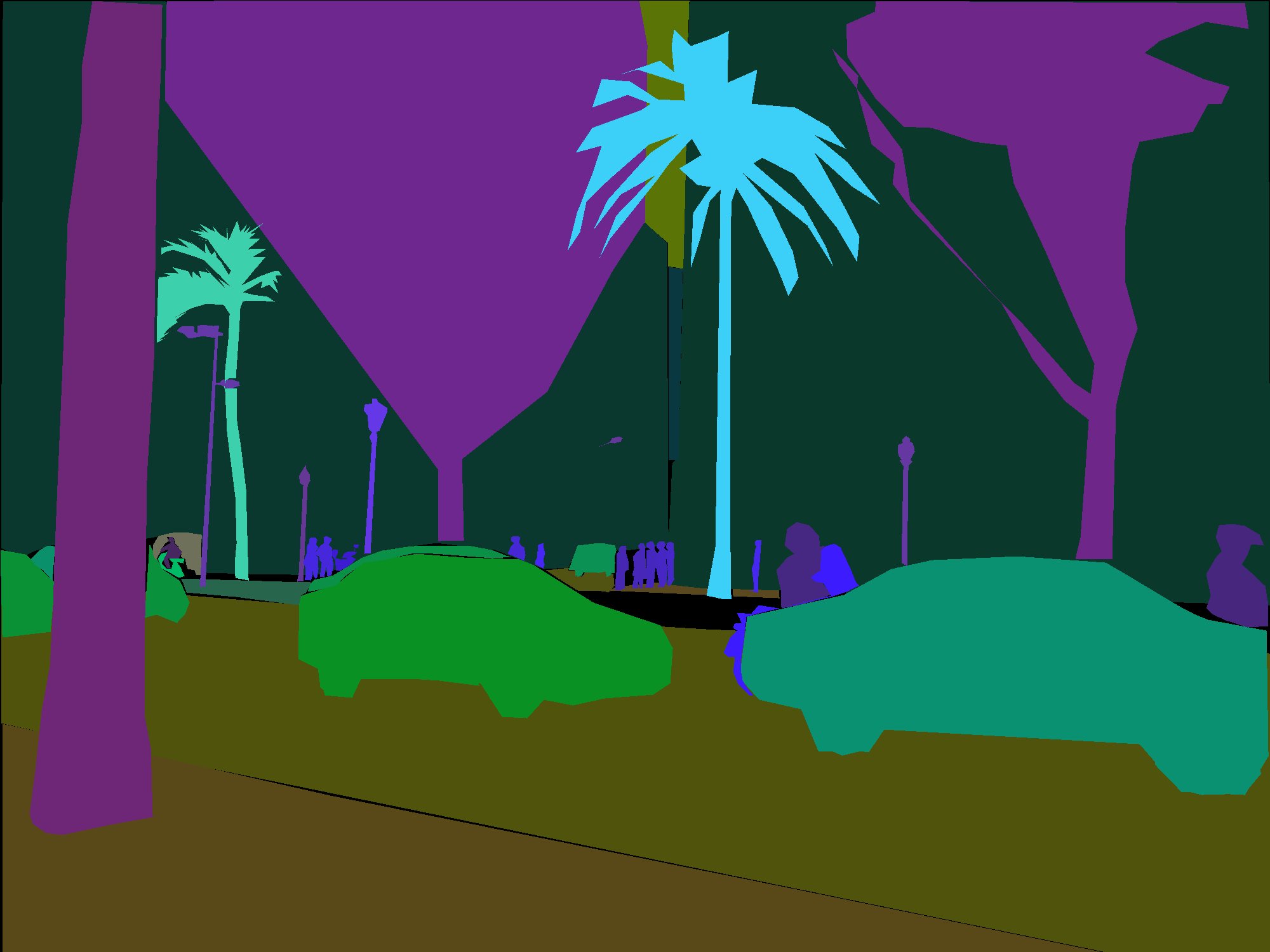} & \includegraphics[width=0.145\textwidth]{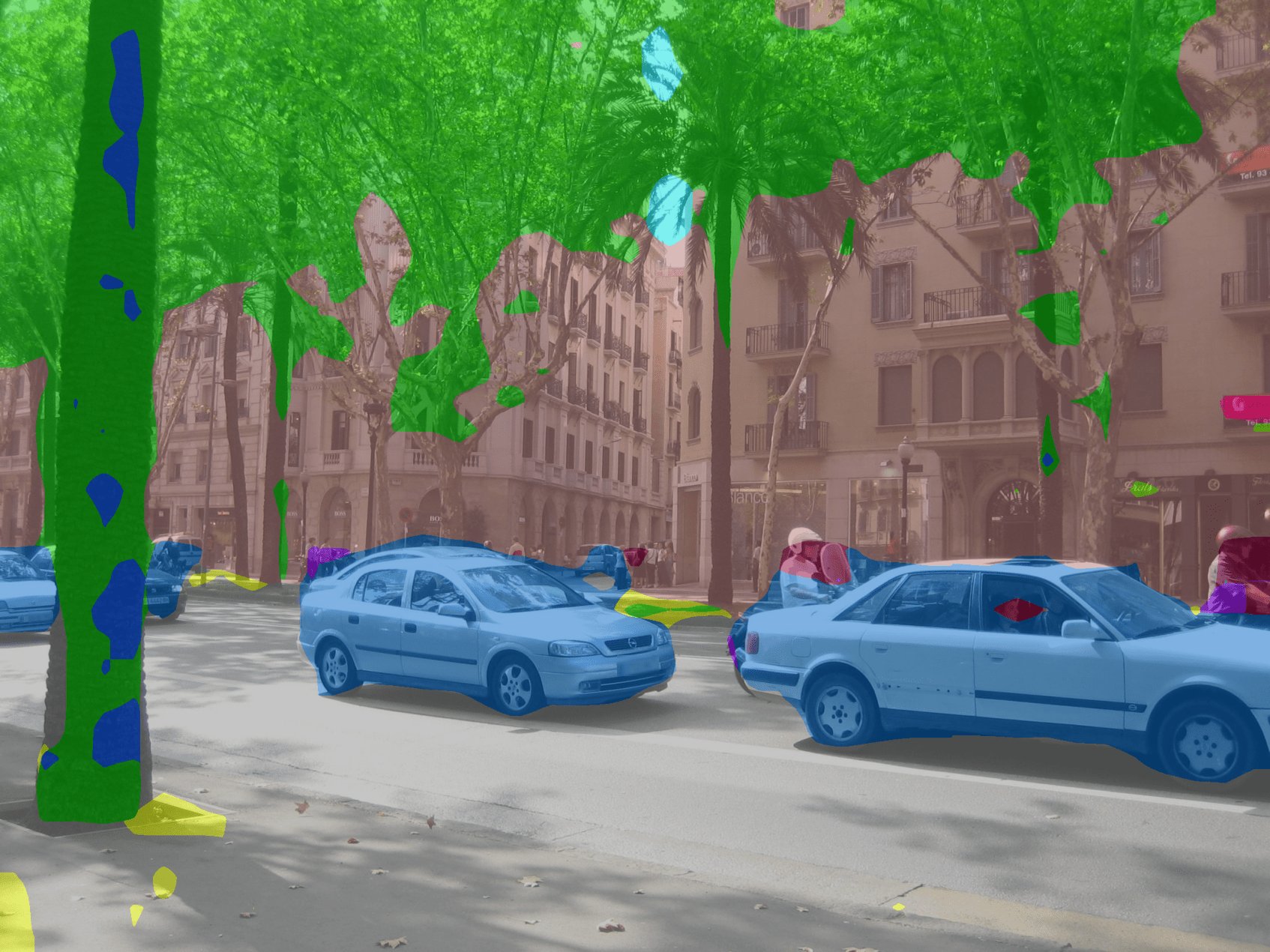} & \includegraphics[width=0.145\textwidth]{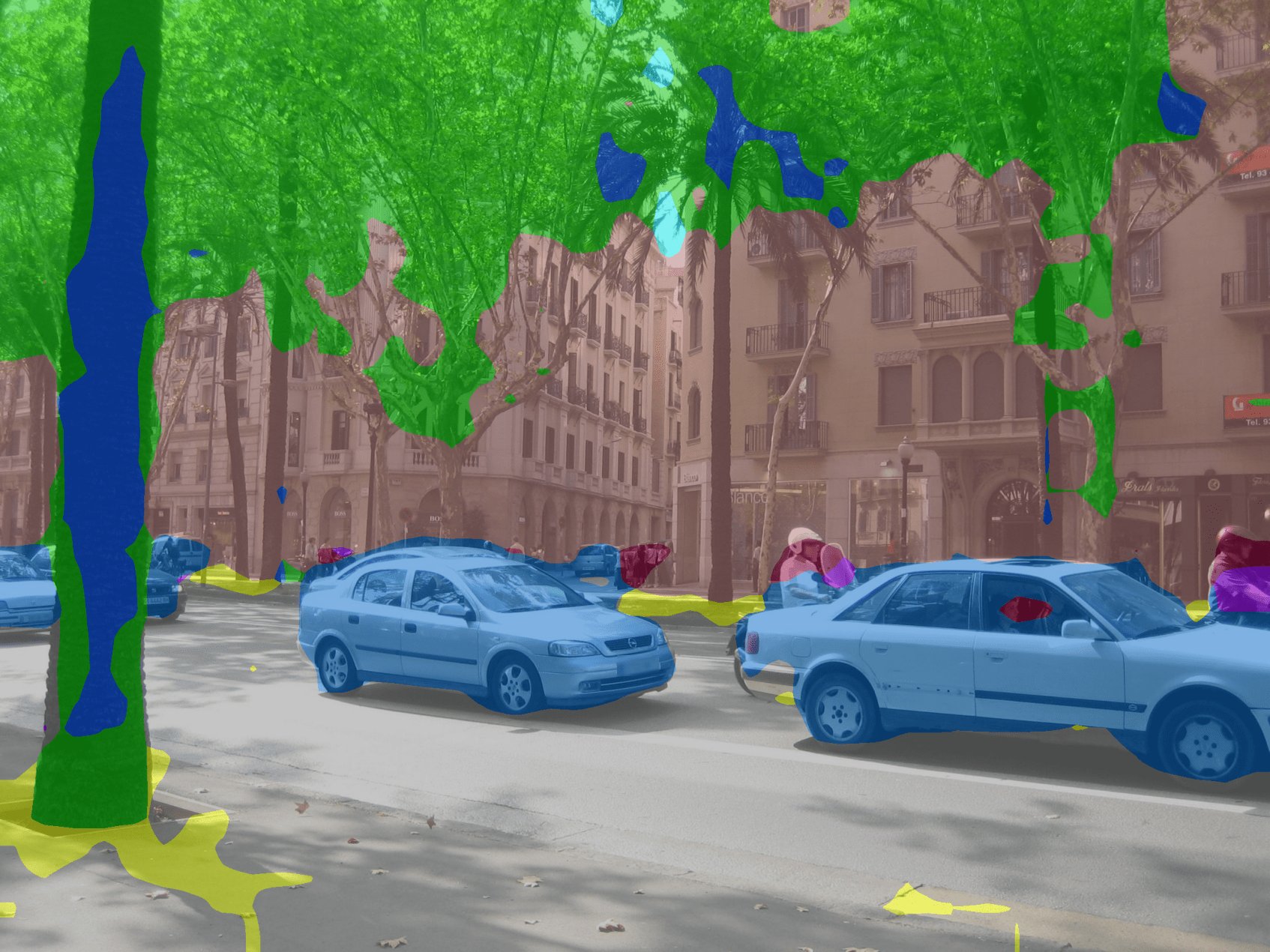}  & \includegraphics[width=0.145\textwidth]{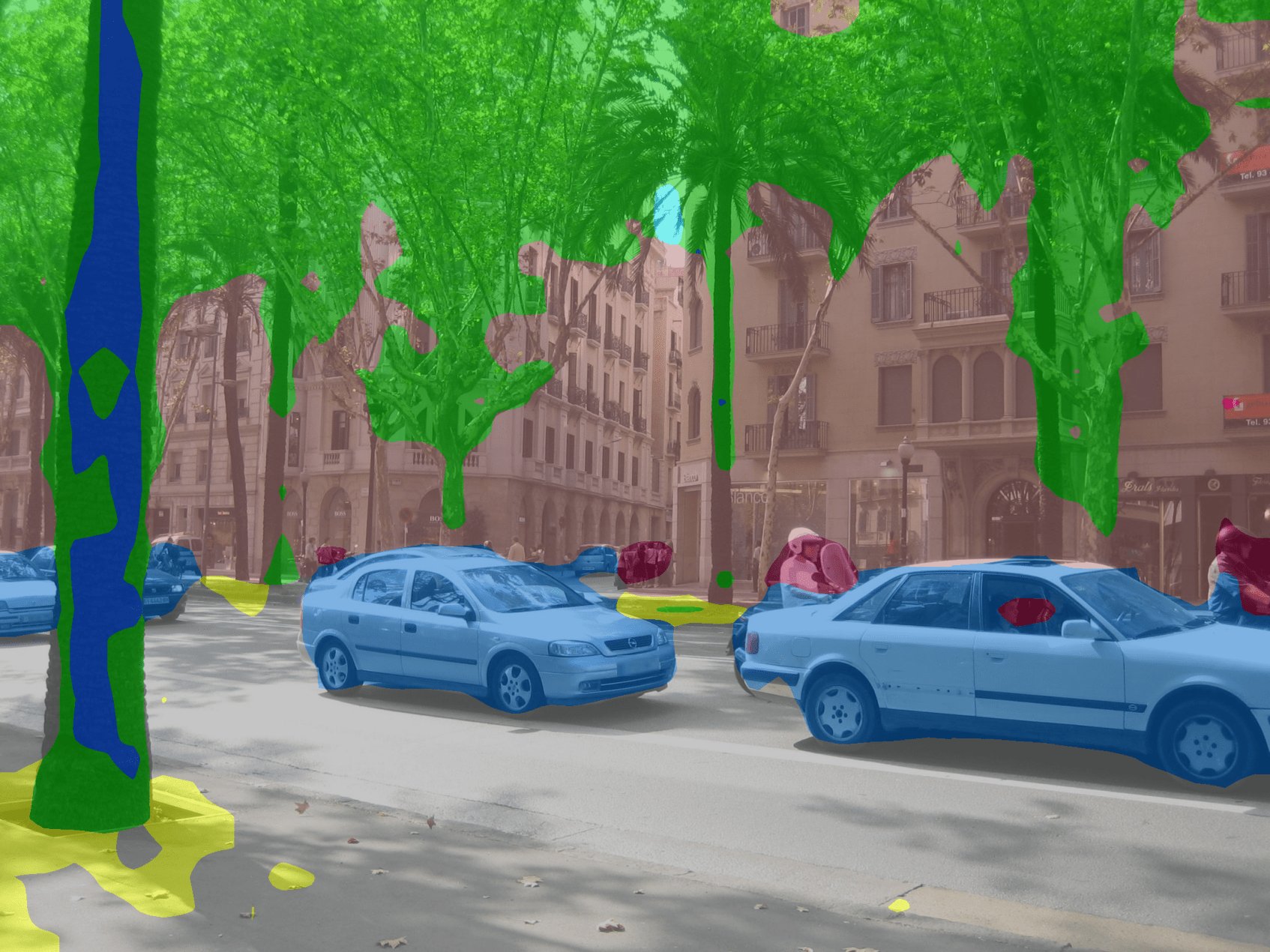}\\

\includegraphics[width=0.145\textwidth]{image_files/semantic_segmentation/raw/ADE_val_00001880.jpg} & \includegraphics[width=0.145\textwidth]{image_files/semantic_segmentation/ground_truth/ADE_val_00001880_seg.jpg} & \includegraphics[width=0.145\textwidth]{image_files/semantic_segmentation/linear_relevant/aclip_25epoch/ADE_val_00001880_seg.jpg} & \includegraphics[width=0.145\textwidth]{image_files/semantic_segmentation/linear_relevant/slip_25epoch/ADE_val_00001880_seg.jpg}  & \includegraphics[width=0.145\textwidth]{image_files/semantic_segmentation/linear_relevant/iclip_25epoch/ADE_val_00001880_seg.jpg}\\

\hline
\hline

\end{tabular}
\caption{Visual comparison of segmentation results using the Linear decoder across different models. The comparison includes SLIP, A-CLIP, and DetailCLIP at 25 epochs.}
\label{sup:fig:seg:linear}
\end{figure*}

This section presents a comprehensive comparative analysis of DetailCLIP's performance against several baseline models, primarily focusing on segmentation and object detection tasks. Through a series of detailed visualizations, we provide side-by-side comparisons that emphasize the superior capability of DetailCLIP in capturing and reconstructing fine-grained details, which is crucial for achieving high segmentation accuracy. Specifically, Figure \ref{fig:seg:upernet} showcases the semantic segmentation results using the UPerNet decoder, while Figure \ref{sup:fig:seg:linear} highlights the results with a linear decoder. These figures emphasize the notable improvements achieved by DetailCLIP in handling complex images, particularly when it comes to accurately delineating boundaries and preserving detailed features, areas where baseline models often struggle. The segmentation comparisons are conducted using the ADE20K dataset.

\begin{figure*}[ht]
\centering
\begin{tabular}{cc}
\hline
Image + Objects & Image + Objects \\ \hline \\

\includegraphics[width=0.38\textwidth]{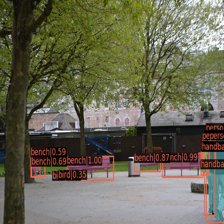} & \includegraphics[width=0.38\textwidth]{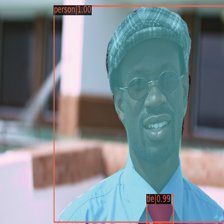}  \\

\includegraphics[width=0.38\textwidth]{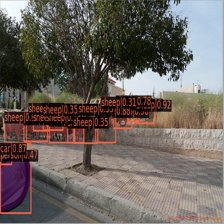} & \includegraphics[width=0.38\textwidth]{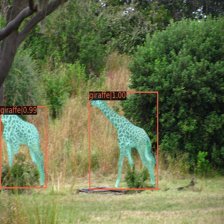}  \\

\includegraphics[width=0.38\textwidth]{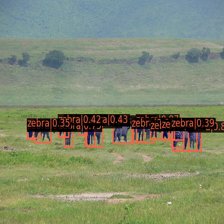} & \includegraphics[width=0.38\textwidth]{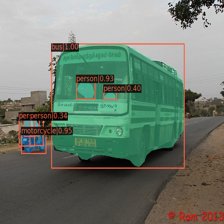}  \\

\hline
\hline

\end{tabular}
\caption{Object detection results using DetailCLIP after 50 training epochs. The image highlights DetailCLIP's ability to detect objects accurately, showcasing its precision and effectiveness in complex scenes}
\label{fig:object:eval}
\end{figure*}

\begin{figure*}[ht]
\centering
\begin{tabular}{cc}
\hline
Image + Objects & Image + Objects \\ \hline \\

\includegraphics[width=0.38\textwidth]{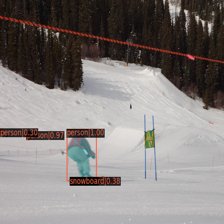} & \includegraphics[width=0.38\textwidth]{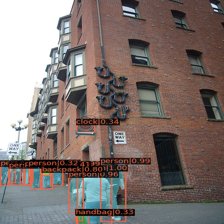}  \\

\includegraphics[width=0.38\textwidth]{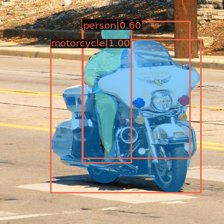} & \includegraphics[width=0.38\textwidth]{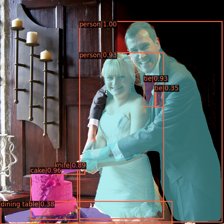}  \\

\includegraphics[width=0.38\textwidth]{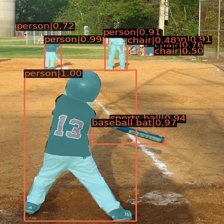} & \includegraphics[width=0.38\textwidth]{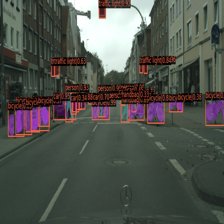}  \\

\hline
\hline

\end{tabular}
\caption{Object detection results using DetailCLIP after 50 training epochs. The image highlights DetailCLIP's ability to detect objects accurately, showcasing its precision and effectiveness in complex scenes}
\label{fig:object:eval:1}
\end{figure*}

In addition to segmentation, Figures \ref{fig:object:eval} and \ref{fig:object:eval:1} present the object detection results using the MS COCO dataset, further demonstrating the effectiveness of DetailCLIP across multiple vision tasks. By visualizing these outputs, we showcase DetailCLIP's enhanced performance and shed light on the limitations of other models, such as SLIP and A-CLIP. These comparisons underscore the significant advancements made by DetailCLIP in tackling detail-oriented tasks, solidifying its robustness and versatility in a wide array of visual challenges.

Overall, these visual and quantitative evaluations illustrate the significant advancements made by DetailCLIP in addressing detail-oriented challenges across both segmentation and object detection. The side-by-side comparisons solidify DetailCLIP's robustness and adaptability, making it a more effective and versatile solution for a wide range of fine-grained vision tasks.

\end{document}